\documentclass{egpubl}  
\usepackage{eg2022}
\usepackage{float}      

\STAR                   

\usepackage[T1]{fontenc}
\usepackage{dfadobe}  

\usepackage{cite}  
\BibtexOrBiblatex
\electronicVersion
\PrintedOrElectronic
\ifpdf \usepackage[pdftex]{graphicx} \pdfcompresslevel=9
\else \usepackage[dvips]{graphicx} \fi

\usepackage{egweblnk} 

\usepackage{xcolor}
\usepackage{soul}

\usepackage{amsmath}
\usepackage{amsfonts}
\usepackage{booktabs}
\usepackage{cleveref}
\usepackage{paralist}
\usepackage{adjustbox}
\usepackage{todonotes}
\usepackage{textcomp}
\usepackage{nicefrac}

\usepackage{mdframed}
\newmdtheoremenv{definition}{Definition}

\newcommand{\jt}[1]{}

\newcommand{\numpapers}{250 }

\newcommand{\field}{\Phi}

\newcommand{\coord}{\mathbf{x}}
\newcommand{\quantity}{\mathbf{q}}
\newcommand{\sensorfeats}{\mathbf{t}_{\text{sens}}}
\newcommand{\sensorcoord}{\coord_{\text{sens}}}
\newcommand{\reccoord}{\coord_{\text{recon}}}
\newcommand{\recfeats}{\mathbf{y}_{\text{recon}}}
\newcommand{\sensor}{\Omega}
\newcommand{\params}{\Theta}
\newcommand{\latent}{\mathbf{z}}
\newcommand{\latentdistrib}{\mathcal{Z}}

\newcommand{\hyponet}[1]{\color{red}\mathbf{\Phi}}

\newcommand{\ie}{\emph{i.e.,}}
\newcommand{\eg}{\emph{e.g.,}}

\newcommand{\parahead}[1]{\paragraph*{#1}}

\newcommand{\posteg}[1]{}

\newenvironment{packed_itemize}
{\begin{itemize}
    \setlength{\itemsep}{1pt}
    \setlength{\parskip}{0pt}
    \setlength{\parsep}{0pt}
}{\end{itemize}}

\newcommand{\chapter}[1]
{
\vspace{0.5cm}
\begin{center}
    {\Large \textbf{#1}}
\end{center}
}

\DeclareMathOperator*{\argmax}{arg\,max}
\DeclareMathOperator*{\argmin}{arg\,min}

\usepackage{longtable}
\usepackage{pdflscape}
\usepackage{subfig}

\title[Neural Fields in Visual Computing]%
      {Neural Fields in Visual Computing and Beyond}
\author[Y.~Xie, T.~Takikawa, S.~Saito, O.~Litany, S.~Yan, N.~Khan, F.~Tombari, J.~Tompkin, V.~Sitzmann, S.~Sridhar]
{\parbox{\textwidth}{\centering 
    Yiheng Xie$^{1,2}$\orcid{0000-0002-2689-3471}  
    Towaki Takikawa$^{3,4}$  
    Shunsuke Saito$^{5}$\orcid{0000-0003-2053-3472}  
    Or Litany$^{4}$\orcid{0000-0001-6700-7379}  
    Shiqin Yan$^{1}$  
    Numair Khan$^{1}$\orcid{0000-0002-7787-5177}  
    Federico~Tombari$^{6,7}$  
    James~Tompkin$^{1}$\orcid{0000-0003-2218-2899}  
    Vincent~Sitzmann$^{8\dag}$\orcid{0000-0002-0107-5704}  
    Srinath~Sridhar$^{1\dag}$\orcid{0000-0003-4663-3324}
}
\\
{\parbox{\textwidth}{\centering 
    $^1$Brown~University\;
    $^2$Unity~Technologies\;
    $^3$University~of~Toronto\;
    $^4$NVIDIA\;
    $^5$Meta Reality Labs Research\;
    $^6$Google\;
    $^7$Technical~University~of~Munich\;
    $^8$Massachusetts~Institute~of~Technology\;
    $^\dag$\textit{Equal advising}\;
}}
\\
{\parbox{\textwidth}{\centering 
}}
}

\begin{document}

\teaser{
    \vspace{-1.5cm}
    \centering
    \subfloat{\href{https://neuralfields.cs.brown.edu/}{\includegraphics[width=\linewidth]{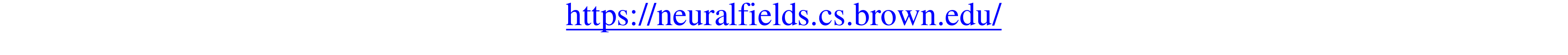}}}\\
    \subfloat{\includegraphics[height=3.3in]{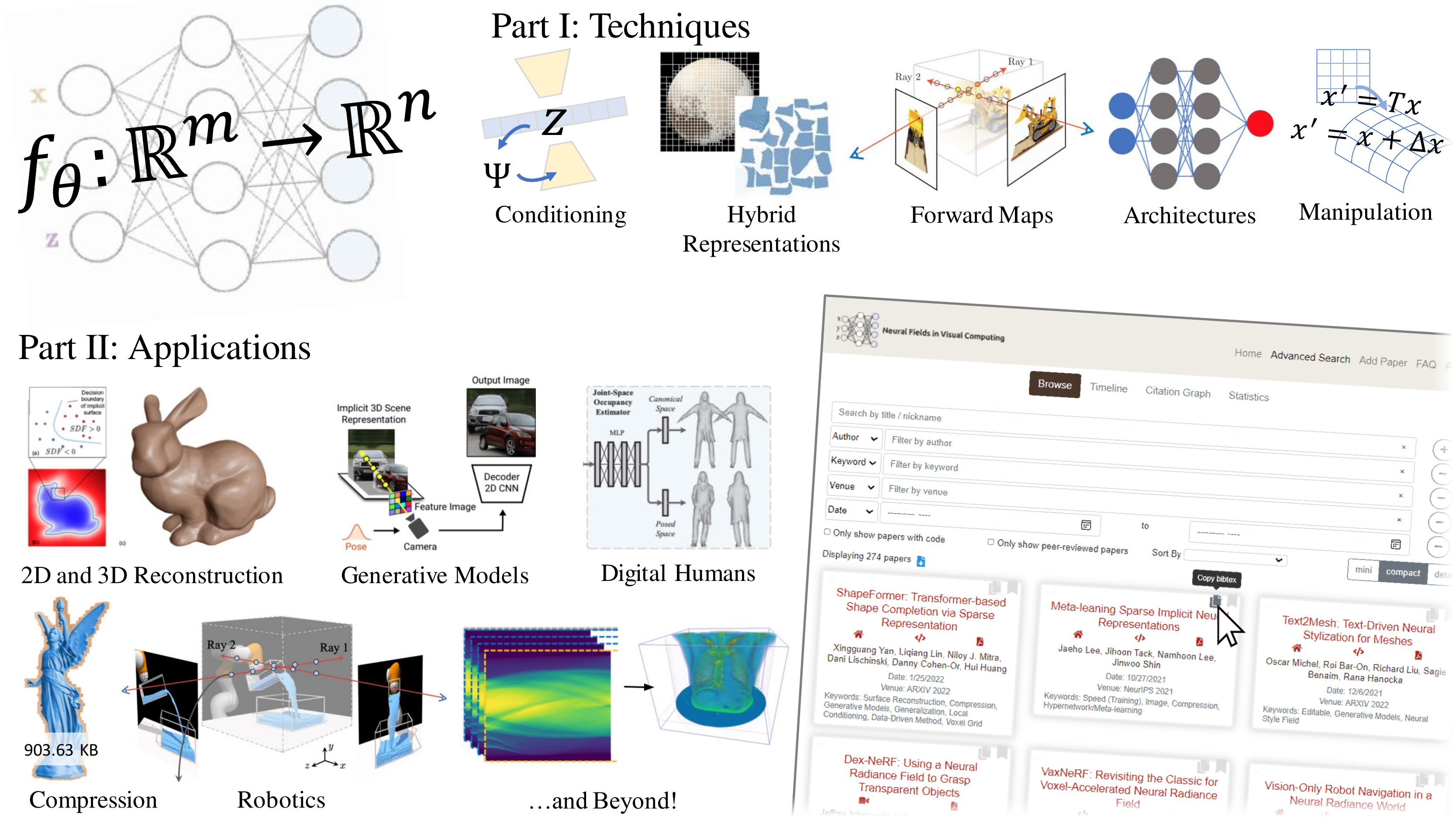}}
    \caption{\textbf{Contribution of this report.} Following a survey of over \numpapers papers, we provide a review of (\textbf{Part I}) techniques in neural fields such as prior learning and conditioning, representations, forward maps, architectures, and manipulation, and of (\textbf{Part II}) applications in visual computing including 2D image processing, 3D scene reconstruction, generative modeling, digital humans, compression, robotics, and beyond.
    This report is complemented by a \href{https://neuralfields.cs.brown.edu/}{community-driven website} with search, filtering, bibliographic, and visualization features.}
    \vspace{0.25cm}
    \label{fig:teaser}
}

\maketitle

\begin{abstract}
Recent advances in machine learning have led to increased interest in solving visual computing problems using methods that employ coordinate-based neural networks.
These methods, which we call \textbf{neural fields}, parameterize physical properties of scenes or objects across space and time.
They have seen widespread success in problems such as 3D shape and image synthesis, animation of human bodies, 3D reconstruction, and pose estimation.
Rapid progress has led to numerous papers, but a consolidation of the discovered knowledge
has not yet emerged.
We provide context, mathematical grounding, and a review of over \numpapers papers in the literature on neural fields.
In \textbf{Part I}, we focus on \emph{neural field techniques} by identifying common components of neural field methods, including different conditioning, representation, forward map, architecture, and manipulation methods.
In \textbf{Part II}, we focus on \emph{applications of neural fields} to different problems in visual computing, and beyond (\eg~robotics, audio).
Our review shows the breadth of topics already covered in visual computing, both historically and in current incarnations, and highlights the improved quality, flexibility, and capability brought by neural field methods.
Finally, we present a \href{https://neuralfields.cs.brown.edu/}{companion website} that acts as a living database that can be continually updated by the community.


\end{abstract}  
\section{Introduction}
\label{sec:intro}

Visual computing involves the synthesis, estimation, manipulation, display, storage, and transmission of data about objects and scenes across space and time.
In computer graphics, we synthesize 3D shapes and 2D images, render novel views of scenes, and animate articulating human bodies. 
In computer vision, we reconstruct 3D appearance, shape, object pose, and deformation. 
In human-computer interaction (HCI), we enable the interactive investigation of spacetime data. 
Beyond visual computing and into adjacent disciplines like robotics, we use the structure of 3D scenes to plan and execute actions. 
Within these disciplines, \textbf{fields} are widely used to continuously parameterize an underlying physical quantity of an object or scene over space and time. For instance, fields have been used to visualize physical phenomena~\cite{sabella1988rendering}, compute image gradients~\cite{schluns1997local}, compute collisions~\cite{osher2003signed}, or represent shapes via constructive solid geometry (CSG)~\cite{evans2015learning}.

In visual computing tasks, we solve many different optimization problems, including using machine learning and training data.
One class of methods has gained significant attention since 2019: coordinate-based neural networks that represent a field.
We refer to these methods as \textbf{neural fields}.
Given sufficient parameters, fully connected neural networks can encode continuous signals over arbitrary dimensions at arbitrary resolution.
Recent success with neural networks has caused a resurgence of interest in visual computing problems, leading to more accurate, higher fidelity, more expressive, and memory-efficient solutions.
This has allowed neural fields to gain traction as a useful parameterization of 2D images~\cite{karras2021alias}, 3D shape~\cite{park2019deepsdf,mescheder2019occupancynetworks,chen2019imnet}, view-dependent appearance~\cite{sitzmann2019srn,mildenhall2020nerf}, and human bodies and faces\cite{niemeyer2019occupancyflow, deng2020nasa, saito2021scanimate, yenamandra2021i3dmm,ramon2021h3dnet}.

\begin{figure}[t]
    \centering
    \includegraphics[width=\linewidth]{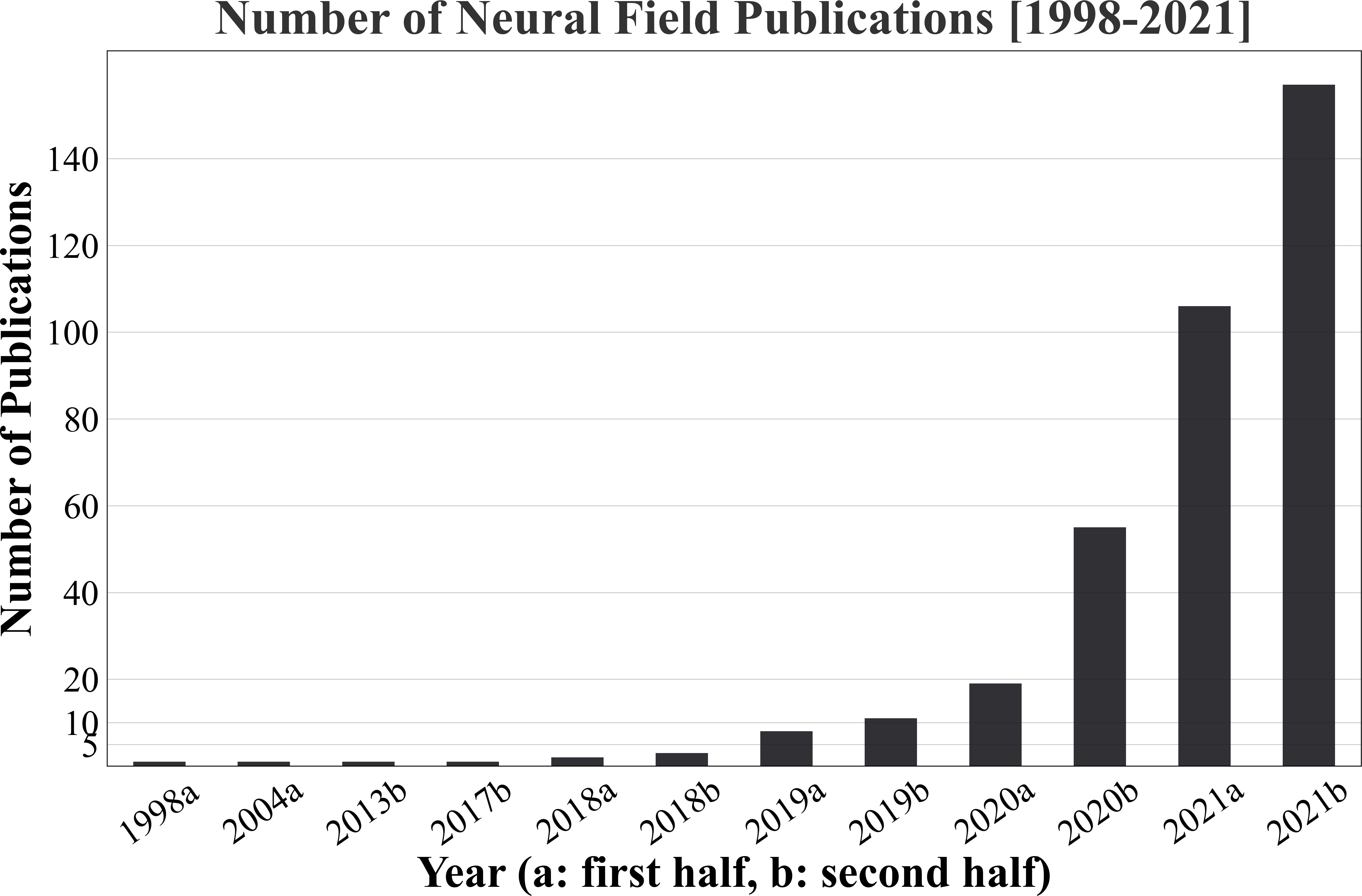}
    \caption{Neural fields were proposed two decades ago, yet their growth in visual computing has been concentrated in the last two years with over \numpapers papers.}
    \label{fig:explosion}
    \vspace{-15pt}
\end{figure}

Such attention manifests as rapid progress and an explosion of papers (Figure~\ref{fig:explosion}), with a need to consolidate the discovered knowledge.
However, a shared mathematical formulation to describe related techniques has not yet emerged, making it hard to communicate ideas and train students.
Furthermore, there is ``\emph{selective amnesia}''~\cite{su2021affective} of older or even concurrent works, causing research repetition.
Finally, rapid progress makes any survey quickly out-of-date, requiring new summarization approaches.

Following a survey of over \numpapers papers, we address the above issues by defining a neural field via fields of physical quantities, providing a shared mathematical formulation across common techniques, categorizing, describing, and relating many applications, and presenting a \textbf{living database} via a community website.

In \textbf{Part I}, we describe \emph{neural field techniques} that are common across papers with a consistent notation and vocabulary. For instance, we identify and formalize recent hybrid discrete-continuous representations of neural fields, and techniques for learning priors such as local and global conditioning and meta-learning. Further, neural fields can be combined with a wide variety of differentiable forward maps, and we identify many such maps including surface and volume renderers and partial differential equations.

In \textbf{Part II}, we describe a broad cross-section of \emph{applications of neural fields} to problems in visual computing. This lets us identify commonalities, connections, and trends across works representing shape and appearance of scenes and objects, including 3D reconstruction, digital humans, generative modeling, data compression, and 2D image processing. We also review works that solve tasks in adjacent communities including robotics (Section~\ref{sec:robotics}), medical imaging (Section~\ref{sec:ct_mri}), audio processing (Section~\ref{sec:audio}), and physics-informed problems (Section~\ref{sec:PINN}).

Our living database is a \href{https://neuralfields.cs.brown.edu/}{companion website} that provides search, filter, and visualization features to present the works described in this report (Figure~\ref{fig:teaser}), along with bibliography exporting features. 
The website allows our community to submit new works in neural fields and automatically updates the databases using keyword categorizations from Part I and II taxonomies.
The website requires minimal maintenance, and its source code is open to allow future reports and surveys to provide the same functionality.

In summary, neural field techniques are powerful and widely applicable to visual computing problems and beyond. This state-of-the-art report provides the mathematical formulation of techniques and discussion of applications to make sense of this exciting area, along with a community-driven website to continue to help researchers keep track of new developments in the future.

\subsection{Background}
\label{sec:background}

\begin{table*}[t]
\adjustbox{valign=t}{\begin{minipage}[t]{0.25\textwidth}
    \caption{\textbf{Examples of fields in physics and visual computing.}}
    \label{tab:field_examples}
\end{minipage}}
\quad
\adjustbox{valign=t}{\begin{minipage}[t]{0.75\textwidth}
    \begin{tabular}{l l l l l}
    \toprule
    Examples                            & Field Quantity             & Scalar/Vector & Coordinates           \\ 
    \midrule
    Gravitational Field                 & Force per unit mass (N/kg) & Vector        & $\mathbb{R}^n$        \\
    3D Paraboloid: $z = x^2 + y^2$ & Height $z$                 & Scalar        & $\mathbb{R}^2$            \\
    2D Circle: $r^2 = x^2 + y^2$          & Radius $r$                 & Scalar        & $\mathbb{R}^2$                \\
    Signed Distance Field (SDF)         & Signed distance            & Scalar        & $\mathbb{R}^n$           \\
    Occupancy Field                     & Occupancy                  & Scalar        & $\mathbb{R}^n$             \\
    Image                               & RGB intensity                  & Vector        & $\mathbb{Z}^2$ pixel locations $x, y$         \\
    Audio                               & Amplitude                  & Scalar        & $\mathbb{Z}^1$ time $t$           \\ 
    \bottomrule
    \end{tabular}
\end{minipage}}
\vspace{-10pt}
\end{table*}

\parahead{Fields}
Following the original definition in physics~\cite{feynman1965feynman,mcmullin2002origins} we define fields as the following:

\vspace{0.25cm}
\begin{definition}
A \emph{field} is a quantity defined for all spatial and/or temporal coordinates.
\end{definition}

We can represent a field as a function mapping a coordinate $\coord$ to a quantity, which is typically a scalar or vector. Table~\ref{tab:field_examples} provides several examples. The physics community has also studied spinor and tensor fields. Finally, field coordinates are not limited to space and time, such as frequency coordinates in spectrograms. This review will focus on scalar and vector fields defined over spacetime, which are most relevant to visual computing.

In practice, the underlying field generation process may not have a known analytic form. Thus, functions may be described by parameters $\params$ that are hand crafted, optimized, or learned.
We denote such a field as producing quantity $\quantity = \field(\coord; \params)$.
Furthermore, we often index sampled functions using discrete values, such as at camera \emph{pixels}, or choose discrete function parameterizations using \emph{voxels} or discretized \emph{level sets}.
However, discrete parameterizations are limited by the Nyquist sampling rate, causing high memory requirements for 3D tasks.
Adaptive grids like octrees and k--d trees can reduce memory, but their generation can lead to costly combinatoric optimizations.

\parahead{Neural Networks}
A neural network connects many layers of artificial neurons to learn to non-linearly map a fixed-size input to a fixed-size output~\cite{lecun2015deep}.
A multi-layer perceptron (MLP) neural network can approximate any function through their learned parameters (universal approximation theorem~\cite{kim2003approximation}).
Since 2010, there has been significant interest in using neural networks with corresponding investment in hardware and software to support neural networks.
This has significantly lowered the barrier to apply neural networks to a wide range of problems.

\parahead{Neural Fields}
Following the universal approximation theorem, any field can be parameterized by an MLP neural network. Thus:
\vspace{0.25cm}
\begin{definition}
A \emph{neural field} is a field that is parameterized fully or in part by a neural network.
\label{def:neuralfield}
\end{definition}

Neural fields are both continuous and adaptive by construction. 
Unlike the memory required for discrete parameterizations that scales poorly with spatio-temporal resolution, the memory required for neural fields instead scales with the number of parameters of the neural network---so-called network complexity. While other continuous parameterizations can represent large extents (for example, a Fourier series is a parameterization of a field), it is often difficult to know the required complexity ahead of time for efficient representation.
Neural fields help to resolve this problem by using their parameters only where field detail is present.
Neural fields are often parameterized as MLPs with activation functions whose gradients are well-defined.
By their analytic differentiability, gradient descent, and over-parameterization~\cite{frankle2018lottery}, neural fields are effective at regressing complex signals through optimizations for problems that are otherwise ill-posed.

Our definition of a neural field does not include neural networks whose co-domain has a spatial extent, for instance, a network that outputs a grid of voxels.
These methods often use convolutional layers and output a 2D or 3D grid of RGB color, occupancy, or latent features~\cite{sitzmann2019deepvoxels, lombardi2019nv}. 
While the regular grid output of these architectures can be seen as samplings of fields, the functions parameterized by the neural network are not neural fields as they do not ingest spatio-temporal coordinates.

\parahead{Terminology}
Within visual computing, neural fields have been called \emph{implicit neural representations}, \emph{neural implicits}, or \emph{coordinate-based neural networks}.
In neuroscience, the term \emph{neural field} may describe theories about the organization and function of the brain~\cite{coombes2014neural}.
The term also describes a regular grid of neurons in a neural network~\cite{lemmon1991aneuralfieldspathplanning}.
We exclude works in these two areas as they do not follow from our definition.

\subsection{Related Surveys and Articles}

Within visual computing, there are survey papers on specific areas like neural rendering~\cite{tewari2020state,tewari2021advances} and 3D reconstruction~\cite{zollhofer2018state}, as well as lists of works in scene representations~\cite{NeRFExpl90:online,NeRFatIC64:online,Sitzmann_Awesome_Implicit_Representations,NeRFNeur57:online}.
Our survey aims to connect our visual computing communities together and share findings. 
Finally, neural fields are domain-agnostic and can model arbitrary quantities beyond shape and appearance. Neural fields have been used extensively in computational physics via physics-informed neural networks (PINNs)~\cite{raissi2019physics}, with an existing survey~\cite{karniadakis2021physics}.

\chapter{Part I. Neural Field Techniques}
\begin{figure*}[t]
    \centering
    \includegraphics[width=0.95\linewidth]{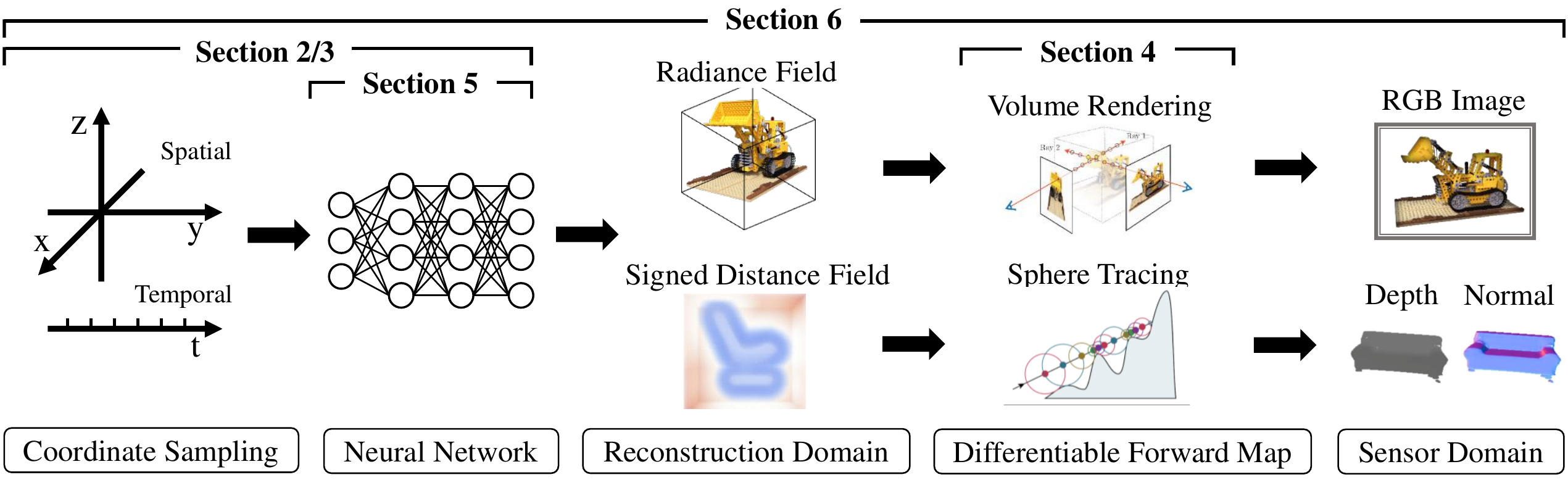}
    \vspace{-0.25cm}
    \caption{\textbf{A typical feed-forward neural field algorithm.} Spatiotemporal coordinates are fed into a neural network which predicts values in the reconstruct a domain. Then, this domain is mapped to the sensor domain where sensor measurements are available as supervision. Figures adapted from \cite{mildenhall2020nerf, liu2020dist}.}
    \label{fig:part1}
    \vspace{-10pt}
\end{figure*}

\begin{table*}[t]
\centering
\begin{tabular}{l l l}
\toprule
\textbf{Class and Section} & \textbf{Problems Addressed}                                                       \\
\midrule
Prior Learning and Conditioning (Section~\ref{sec:generalization})               & Inverse problems, ill-posed problems, edit ability, symmetries. \\
Hybrid Representations (Section~\ref{sec:hybrid_rep})           & Computation \& memory efficiency, representation capacity, edit ability. \\
Forward Maps (Section~\ref{sec:forward_maps})                   & Inverse problems.                                                       \\
Network Architecture (Section~\ref{sec:networkarch})            & Spectral bias, integration and derivatives.                      \\
Manipulating Neural Fields (Section~\ref{sec:editing})          & Edit ability, constraints, regularization.                                            \\
\bottomrule
\end{tabular}
\caption{The five classes of techniques in the neural field toolbox each addresses problems that arise in learning,  inference, and control.}
\label{tab:part1}
\vspace{-10pt}
\end{table*}
\label{sec:part1}
A typical neural fields algorithm in visual computing proceeds as follows (Figure \ref{fig:part1}): Across space-time, we \emph{sample coordinates} and feed them into a \emph{neural network} to produce field quantities. The field quantities are samples from the desired \emph{reconstruction domain} of our problem. Then, we apply a \emph{forward map} to relate the reconstruction to the \emph{sensor domain} (e.g. RGB image), where supervision is available. Finally, we calculate the reconstruction error or \emph{loss} that guides the neural network optimization process by comparing the reconstructed signal to the sensor measurement. 

During this process, many problems affect our ability to successfully reconstruct the signal. 
To better understand and apply neural fields, we identify five classes of techniques (Table~\ref{tab:part1}). 
We can aid reconstruction from incomplete sensor signals via \emph{prior learning and conditioning} (Section~\ref{sec:generalization}). We can improve memory, computation, and neural network efficiency via \emph{hybrid representations} using discrete data structures (Section~\ref{sec:hybrid_rep}). We can supervise reconstruction via differentiable \emph{forward maps} that transform or project our domain (\eg~3D reconstruction via 2D images; Section~\ref{sec:forward_maps}). With appropriate \emph{network architecture} choices, we can overcome neural network spectral biases (blurriness) and efficiently compute derivatives and integrals (Section~\ref{sec:networkarch}). Finally, we can \emph{manipulate neural fields} to add constraints and regularizations, and to achieve editable representations (Section~\ref{sec:editing}).
Collectively, these classes constitute a `toolbox' of techniques to help solve problems with neural fields.

\section{Prior Learning and Conditioning}
\label{sec:generalization}
Suppose we wish to estimate a plausible 3D surface shape given a partial point cloud. This problem arises in reconstructing 3D street scenes from Lidar scans as we only observe points upon surfaces.
To accomplish this task, we need a suitable prior over 3D surfaces. One option is to hand-craft a prior via heuristics such as smoothness or sparseness. However, this approach is limited in the complexity of heuristics that we can conceive. Alternatively, we can \emph{learn} such a prior from data, and these can be encoded within the parameters and architecture (Section~\ref{sec:networkarch}) of a neural network. 

Beyond this, we might wish to adapt our prior based on specific conditions. In our street scene Lidar example, we might vary the expected surface shape of vehicles based on which kind of vehicle it is: a bicycle, a car, or a truck.
For neural fields, this is accomplished by \emph{conditioning} the neural field on a set of latent variables $\mathbf{z}$ that encode the properties of a specific field.
By varying the latent variables, we can then vary the neural field.
In this section, we discuss how to pose optimization problems that learn latent variables $\mathbf{z}$, how to infer $\mathbf{z}$ given a set of incomplete observations, and how to condition the neural field on $\mathbf{z}$ to decode them into different fields.



\subsection{Conditional Neural Fields}
\label{sec:cond_nf}

A conditional neural field lets us vary the field by varying a set of latent variables $\textbf{z}$. These latent variables could be samples from an arbitrary distribution, or semantic variables describing shape, type, size, color, etc., or come from an encoding of other data types such as audio data~\cite{guo2021adnerf}. 
For example, by conditioning our neural fields on semantic variables which describe cars, we would like to decode these to a field that represents the shape and appearance of the corresponding car.
Instance-specific information can then be encoded in the conditioning latent variable $\textbf{z}$, while shared information can be encoded in the neural field parameters. If this latent variables are defined on a semantic or smooth space, they can be interpolated or edited.
%

In the following subsections, we discuss techniques to learn in an unsupervised manner what the latent variables $\mathbf{z}$ are, how to infer them given (partial) observations $\mathcal{O}$, and how to condition neural fields upon $\mathbf{z}$ to be decoded into a corresponding field. 
$\mathbf{z}$ is typically a low-dimensional vector, and is often referred to as a \emph{latent code} or \emph{feature code}.
We discuss different encoding schemes (Section~\ref{sec:generalization_encoder}), both global and local conditioning (Section~\ref{sec:generalization_global_local}), and different mapping functions $\Psi$ (Section~\ref{sec:generalization_map_to_nf_param}).

\subsubsection{Encoding the Conditioning Variable $\mathbf{z}$}
\label{sec:generalization_encoder}

\begin{figure}[t]
	\centering
    \includegraphics{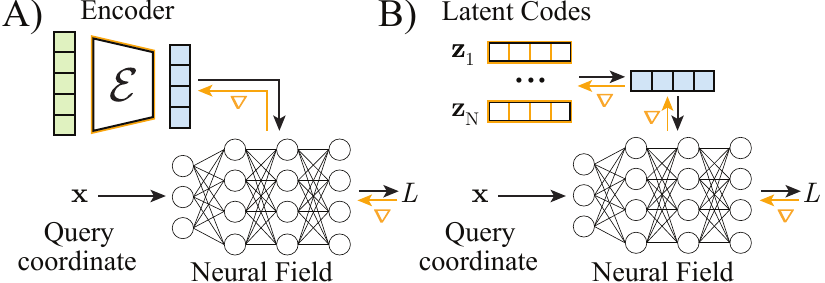}
	\caption{\textbf{Inference: Encoding vs.~Auto-decoding.} (A) In amortized or encoder-based inference, encoder $\mathcal{E}$ maps observations $\mathcal{O}$ to latent variable $\mathbf{z}$. Parameters of $\mathcal{E}$ are jointly optimized with the conditional neural field. (B) Auto-decoding lacks an encoder, and each neural field is represented by a \emph{separately-optimized} latent code $\mathbf{z}_i$, which are jointly optimized with the conditional neural field. Gradient flow displayed in orange.}
	\label{fig:encoder_vs_autodecoder}
	\vspace{-15pt}
\end{figure}

\parahead{Feed-forward Encoders/Amortized Inference}
For feed-forward encoder methods, the conditioning latent code $\mathbf{z} = \mathcal{E}(\mathcal{O})$ is generated via an encoder $\mathcal{E}$, typically a neural network (Figure~\ref{fig:encoder_vs_autodecoder}, left). 
The parameters in $\mathcal{E}$ can encode priors that can be pre-trained on data or auxiliary tasks.
The decoder is the neural field that is conditioned by the latent code.
This conditioning method is fast since inference requires only a single forward pass through the encoder and decoder. 
Examples of encoders for different data types include PointNet~\cite{qi2017pointnet} for point clouds, ResNet~\cite{he2016resnet} for 2D images, or VoxNet~\cite{maturana2015voxnet} for voxel grids.
%
%

\parahead{Auto-decoders}
%
%
%
Only the decoder is explicitly defined, and encoding happens through stochastic optimization. 
That is, the latent code $\textbf{z} = \argmin_\textbf{z} \mathcal{L}(\mathbf{z}, \Theta) $ is obtained by minimizing some loss function $\mathcal{L}$, which may be an expectation over a dataset (Figure~\ref{fig:encoder_vs_autodecoder}, right).
%
Every training observation is initialized with its own latent code $\mathbf{z}_i$.
To optimize a particular latent code $\mathbf{z}_i$, we map it to neural field parameters $\params$ (Section~\ref{sec:generalization_map_to_nf_param}), 
compute reconstruction loss, and back-propagate that loss to $\mathbf{z}_i$ to take a gradient descent step.
At training time, the per-observation latent codes $\mathbf{z}_i$ and the neural field $\Phi_{\Theta}$ are jointly optimized.
At test time, given a new observation $\hat{\mathcal{O}}$, we freeze parameters of $\Phi$ and optimize the latent code $\hat{\mathbf{z}}$ that minimizes reconstruction error. 
Thus, auto-decoders solve $\argmax_{\textbf{z}}\mathbb{P}(\textbf{z}|\mathcal{O})$ via first-order optimization; the optimization acts as encoder $\mathcal{E}$ to map observations $\mathcal{O}$ to latent variables $\mathbf{z}$. 

Given the inference-time optimization, auto-decoding is significantly slower than a feed forward encoder approach.
However, auto-decoding does not introduce additional parameters and does not make assumptions about observations $\mathcal{O}$.
For instance, a 2D CNN encoder assumes $\mathcal{O}$ to be on a 2D pixel grid, whereas an auto-decoder can ingest tuples of pixel coordinates and colors independently of their spatial arrangement.
This can lead to robustness in certain out-of-distribution scenarios. For instance, in 3D reconstruction from images when observing a camera pose not in the training set, a convolutional encoder is constrained by the 2D geometry of its kernels whereas an auto-decoder is not~\cite{sitzmann2019srn,li20213d}.
Due to these benefits, several works have adopted the auto-decoder approach~\cite{park2019deepsdf,gao2020portraitnerf,ramon2021h3dnet,yang2021deep,liu2021editing,jang2021codenerf,tretschk2021nrnerf,sitzmann2021lfns}.

\parahead{Hybrid Approaches}
These approaches initialize $\mathbf{z}$ with a forward pass of a parametric encoder $\mathcal{E}$, then continue to optimize $\mathbf{z}$ iteratively via auto-decoding~\cite{mu2021asdf}.

\subsubsection{Global and Local Conditioning}
\label{sec:generalization_global_local}

\begin{figure}[t]
	\centering
    \includegraphics{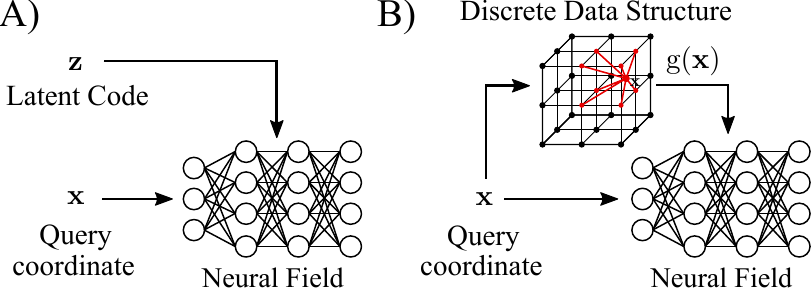}
	\caption{\textbf{Local vs.~Global Conditioning.} (A) In global conditioning, a latent code $\mathbf{z}$ defines the neural field across \emph{all} input coordinates $\coord$. (B) In local conditioning, a discrete data structure provides a \emph{coordinate-dependent} latent code $\mathbf{z}=g(\coord)$ such that the neural network is coordinate-dependent. Figure adapted from~\cite{peng2020convolutional}.}
	\label{fig:local_vs_global}
	\vspace{-15pt}
\end{figure}

In \emph{global conditioning}, a single latent code $\mathbf{z}\in \mathbb{R}^n$ determines the entire neural field $\Phi$ across coordinate values (Figure~\ref{fig:local_vs_global}, left). This is desirable in representation learning, when we want $\mathbf{z}$ to be \emph{compact}---to encode as much information as possible. 
However, certain signals are not amenable to the learning of a single, continuous space of latent variables $\mathbf{z}$. Specifically, signals that are a \emph{combination} of otherwise independent parts, such as the geometry of a room that is made up of separate objects. In such cases, representing the whole room via a global set of latent variables $\mathbf{z}$ means that these latent variables have to represent all the possible \emph{configurations} of the room, which grows exponentially with the number of objects.
As such, global latent codes succeed at modeling distributions with relatively few degrees of freedom, such as the shape and appearance of single objects or human bodies.

In \emph{local conditioning}, we introduce multiple latent codes $\mathbf{z}$. Each $\mathbf{z}$ has spatial extent over a local neighborhood in the space of coordinates $\coord$. Thus, latent codes are a function of $\coord$: $\textbf{z} = g(\coord)$ (Figure~\ref{fig:local_vs_global}, right).
Example $g$ are discrete data structures like 2D raster grids \cite{saito2019pifu,yu2021pixelnerf,trevithick2021grf}, 3D voxel grids \cite{peng2020convolutional,liu2020nsvf,jiang2020local,chibane2020ifnet,chabra2020deepls}, surface patches \cite{tretschk2020patchnets}, or orthographic 2D projections of 3D grids like floor maps~\cite{devries2021unconstrained,peng2020convolutional}.
As each latent code $\mathbf{z}$ only has to encode information about its \emph{local} neighborhood, it consequently does not need to store information about the configuration anymore. For instance, imagine that we split a room into small 3D cubes, and in each cube store a latent code that describes the geometry in that cube. These latent codes do not need to encode all possible configurations of the room anymore.

Similarly, for local conditioning, an encoder $\mathcal{E}$ only has to encode local properties into latent variables $\mathbf{z}$. 
The encoder can preserve the spatial extent of the observation $\mathcal{O}$, such as 2D or 3D CNNs to extract features from images or discretized volumes.
This can leverage encoder properties such as translation equivariance, granting better out-of-distribution generalization.
On the other hand, latent codes $\mathbf{z}$ do not capture global scene information, which provides fewer constraints and less high-level control.
%

\parahead{Hybrid Approaches}
Global and local conditioning can be combined. For instance, for human face images, to attempt to disentangle a property that is shared across instances (like hair color) from another that is region specific (like skin wrinkle).

\subsubsection{Mapping $\mathbf{z}$ to Neural Field Parameters $\params$}
\label{sec:generalization_map_to_nf_param}

Given a set of latent variables $\mathbf{z}$, we now wish to use them to parameterize the neural network that represents the corresponding field.
Different approaches have been proposed in prior work. Though on first sight, they seem incomparable, all of them follow the same principle of defining a function $\Psi$ that maps latent variables $\mathbf{z}$ to a subset of neural network parameters $\params=\Psi(\mathbf{z})$ that then parameterize the neural field $\Phi_\params$.
Different methods of conditioning differ in which parameters $\params$ are output by $\Psi$, as well as the form of $\Psi$ itself. These design choices may impact generalization ability, parameter count, and computational cost.

\parahead{Conditioning by Concatenation}
Neural fields can be conditioned on latent variables $\textbf{z}$ by directly concatenating coordinate inputs $\textbf{x}$ with $\textbf{z}$.
For instance, consider a neural field $\Phi: \mathbb{R}^2 \to \mathbb{R}^3$ that maps 2D pixel coordinates to RGB colors, and consider a vector of latent variables $\textbf{z} \in \mathbb{R}^n$.
Conditioning via concatenation would yield a conditional neural field $\Phi': \mathbb{R}^{2 + n} \to \mathbb{R}^3$, that takes as input concatenated coordinates $\coord$ and latent codes $\mathbf{z}$.

It is not obvious how conditioning via concatenation falls into the framework defined above, where all conditional neural fields are expressed as predicting a subset of the parameters $\params$ of a neural field $\Phi_\params$ via a function $\Psi$.
However, conditioning via concatenation is equivalent to defining an affine function $\Psi(\mathbf{z})=\mathbf{b}$ that maps latent codes $\mathbf{z}$ to the vector of biases $\mathbf{b}$ of the first layer of $\Phi$~\cite{sitzmann2020metasdf,dumoulin2018feature,mehta2021modulated}. In other words, in the case of conditioning via concatenation, the subset of parameters that is predicted by $\Psi$ is only the biases of the first layer of the neural network $\Phi$, and $\Psi$ is parameterized as a simple affine mapping.
%

\parahead{Hypernetworks}
Hypernetworks~\cite{ha2016hypernetworks} parameterize the function $\Psi$ as a neural network that takes the latent code $\mathbf{z}$ as input and outputs neural field parameters $\params$ via a forward pass~\cite{sitzmann2019srn,sitzmann2020siren,sitzmann2021lfns,nirkin2021hyperseg,chiang2021stylizing}.
We can view this as a general form of conditioning because every other form of conditioning may be obtained from a hypernetwork by outputting only subsets of parameters $\params$, by factorizing parameters of $\Phi$ via low-rank approximations or via additional scales and biases, or by varying $\Psi$ architectures such as using only a single linear layer.
For instance, conditioning via concatenation is a special case of a hypernetwork where $\Psi$ is an affine transform and only the biases of the first layer of $\Phi$ are predicted.
%
Full hypernetworks provide more complex embeddings of network weights than concatenation~\cite{galanti2020modularity}.

\parahead{FiLM and Other Conditioning}
Between conditioning via concatenation and full hypernetworks, one can condition an MLP by predicting feature-wise transformations (FiLM)~\cite{dumoulin2018feature,chan2021pigan,mehta2021modulated}.
To FiLM-condition a neural field $\Phi$, we use a network $\Psi$ to predict a per-layer (and potentially per-neuron) scale $\gamma$ and bias $\beta$ vector from latent variables $\latent$: $\Psi(\latent)=\{\gamma, \beta\}$.
The input $\mathbf{x}_i$ to the $i$-th layer $\Phi_i$ is transformed as $\Phi_i = \gamma_i(\latent) \odot \coord_i + \beta_i(\latent)$.
%
%
Yet another trade-off in the size of the subset of neural field parameters predicted is struck by predicting the factors of a low-rank decomposition of the weight matrices of the neural field $\Phi$~\cite{skorokhodov2021inrgan}.
%

\subsection{Gradient-based Meta-learning}
An alternative to the conditional neural field approach is gradient-based meta-learning~\cite{finn2017model}.
Here, all neural fields in our target distribution are viewed as specializations of an underlying \emph{meta-network} with parameters $\theta$~\cite{sitzmann2020metasdf,tancik2021learned}. 
Individual instances are obtained from fitting this meta-network to a set of observations $\mathcal{O}$, minimizing a reconstruction loss $\mathcal{L}$ in a small number of gradient descent steps with step size $\lambda$:
\begin{align}
    \Theta^{j+1} = \Theta^j - \lambda \nabla \sum_{\mathcal{O}} \mathcal{L}(\Phi(\mathcal{O}; \Theta_i^j)), \quad \Theta_i^0 = \theta.
\end{align}
Similar to the auto-decoder framework in conditional neural fields, the inference function $\mathcal{E}$ is implemented via an iterative optimization algorithm.
The meta-network can be seen as an initialization that is sufficiently close to all neural fields in the target distribution.

In a conditional neural field, the prior is expressed via the parameters of $\Psi$ that enforce that the parameters of $\Phi$ lie in a low-dimensional space as defined by latent variable $\mathbf{z}$. 
However, in gradient-based meta-learning, the prior is expressed by constraining the optimization to not move the neural field parameters $\params$ too far away from the parameters of the meta-network $\theta$.
Gradient-based meta-learning enables fast inference, as only a few gradient descent steps are required to obtain $\params$. As this does not assume a low-dimensional set of latent variables, in principle we retain the full expressivity of the neural field $\Phi$.


\section{Hybrid Representations}
\label{sec:hybrid_rep}
\begin{figure}[tbp]
  \centering
  \includegraphics[width=\linewidth]{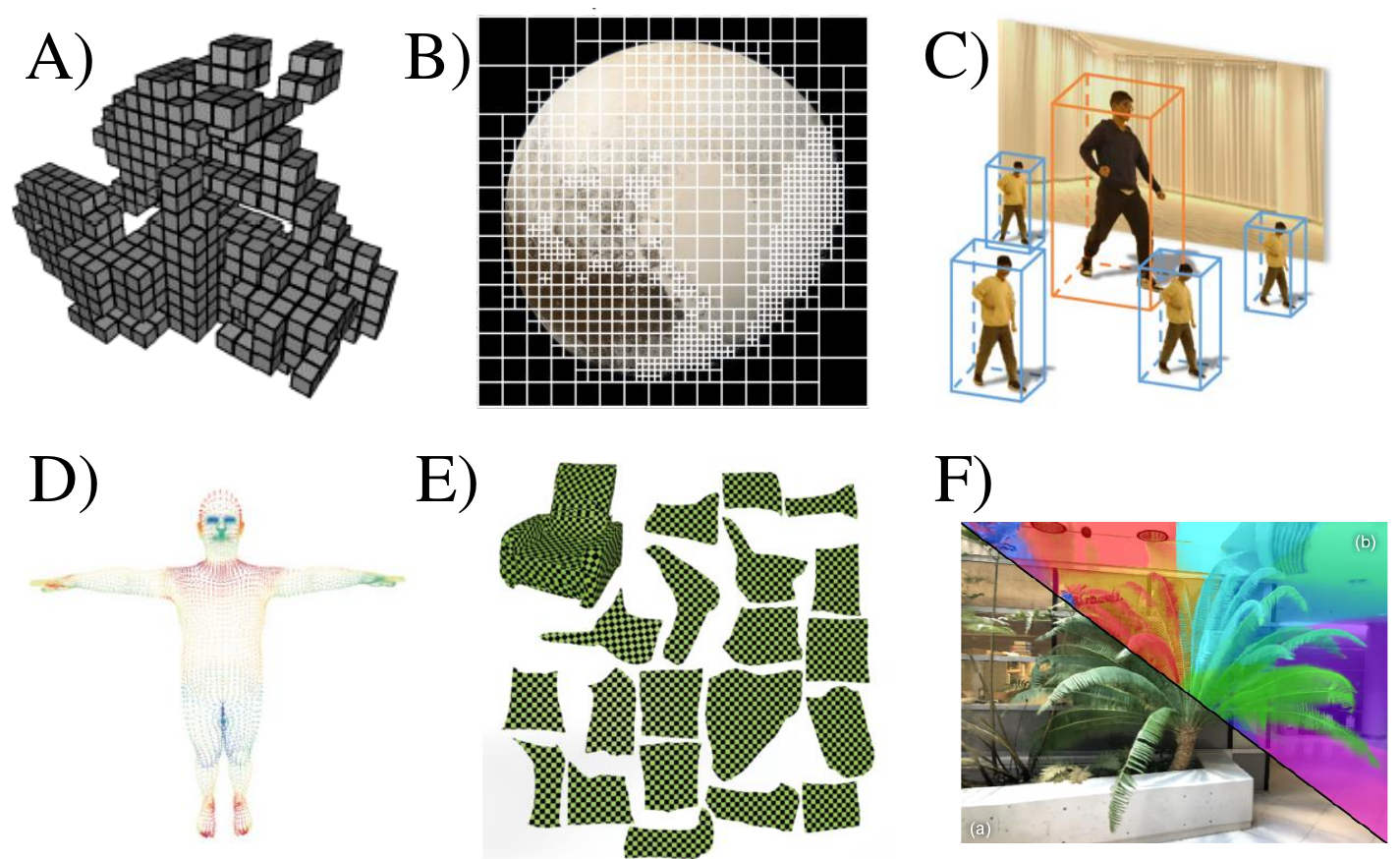}
  \caption{Examples of hybrid representations: (A) neural sparse voxel grid \cite{liu2020nsvf}, (B) multi-scale voxel grid with neural 2D image compression \cite{martel2021acorn}, (C) object bounding boxes with neural radiance fields \cite{zhang2021stnerf}, (D) mesh \cite{peng2021neuralbody} (E) atlas \cite{groueix2018atlasnet}, and (F) Voronoi decomposition with neural radiance fields \cite{rebain2021derf}. \jt{Need help with what d) and e) are combined with.}
  }
  \label{fig:hybrid_rep}
  \vspace{-15pt}
\end{figure}

The second category of items in the neural field toolbox are \emph{hybrid representations}. 
These combine neural fields with discrete data structures that decompose the space of input coordinates. This allows neural fields to scale up to large signals~\cite{ren2013global} (Figure~\ref{fig:hybrid_rep}).
Discrete data structures are used extensively in visual computing, including regular grids, adaptive grids, curves, point clouds, and meshes. 
Discrete structures have several important benefits: 1) They typically reduce computation. For instance, bounding volume hierarchies (BVHs)~\cite{rubin19803, meister2021survey} enable fast queries on hardware accelerators~\cite{parker2010optix,wald2014embree}. 2) They also allow for more efficient use of network capacity, since large MLP networks have diminishing returns in representation capacity~\cite{rebain2021derf}. 3) When representing geometry, discrete structures allow for empty space skipping, and so accelerate rendering. 4) Discrete structures are also suitable for simulation such as via finite element methods, and can help in manipulation and editing tasks (Section \ref{sec:editing}).

The general approach to spatial decomposition is to store some or all of the neural field parameters $\params_\field$ in a data structure $g$.
Given a coordinate $\coord$, we query $g$ to retrieve parameters $\params$ of the neural field. Two common approaches to how we map parameters $\params$ to data structure $g$ are network tiling and embedding.

\parahead{Network Tiling}
A collection of \emph{separate} (usually small) neural fields $\Phi$ are tiled across the input coordinate space, covering disjoint regions. The network architecture is shared, but their parameters are distinct for each disjoint region. Given a coordinate $\coord$, we simply look up the network parameters in the data structure $g$: 
\begin{equation}
    \quantity = \field(\coord, \params) = \field(\coord, g(\coord)).
\end{equation}

\parahead{Embedding} 
We store latent variables $\mathbf{z}$ in the data structure, as in Section~\ref{sec:generalization_global_local}.
Neural field parameters $\params$ become a function of the \emph{local} embedding $\mathbf{z} = g(\coord)$ via a mapping function $\Phi$:
\begin{equation}
    \quantity = \field(\coord, \params) = \field(\coord, \Psi(\mathbf{z})) = \field(\coord, \Psi(g(\coord)))
\end{equation}
%
Tiling is a special case of embedding where $\Psi$ is the identity function and $\mathbf{z}$ is the parameters $\params$. 
For design choice details of function $\Psi$, please refer to Section ~\ref{sec:generalization}.

\subsection{Defining $g$ and its Common Forms}

We define a discrete data structure $g$ as a vector field that maps coordinates to quantities using a sum of Dirac delta functions $\delta$:
\begin{equation}
    g(\coord) = \alpha_0\delta(\coord_0 - \coord) + \alpha_1\delta(\coord_1 - \coord) + ... + \alpha_n\delta(\coord_n - \coord),
\end{equation}
where coefficients $\alpha_i$ (scalar or vector) are the quantities stored at coordinates $\coord_i$.
For uniform voxels, the coordinates $\coord_i$ are distributed on a regular grid, whereas for a point cloud the coordinates $\coord_i$ are distributed arbitrarily.
For network tiling, $\alpha_i$ is an entire set of network parameters $\params_i$; for embedding, $\alpha_i$ is latent variable $\mathbf{z}_i$.

\parahead{Interpolation} The discrete data structure $g$ may use an \emph{interpolation scheme} to define $g$ outside the coordinates $\coord_i$ of the Dirac deltas, such as nearest neighbor, linear, or cubic interpolation.
If so, instead of Dirac deltas, function $\delta$ are basis functions of non-zero, compact, and local support.
For instance, voxel grids are regularly combined with nearest-neighbor interpolation, where $g(\coord)$ is defined as $\alpha_i$ at the $\coord_i$ closest to $\coord$.

\subsubsection{Regular Grids}
Regular grids, such as 2D pixels and 3D voxels, discretize the coordinate domain with regular intervals.
Their regularity makes them simple to index and apply standard signal processing techniques.
Although simple, grids suffer from poor memory scaling in high dimensions, and the Nyquist-Shannon theorem requires dense sampling for high-frequency signals.
To overcome this, grids can adaptive~\cite{martel2021acorn} or sparse~\cite{chabra2020deepls} to focus the capacity around higher frequency regions, and can be implemented with data structures like hierarchical trees~\cite{liu2020nsvf, takikawa2021nglod} and  textures~\cite{saito2019pifu, peng2020convolutional}.

\textit{Grid tiling} discretizes the coordinate domain with a grid and define each local region with smaller neural networks~\cite{ren2013global}. This can help learn larger scale signals~\cite{jagtap2020extended} , make inference faster~\cite{reiser2021kilonerf}, and can be suitable for parallel computing~\cite{shukla2021parallel}.
Tiling may increase overfitting given sparse training data~\cite{hu2021when}, and tile boundary artifacts are possible, though network parameter interpolation can be used to reduce boundary artifacts~\cite{dehesa2021gfnn, moseley2021fbpinns}.

\textit{Grids of embeddings} can similarly model larger-scale signals~\cite{chabra2020deepls}, enable the use of small neural networks~\cite{takikawa2021nglod}, and benefit from interpolation~\cite{liu2020nsvf}.
%
Grids of embeddings can also be generated from other neural fields~\cite{martel2021acorn}, generative models~\cite{ibing20213d}, images~\cite{saito2020pifuhd, trevithick2021grf, chan2021pigan}, or user input~\cite{hao2021gancraft}.
Please see Section \ref{sec:generativemodeling} for a more detailed discussion.

\subsubsection{Irregular Grids \jt{`irregular data structures'?}}

Irregular grids discretize the coordinate domain with a grid that does not follow a regular sampling pattern (and hence avoding the Nyquist-Shannon sampling limit). These can be morphed to adaptively increase capacity in complex data regions. They may declare connectivity between coordinates explicitly such as in meshes, or implicitly such as in Voronoi cells. They too may also be organized into hierarchies, such as within a BVH or scene graph~\cite{granskog2021neural, ost2021neural}.

\parahead{Point Clouds} Point clouds are a collection of sparse discrete coordinates. Each location can hold an embedding~\cite{tretschk2020patchnets} or a network~\cite{rebain2021derf}. Although sparse in its support, point clouds can volumetrically define regions through Voronoi cells via nearest neighbor interpolation. For continuous interpolation, we can use Voronoi cells with natural neighbor interpolation~\cite{sibson1981brief} or soft-Voronoi interpolation~\cite{williams2020voronoinet}.

\parahead{Object-centric Representations}\label{sec:object_centric} These are also a collection of points but where each has an orientation and a bounding box or volume~\cite{wang2019normalized}. Neural field parameters are stored at each point or at each vertex of the bounding volume, and can store embeddings~\cite{ost2021neural} or networks~\cite{guo2020osfs, zhang2021stnerf}.

\parahead{Mesh} Meshes are a common data structure in computer graphics with well understood properties and processing operations. For triangle meshes, embeddings can be stored on vertices \cite{peng2021neuralbody} and interpolated with barycentric interpolation. For complex polygons, mean-value coordinates~\cite{floater2003mean} and harmonic coordinates~\cite{joshi2007harmonic} are options.

\jt{2021-11-20 21:40 Up to here on my front-to-back edit.}

\section{Forward Maps}
\label{sec:forward_maps}


\begin{figure}[tbp]
  \centering
  \includegraphics[width=\linewidth]{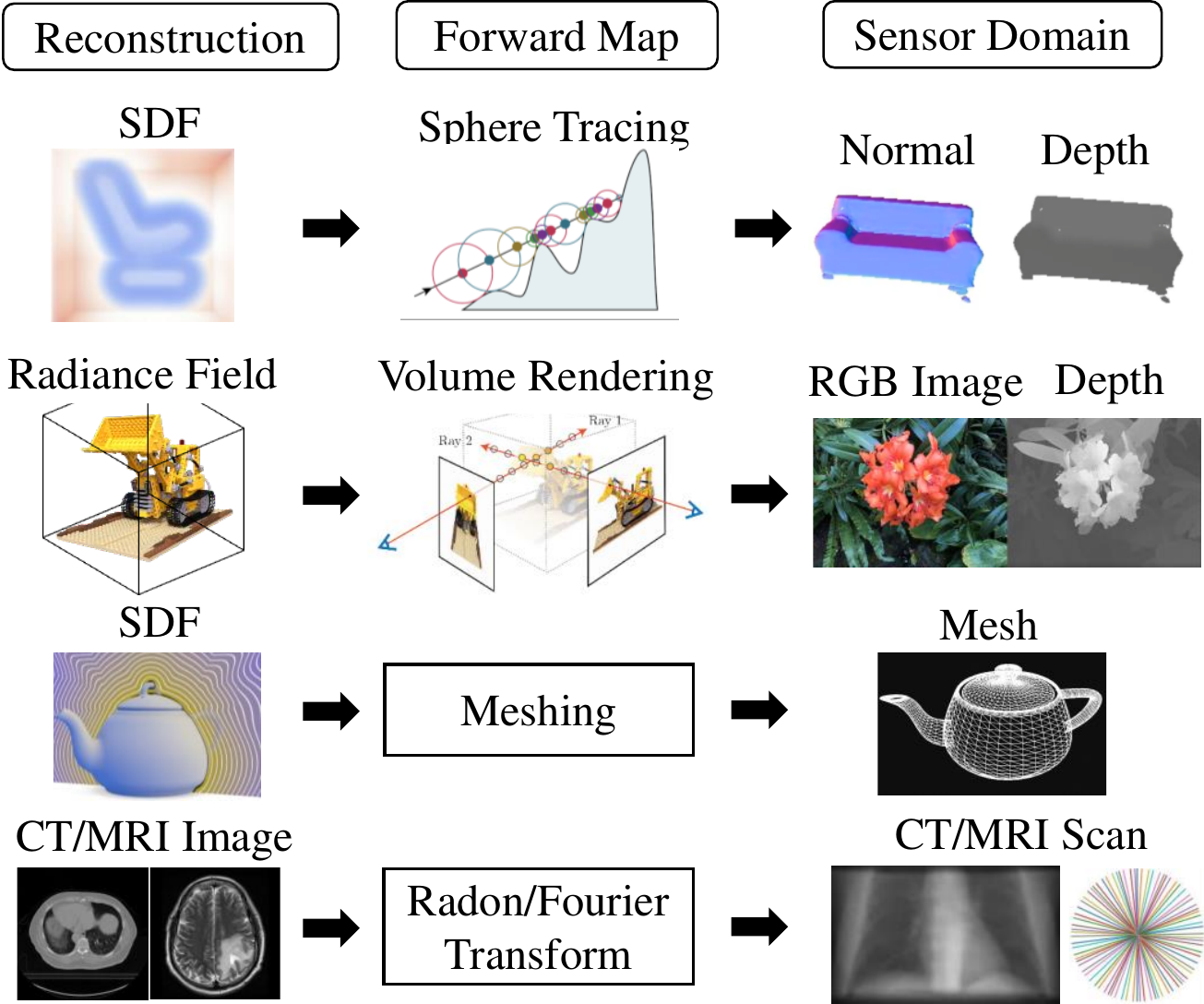}
  \caption{Forward maps relate reconstruction domains to sensor domains. Types of differentiable forward maps: a) sphere tracing \cite{liu2020dist}, b) volume rendering \cite{lombardi2019nv, mildenhall2020nerf}, c) meshing (e.g., marching cubes), d) Radon and Fourier Transformation \cite{shen2021nerp}, e) partial derivatives \cite{raissi2019physicsinformed}
}
  \label{fig:forward_maps}
  \vspace{-15pt}
\end{figure}

\begin{table*}[t]
\centering
\begin{adjustbox}{max width=\textwidth}
\begin{tabular}{l l l l l l}
\toprule
Problem                & Sensor                   & Sensor Domain              & Forward Module                  & Reconstruction Domain         & Literature                                                                              \\
\midrule
3D Reconstruction      & Digital Camera           & Image (2D discrete array)  & Rendering                       & Geometry, Appearance          & \cite{saito2019pifu, lombardi2019nv, mildenhall2020nerf, yariv2020idr} \\
Geodesy Estimation     & Accelerometer            & Gravitational acceleration & $F=m\vec{a}=G m_1 m_2/\vec{r}^2$ & Density (mass)                & \cite{izzo2021geodesynets}                                             \\
CT Reconstruction      & X-ray Detector           & Projection domain          & Radon Transform                 & Density/Intensity             & \cite{9606601, zang2021intratomo, shen2021nerp}                    \\
MRI Reconstruction     & RF Detector & Frequency domain           & Fourier Transform               & Density/Intensity             & \cite{shen2021nerp}                                                    \\
Audio Reconstruction   & Microphone               & Waveform                   & Fourier Transform               & Spectrogram                  & \cite{gao2021objectfolder}                                             \\
Synthetic Aperture Sonar & Microphone               & Waveform                   & Convolution (w/ PSF)            & Point Scattering Distribution & \cite{reed2021implicit}        \\                                       
\bottomrule
\end{tabular}
\end{adjustbox}
\caption{\label{tab:forward_module}Examples of forward maps. Detailed discussion on CT, MRI, Synthetic Aperture Sonar can be found in \ref{sec:beyond}. RF: Radio Frequency, CT: Computed Tomography, MRI: Magnetic Resonance Imaging, PSF: Point Spread Function.}
\vspace{-15pt}
\end{table*}


In many applications, the \emph{reconstruction domain} (how we represent the world) is different from the \emph{sensor domain} (how we observe the world).
Solving an \emph{inverse problem} recovers the reconstruction from observations obtained from sensors, i.e., finding the parameters $\Theta$ of a neural field $\Phi$ given observations from the sensor $\sensor$.

We represent the (unknown) reconstruction as a neural field $\Phi: \mathcal{X}\to\mathcal{Y}$ that maps world coordinates $\reccoord\in\mathcal{X}$ to quantities $\recfeats\in\mathcal{Y}$. 
A sensor observation is often also a field $\sensor: \mathcal{S}\to\mathcal{T}$ that maps sensor coordinates $\sensorcoord \in \mathcal{S}$ to measurements $\sensorfeats\in\mathcal{T}$. 

For instance, in 3D reconstruction from 2D images, the sensor domain contains camera images lying on a 2D raster grid, and the reconstruction domain could be some continuous 3D representation (e.g. SDF, radiance field). A 2D photograph from a camera is a field $\sensor : \mathbb{R}^2 \to \mathbb{R}^3$ representing irradiance. If we wish to model the reconstruction as a radiance field $\Phi : \mathbb{R}^5\to\mathbb{R}^3$ of anisotropic photon flux, we need a mapping between $\field$ and $\sensor$. We call such models \emph{forward maps}; Figure~\ref{fig:forward_maps} shows several examples.

A forward map is an operator 
$F : (\mathcal{X} \to \mathcal{Y}) \to (\mathcal{S} \to \mathcal{T})$,
that is a mapping of functions. The forward map may depend on additional parameters, and may be composed with downstream operators such as sampling or optimization.
We call a forward map \textit{parameter differentiable} if for $\textbf{y} = F(\field(\coord))$ we can calculate the derivative $\frac{\partial \textbf{y}}{\partial \theta}$, and \textit{input differentiable} if we can calculate $\frac{\partial \textbf{y}}{\partial \textbf{x}}$.
%

Given a parameter differentiable forward map, we can solve the following optimization problem to recover the neural field $\Phi$:
\begin{equation}
    \argmin_\Theta \int_{(\reccoord, \sensorcoord) \in (\mathcal{X}, \mathcal{S})} \| F(\field(\reccoord)) - \sensor(\sensorcoord) \|
\end{equation}

using differentiable programming and an algorithm such as stochastic gradient descent.

In the 2D image example, the forward map can be modeled with equations of radiative transfer, which integrate over a volumetric anisotropic vector field.
Neural Radiance Fields (NeRFs)~\cite{mildenhall2020nerf} are one example that recovers isotropic density and anisotropic radiance fields from 2D image measurements.
%
Even though many inverse problems are ill-posed with no guarantee that a solution exists or is unique, empirically forward maps help us to find good solutions in a variety of applications (Part II).
\subsection{Rendering}
\label{sec:differentiable_rendering}

We define a \emph{renderer} as a forward map which converts some neural field representation of 3D shape and appearance to an image. Renderers take as input camera intrinsic and extrinsic parameters, as well as the neural field $\Phi$ to generate an image. Extrinsic parameters define the translation and rotation of the camera, while intrinsic camera parameters define any other information for the image formation model such as field of view and lens distortion~\cite{hartley2003mvg}.
Renderers often utilize a \emph{raytracer} which takes as input a ray origin and a ray direction (\ie~a single pixel from the camera model), and returns some information about the neural field. 
The information may be geometric like surface normals and intersection depth, or they may also aggregate or return some arbitrary features.

\subsubsection{Ray-surface Intersection}
If a neural field represents a shape's surface, we can use several methods to obtain geometric information such as the surface normal and intersection point. 
For common shape neural fields representations such as occupancy and signed distance fields, ray-surface intersection amounts to a root finding algorithm. 
One such method is ray marching, which takes discrete steps on a ray to find the surface. 
Given the first interval, a method such as the Secant method can be used to refine the root within the interval.
This method will fail to converge at the right surface if the surface is thinner than the interval which discrete steps are taken. 
Taking jittered steps can alleviate this by stochastically varying the step length and using supersampling~\cite{dippe1985antialiasing}. 
Interval arithmetic can be used to guarantee convergence at the cost of iteration count~\cite{florez2007efficient}.
In some cases, a neural network may be employed within ray marching to predict the next step~\cite{sitzmann2019srn,neff2021donerf}.
If the surface is Lipschitz-bounded (as is the case for signed distance fields), then sphere tracing~\cite{hart1996sphere} can efficiently find intersections with guaranteed convergence.
Segment tracing~\cite{galin2020segment} can further speed up convergence at the cost of additional segment arithmetic computation at each step.

These methods are all differentiable, but naively back-propagating through the iterative algorithm is computationally expensive. Instead, the surface intersection point alone can be used to compute the gradient~\cite{niemeyer2020dvr, yariv2020idr}.
If a hybrid representation is used, we can exploit a bounding volume hierarchy to speed up raytracing. In some cases, the data structure can also be rasterized onto the image to reduce the number of rays.

\subsubsection{Surface Shading and Lighting}
\label{sec:shadingandlighting}
Once the ray-surface intersection point has been retrieved, we can calculate the radiance contribution \emph{from} the point towards the camera. 
This is done with a bidirectional scattering distribution function (BSDF)~\cite{cook1982reflectance, burley2012physically} which can be differentiable or be parameterized as a neural field~\cite{gargan1998approximating}.
To aggregate lighting contribution \emph{to} the point, the most phyiscally accurate method is to solve Kajiya's rendering equation~\cite{kajiya1986rendering} with a multi-bounce Monte Carlo algorithm~\cite{pharr2016physically}.
However, this can be especially computationally expensive combined with a neural field.

To overcome this issue, approximations of incident lighting (related to precomputed radiance transfer) can be used.
This include cubemaps~\cite{greene1986environment} or the use of spherical basis functions such as spherical harmonics~\cite{green2003spherical}.
Neural fields can also be employed to approximate incident lighting~\cite{wizadwongsa2021nex}.
In the case where the exact material properties or the lighting environment does not need to be modeled, a neural field with position and view direction as input can directly model the radiance towards the camera~\cite{mildenhall2020nerf}.

\subsubsection{Volume Rendering}
\label{vol_render}
The surface rendering equation cannot model scenes with inhomogeneous media such as clouds and fog. 
Even for scenes with opaque surfaces, inverse surface rendering has difficulty recovering high frequency or thin geometry such as hair and other meso-scale details because the gradients are only defined at surfaces.
Instead of Kajiya's rendering equation, volume rendering uses the \emph{volume rendering integral}~\cite{kajiya1984ray} based on the equations of radiative transfer~\cite{chandrasekhar2013radiative}, for which the integral can be numerically approximated using quadrature (in practice, stochastic ray marching~\cite{perlin1989hypertexture, mildenhall2020nerf}). 

In a differentiable setting, under the assumptions of exponential transmittance, integration can be performed with a simple cumulative sum of samples across a ray, making backpropagation efficient and dense with respect to coordinates~\cite{mildenhall2020nerf}. 
This also propagates gradients throughout space, making the optimization easier. 
Non-exponential formulations exist~\cite{liu2019learning,vicini2021non}; some are not physically accurate but work as an approximation.

One of the important factors in stochastic ray marching-based volume rendering is the number of samples. A higher sample count will mean more accurate models, but at the cost of computational cost and memory. 
Some approaches to mitigate this include using a coarse neural field model to importance sample~\cite{mildenhall2020nerf}, using a classifier network~\cite{neff2021donerf}, or using analytic anti-derivatives~\cite{lindell2021autoint}.


\subsubsection{Hybrid Volume and Surface Rendering}

Many differentiable renderers try to combine the strengths of a opaque surface-based renderer and a volume renderer.
Volume rendering is typically under-constrained due to the stochastic samples taken on intervals, resulting in noise near surfaces. 
Surface rendering only provides gradients at the object surface, which do not smoothly propagate across the spatial domain.

We can combine approaches by re-parameterizing the implicit surface as a density field with soft boundaries using a Laplace distribution~\cite{yariv2021volsdf}, a logistic density function~\cite{wang2021neus}, a Gaussian distribution, or a smoothed step function. We can also importance sample around the surface intersection point~\cite{oechsle2021unisurf}.

\posteg
{
\subsection{Differentiable Meshing / Differentiable Cube-Meshing}

\tt{is anyone writing this?}
}

\subsection{Physics-informed Neural Networks}
\label{sec:pinn}
Partial differential equations (PDEs) are also powerful forward modules that map network outputs to gradient space supervision. Most of the works thus far have relied on a completely data-driven paradigm where additional constraints can be imposed by the choice of representation or implicit biases and invariance from the network architecture. 
Another class of methods, known as physics-informed neural networks (PINNs), use \emph{learning bias}~\cite{karniadakis2021physics} which supervises boundary and initial values (from an incomplete simulation or observations) using a loss, and the rest of space by sampling or regularizing with equations of physics, typically partial differential equations (PDEs).

In visual computing, the PINN paradigm is often seen with signed distance functions~\cite{gropp2020igr}. One of the core properties of an SDF is that they satisfy the Eikonal equation:
\begin{equation}
    \|\nabla u(x)\| = \frac{1}{f(x)},\; x\in \mathbb{R}^n
\end{equation}
. The boundary values are the point cloud $\mathcal{X}$ which correspond to the 0-level set. In the special case when $f(x)=1$, $u(x)$ becomes an SDF. The Eikonal loss is therefore: $\mathcal{L} = \sum_{\coord \in \mathcal{X}} \|\|\nabla \Phi(x)\| - 1\|$.
%
%
%

PDEs are a natural description of the dynamics in the natural world. As such, a large collection of PDEs have been proposed in the discipline of physics, and naturally many of those PDEs have been used in conjunction with Neural Fields as PINNs. We refer readers to Karniadakis et. al for a much more comprehensive review of PINNs~\cite{karniadakis2021physics}.

\subsection{Identity Mapping Function}
%
In some applications, the sensor domain may be the same as the reconstruction domain. In these cases, the forward model is the identity mapping function. The task is simply overfitting a neural field to data. Some examples are: supervising a neural signed distance field directly by ground-truth values~\cite{park2019deepsdf}, using neural fields to "memorize" images, or audio signals~\cite{sitzmann2020siren}.
\section{Network Architecture}
\label{sec:networkarch}
The design choices that we make about the structure and components of a neural network have a significant impact on the quality of the  field it parameterizes.
The most obvious of these choices is the \emph{network structure} itself, such as how many layers are in the MLP and how many neurons are in each layer.
We assume that a reasonable structure with enough learning capacity has been chosen, and focus our discussion on other design decisions that provide inductive biases for effectively learning neural fields.

\subsection{Overcoming Spectral Bias}
\label{sec:pos_enc}
Real-world signals are complex, making it challenging for neural networks to achieve high fidelity.
Furthermore, neural networks are biased to fit functions with low spatial frequency~\cite{rahaman2019spectral, huh2021low}.
Several designs address this shortcoming.

\parahead{Positional Encoding} 
First proposed in the Natural Language Processing community, the coordinate input of the neural network may be transformed by a \emph{positional encoding} $\gamma:\mathbb{R}^n \to \mathbb{R}^m$, which is a set of scalar functions $\gamma_i:\mathbb{R}^n\to \mathbb{R}$ which maps a coordinate vector $\coord = \left[ x_1, x_2, ..., x_n \right]^T \in \mathbb{R}^n$ to a vector of embedded coordinates:
\begin{equation}
    \gamma(\coord) = \left[\gamma_1(\coord), \gamma_2(\coord), ..., \gamma_m(\coord) \right].
\end{equation}
Sinusoidal functions are widely used to equip neural fields with the ability to fit high-frequency signals. Proposed by~\cite{zhong2019reconstructing}, they can be formally written as $\gamma_i$ where:
\begin{align}
    &\gamma_{(2i)}(x) = sin(2^{i-1} \pi x), \\
    &\gamma_{(2i+1)}(x) = cos(2^{i-1} \pi x).
\end{align}
These sinusoidal embeddings, also known as Fourier feature mapping, were subsequently popularized for the task of novel view synthesis~\cite{mildenhall2020nerf,tancik2020ffn}.
%
In the context of neural tangent kernels~\cite{jacot2018neural}, sinusoidal positional encoding can be shown to induce a kernel with a spatial selectivity that increases with the frequency of the functions $\phi_i$.
Positional encodings can thus also be seen as controlling the interpolation properties of a neural field.

Wang et al.~\cite{wang2021spline} showed that the choice of the frequency of ${\gamma_i}$ biases the network to learn certain, band-width limited frequency content, where lower encoding frequencies result in blurry reconstruction, and higher encoding frequencies introduce salt-and-pepper artifacts.
For more stable optimization, one approach is to mask out high-frequency encoding terms at the beginning of the optimization, and progressively increase the high-frequency encoding weights in a coarse-to-fine manner \cite{lin2021barf,benbarka2021seeing}.
SAPE proposed a masking scheme~\cite{hertz2021sape} that also allowed the encoding of spatially varying weights. 
Finally, alternative positional encoding functions $\phi_i$ have also been proposed \cite{zheng2021rethinking,wang2021spline,muller2021realtime}.
Zheng et al.~\cite{zheng2021rethinking} conducted a comprehensive study of various positional encoding functions.

\parahead{Activation Functions}
An alternative approach to enable the fitting of high-frequency functions is to replace standard, monotonic nonlinearities with periodic nonlinearities, such as the periodic sine as in SIREN~\cite{sitzmann2020siren} enabling fitting of high-frequency content.
While a derivation of the properties of the neural tangent kernel is outstanding, some theoretical understanding can be gained by analyzing the relationships of the gradients of the output of the Neural Field with respect to neighboring coordinate inputs---for high frequencies of the sinusoidal activations, these gradients have been shown to be orthogonal, leading to an ability to perfectly fit values at any input coordinates without any interpolation whatsoever.
It has been pointed out that positional encoding with Fourier features is equivalent to periodic nonlinearities with one hidden layer as the first neural network layer~\cite{benbarka2021seeing}.

\posteg{\todo{Perlin noise / noise inputs?}}

\subsection{Integration and Derivatives}
\begin{figure}[h]
	\centering
    \includegraphics[width=\columnwidth]{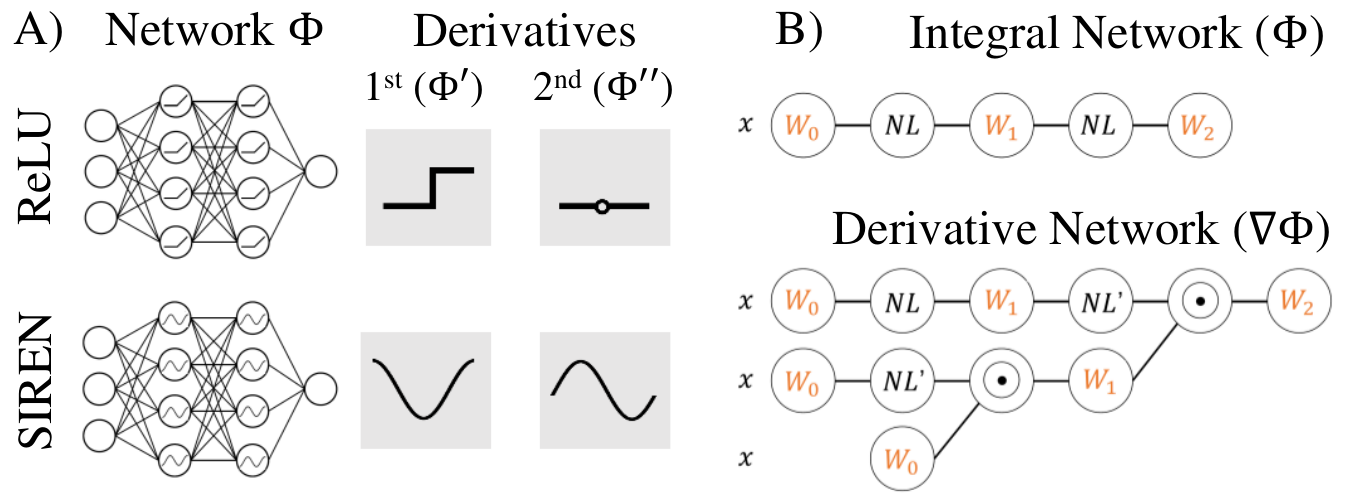}
	\caption{Derivatives and anti-derivatives of neural fields: A) Neural field derivatives depend on the activation function. ReLU yields (piece-wise) constant derivatives. Sinusoidal activations~\cite{sitzmann2020siren} yields (co-)sinusoidal derivatives. B) The computation graph of a neural network can be assembled via automatic differentiation, and shares parameters with its anti-derivative. This enables solving of optimization problems that require supervision of its integral~\cite{lindell2021autoint}. $NL$ and $NL'$ denote a non-linear activation function and its derivative, and $\odot$ denotes element-wise multiplication. Figures adapted from~\cite{sitzmann2020siren,lindell2021autoint}.
	\newline
	}
	\label{fig:integration_derivatives}
	\vspace{-15pt}
\end{figure}
\parahead{Derivatives}
A key benefit of neural fields is that they are highly flexible general function approximators whose derivatives $\nabla_\coord \Phi(\coord)$ are easily obtained via automatic differentiation~\cite{griewank1989automatic,baydin2018automatic}, making them attractive for tasks that require supervision of derivatives via PDEs. 
This, however, raises additional requirements in terms of the architecture of the MLP parameterizing the field. Specifically, the derivatives of the network must be nontrivial (\ie~nonzero) to the degree of the PDE (Figure~\ref{fig:integration_derivatives}, left).
For instance, to parameterize solutions of the wave equation --- a second-order PDE --- the second-order derivatives of the neural field must be nontrivial.
This places restrictions on the activation function used in the network.
For instance, ReLU nonlinearity has piece-wise constant first derivatives, and zero high-order derivatives, and so cannot generally parameterize solutions to the wave equation.
Other common nonlinearities, such as softplus, tanh, or sigmoid, may address this issue, but often need to be combined with positional encodings to parameterize high-frequency content of neural fields.
Alternatively, the periodic sine may be used as an activation function~\cite{sitzmann2020siren}.

\parahead{Integrals} In some cases, we are further interested in solving an inverse problem where we measure the \emph{derivative} quantity of a field, but are later interested in obtaining an expression for an integral over the neural field. 
In these cases, it is possible to sample the neural field and approximate the integral using numerical quadrature, as is often done for instance in volume rendering~\cite{mildenhall2020nerf}.
However, the number of samples needed to approximate the integral may be arbitrarily large for a given accuracy.

While automatic differentiation is ubiquitous in deep learning, automatic integration is not well-explored.
The seminal work AutoInt~\cite{lindell2021autoint} proposed to directly parameterize the \emph{antiderivative} of the field as a neural field $\Phi$.
We may then instantiate the computational graph of its \emph{gradient network} $\nabla_\coord \Phi(\coord)$, and fit the gradient network to the given derivative values.
At test time, a single integral can be calculated in one forward pass, as a consequence of the universal theorem of calculus (Figure~\ref{fig:integration_derivatives}, right).

\section{Manipulating Neural Fields}
\label{sec:editing}
While many data structures in visual computing have post-processing tools (e.g. smoothing a polygonal mesh), neural fields have limited tools for editing and manipulation, which significantly limits their use cases. Fortunately, this is an active area of research.

\subsection{Input Coordinate Remapping}
\label{sec:coordinate_remap}
The simplest approach to editing neural fields is to transform its spatial or temporal coordinate inputs.
A simple rigid translation of a neural field can be expressed as $g(x) = f(x + b)$.
We discuss other more sophisticated coordinate remapping methods below (Table~\ref{tab:warp_flow}).

\subsubsection{Spatial Transformation via Explicit Geometry}
An intuitive solution for controlling shape or appearance neural fields is to leverage explicit shape information.
For object modeling problems (Section \ref{sec:object_centric}), structural priors are often available in the form of bounding boxes and coarse explicit geometry.
Object bounding boxes offer a convenient, explicit handle to add and rigidly transform each object~\cite{zhang2021stnerf, ost2021neural, liu2020nsvf}. 



\begin{table*}[th!]
\centering
\begin{adjustbox}{max width=\textwidth}
\begin{tabular}{llll}
\toprule
\textbf{Task}                                      & \textbf{Updated Parameter}       & \textbf{Update Method}      & \textbf{References}                               \\
\midrule
Geometry editing                          & Latent code             & Backprop           & DualSDF \cite{hao2020dualsdf}            \\
Shape removal/color editing               & Network, latent code    & Backprop           & EditNeRF \cite{liu2021editing}           \\
\midrule
Texture style transfer                    & Appearance network      & Hypernetwork       & Chiang et al. \cite{chiang2021stylizing} \\
\midrule
Facial expression animation               & Latent code             & Interpolation/swap & Gafni et al. \cite{gafni2021nerface}, FLAME-in-NeRF \cite{athar2021flameinnerf}  \\
3D Geometry and/or texture editing              & Latent code             & Interpolation/swap & SRN \cite{sitzmann2019srn}, DVR \cite{niemeyer2020dvr}, IDR~\cite{yariv2020idr}             \\
Camera view, lighting, time interpolation & Latent code             & Interpolation/swap & XFields \cite{bemana2020xfields}         \\
Diffuse albedo                            & Latent code             & Interpolation/swap & NeRFactor \cite{zhang2021nerfactor}         \\
\bottomrule
\end{tabular}
\end{adjustbox}
\caption{\label{tab:edit_via_update}Neural fields can be edited by directly changing network parameters such as weights and latent features.
We list related work that performs editing through parameter editing.}
\end{table*}

For articulated objects, the kinematic chain offers explicit control over object geometry through its joint angles.
At inference time, animating object geometry is achieved by changing joint angle inputs~\cite{mu2021asdf}.
While this approach is effective for a small number of joints, it may introduce spurious correlations in case of long kinematic chains such as human body~\cite{loper2015smpl}.
To avoid this issue, we can represent target shapes as the composition of local neural fields~\cite{genova2020ldif,deng2020nasa}.
Each local field is defined with respect to the joint transformations on a template mesh, and composed by max operation~\cite{deng2020nasa} or weighted blending based on relative joint locations~\cite{su2021anerf,noguchi2021narf}.
The primary drawback of articulated compositional neural fields is that each local field is independently modeled, often producing artifacts around joints with unobserved joint transformations.
An alternative is to warp an observation space into a canonical space, where the reconstruction is defined, using the articulation of a template model~\cite{huang2020arch,saito2021scanimate}.
The warping is commonly represented as linear blend skinning (LBS), where the deformations of surface is defined as the weighted sum of joint transformations.
The blending weights are computed by a nearest-neighbor query on a template mesh~\cite{huang2020arch} or learning as neural skinning fields~\cite{saito2021scanimate, mihajlovic2021leap, chen2021snarf, tiwari2021neuralgif} (Table~\ref{tab:warp_flow}).

%


\subsubsection{Spatial Transformation via Neural Fields}
\label{subsubsec:st_warp}
\begin{table*}[ht!]
\centering
\begin{tabular}{p{4cm} p{5cm} p{2cm} p{5cm}}
\toprule
\textbf{Warp Representation}    & \textbf{Definition}                   & \textbf{Optimized Parameter}      & \textbf{Related Works} \\
\midrule
Position                        & $\mathbf{x}' = \field(\mathbf{x})$    & $\mathbf{x}'$                     & \cite{chen2021mocoflow,athar2021flameinnerf,nam2021neural} \\
Rigid Transformation            & $\mathbf{x}' = \mathbf{T} \mathbf{x}$ & $\mathbf{T}$                      & \cite{nam2021neural,park2021nerfies,park2021hypernerf} \\
Displacement                    & $\mathbf{x}' = \mathbf{x} + \Delta \mathbf{x}$    & $\Delta \mathbf{x}$   & ~References in caption \\
Velocity                        & $\mathbf{x}' = \mathbf{x} + \int_0^\tau{\mathbf{v} dt}$& $v$              & \cite{niemeyer2019occupancyflow,bemana2020xfields} \\
Discrete Cosine Transform & $\mathbf{x}' = (\frac{2}{T})^{1/2} \sum_{k=1}^K{\phi_{\mathbf{x},k} cos(\frac{\pi(2t+1)k}{2T})}$ & $\phi_{\mathbf{x},k}$     & \cite{wang2021dctnerf} \\
Vector Field                    & $\mathbf{x}' = \mathbf{x} + t \textbf{v}(\mathbf{x})$ & $\mathbf{v}(\mathbf{x})$  & \cite{atzmon2021augmenting} \\
\midrule
Linear Blend Skinning (LBS)     & $\mathbf{x}' = \sum_i {\mathbf{T}_i w_i \mathbf{x}}$  & $\mathbf{w}$              & \cite{chen2021snarf,saito2021scanimate,wang2021metaavatar} \\
Residual + LBS                  & $\mathbf{x}' = \sum_i {\mathbf{T}_i w_i \mathbf{x}} + \Delta \mathbf{x}$  & $\Delta \mathbf{x}$   & \cite{liu2021na,tiwari2021neuralgif} \\
\bottomrule
\end{tabular}
\caption{\label{tab:warp_flow} Spatial transformation of coordinates:
$x$ and $x'$ are source and target coordinates; $t$ is time; $\field: \mathbb{R}^m \rightarrow \mathbb{R}^3$ is neural deformation field; $\mathbf{T}\in \mathbb{R}^{4\times 4}$ is rigid transformation matrix; $\Delta \mathbf{x}\in \mathbb{R}^3$ is displacement; $\mathbf{v}\in \mathbb{R}^3$ is velocity;  
$\phi_{\mathbf{x},k}\in \mathbb{R}$ is the $k$-th DCT coefficient at point $\mathbf{x}$; $T$ is number of frames; $K\in \mathbb{Z}^+$ is number of DCT coefficients; 
$\mathbf{v}$ is flow vector; 
$w_i$ are skinning weights given by the human body model. 
References: \cite{li2021nsff,pumarola2021dnerf,yenamandra2021i3dmm,tretschk2021nrnerf,palafox2021npms,wang2021mirrornerf,zhang2021stnerf,gao2021dynamic}}
\vspace{-15pt}
\end{table*}
Unlike articulated objects with known transformations, modeling general dynamic scenes requires flexible representations that can handle arbitrary transformations.
As target geometry and appearance are often modeled with neural fields, a continuous transformation of the field itself is a natural choice.
%
Learning neural fields for spatial transformations is a highly under-constrained problem without 3D supervision~\cite{niemeyer2019occupancyflow}.
This motivates the use of regularization loss terms based on physical intuitions:

\begin{packed_itemize}
    \item \textbf{Smoothness}: The first derivative of warp fields w.r.t. spatiotemporal coordinates should be smooth, assuming no sudden movements. This is necessary to constrain unobserved regions~\cite{gao2021dynamic, park2021nerfies, tretschk2021nrnerf, du2021nerflow}.
    
    \item \textbf{Sparsity}: 3D scenes generally contain large empty space. Thus, enforcing sparsity of the predicted motion fields avoids sub-optimal local minima~\cite{gao2021dynamic, xian2021spacetime, tretschk2021nrnerf, du2021nerflow}.

    \item \textbf{Cycle Consistency}: If a representation provides both forward and backward warping (e.g.~\cite{gao2021dynamic, wang2021dctnerf, li2021nsff}, forward/inverse LBS~\cite{saito2021scanimate, mihajlovic2021leap}), we can employ cycle consistency as a loss function. Unlike other regularization terms, cycle consistency does not dampen the prediction as the global optimum should also satisfy this constraint. 
    
    \item \textbf{Auxiliary Image-space Loss}: Image-space information such as optical flows~\cite{li2021nsff, xian2021spacetime, du2021nerflow, gao2021dynamic, wang2021dctnerf} and depth maps~\cite{li2021nsff, gao2021dynamic, wang2021dctnerf} can also be used in auxiliary loss functions. 
    
\end{packed_itemize}

\subsubsection{Temporal Re-mapping}
Conditioning the neural field on temporal coordinates allows time editing such as speed-up/slow-down ($\field(a\cdot t)$, offset ($\field(t + b)$), and reversal ($\field(-t)$) \cite{zhang2021stnerf,li2021nsff}.


\subsection{Editing via Network Parameters}
Neural fields can also be edited by directly manipulating the latent features or the learned network's weights.
These methods are task-agnostic, and are applicable to editing geometry, texture, as well as other physical quantities beyond visual computing.
A subset of parameters can be selectively modified (\eg~geometry network, texture latent code).
Network weight editing and manipulation methods include (see also Table~\ref{tab:edit_via_update}):
\begin{itemize}
    \item \textbf{Latent Code Interpolation/Swapping}: For neural fields conditioned on latent codes, interpolation or sampling in the latent space can change properties of the representation~\cite{chen2019imnet}.
    \item \textbf{Latent Code/Network Parameters Fine-Tuning}: After pre-training, we can fine-tune parameters to fit new, edited observations at test time.
    Editing may leverage explicit representations, such as sketches on 2D images~\cite{liu2021editing}, or moving primitive shapes~\cite{hao2020dualsdf}. The neural field is coupled to the explicit supervision via differentiable forward maps.
    \item \textbf{Editing via Hypernetworks}: Hypernetworks can learn to map a new statistical distribution (\eg~a new texture style~\cite{chiang2021stylizing}) to a pre-trained neural field, by replacing its parameters.
\end{itemize}




\vspace{0.15in}

\chapter{Part II. Applications of Neural Fields}
In Part II of this report, we review neural field works based on their application domain.
The recent explosion of interest has seen neural fields used for a wide range of problem in visual computing such as 3D shape and appearance reconstruction, novel view synthesis, human modeling, and medical imaging.
Additionally, neural fields are increasingly being used in applications outside of visual computing including in physics and engineering.
For each application domain, we limit our discussion to only neural field work while providing pointers to more traditional methods as context.


\section{3D Scene Reconstruction}
\label{sec:3drecon}
The reconstruction of representations of 3D scenes from real-world measurements is critical for robotics and autonomous vehicles, and for graphics applications like games and visual effects.
Unsurprisingly, the earliest work that we are aware of that uses neural fields was for 3D shape representation~\cite{lim20043d}.
3D scenes have properties including geometry, appearance, materials, and lighting for both static and dynamic parts.
%
%
\emph{Reconstruction} is the solution to an inverse problem that maps available observations to a representation. 
For 3D, available observations are discrete (due to sensors), often sparse (few images), incomplete (partial point clouds), residing in a lower dimension (2D images), or lack vital topological information (point clouds). 
%
In this section, we discuss reconstructing, displaying, and editing 3D scenes while identifying which techniques from Part I that they use.
\subsection{Reconstruction of 3D Shape and Appearance}
\label{sec:recon_geo}
\parahead{Reconstructing 3D Scenes with 3D Supervision} 
A large amount of work is focused on reconstructing representations of geometry in the form of signed distance functions or occupancy functions, given 3D supervision. An example of 3D supervision are point clouds: A point cloud is a set $\mathcal{X} \subset \mathbb{R}^3$ of 3D points. Point clouds often originate from sampling an underlying surface (\eg~with LiDAR). In addition to spatial location, they may contain information such as the surface normal, or appearance information.

\begin{figure}
	\centering
    \includegraphics[width=0.9\linewidth]{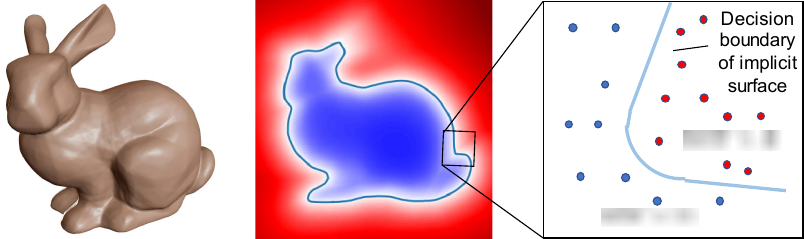}
	\caption{Neural fields for geometry commonly parameterize a surface as the level set of either an SDF~\cite{park2019deepsdf,yang2021nfgp}, or occupancy function ~\cite{mescheder2019occupancynetworks,chen2019imnet}. Figure adapted from DeepSDF~\cite{park2019deepsdf}.}
	\label{fig:deepsdf}
	\vspace{-15pt}
\end{figure}

%
Work on neural fields for geometry reconstruction often focuses on learned priors for reconstruction.
AtlasNet~\cite{groueix2018atlasnet} proposed to represent 3D shapes via predicting a set of 2D neural fields that lift 2D local patch coordinates to 3D (an ``atlas'' of patches).
Concurrently, FoldingNet~\cite{yang2018foldingnet} proposed a similar idea but is not continuous.
These patches were globally conditioned on a latent inferred from either a PointNet~\cite{qi2017pointnet} or ResNet~\cite{he2016resnet} encoder.
%
IM-Net~\cite{chen2019imnet} and Occupancy Networks~\cite{mescheder2019occupancynetworks} proposed to represent 3D geometry via an occupancy function.
Both proposed to generalize across ShapeNet~\cite{shapenet2015} via global conditioning-via-concatenation, where the latent was regressed from either a CNN (from an image) or a PointNet encoder (from a point cloud).
Concurrently, DeepSDF~\cite{park2019deepsdf} proposed to represent 3D surfaces via their signed distance function, similarly generalizing across ShapeNet objects with global conditioning-via-concatenation, but performing inference in the auto-decoder framework instead. Figure~\ref{fig:deepsdf} visualizes the SDF representation.
Even though these methods were proposed in conjunction with learning a shape prior, it is possible to overfit a DeepSDF or neural occupancy function on a single 3D shape.

AtlasNet was further extended~\cite{deprelle2019learning} by learning elementary structures in a data-driven manner, and inferring point correspondences across instances. 
Four concurrent papers proposed to leverage local features stored in voxel grids for local conditioning (Section~\ref{sec:generalization_global_local}) to improve on prior-based inference. 
IF-Net~\cite{chibane2020ifnet}, Chibane et al.~\cite{jiang2020local}, and Convolutional Occupancy Networks~\cite{peng2020convolutional} use 3D CNNs or PointNet operating to process voxelized point clouds into embedding grids to locally parameterize an occupancy network or signed distance function, where conditioning proceeds via concatenation.
In contrast, Chabra et al.~\cite{chabra2020deepls} similarly leverages local conditioning from a 3D voxel grid, but infers the latent codes in the voxel grid via auto-decoding.
CvxNet~\cite{deng2020cvxnet} proposes to represent a 3D shape as a composition of \emph{implicitly defined} convex polytopes.
LDIF~\cite{genova2020ldif} proposes to represent a 3D shape via a collection of local occupancy functions whose weighted sum represents the global geometry.
Littwin et al. \cite{littwin2019deep} infer an occupancy function via an image encoder and global conditioning via a hypernetwork. 
%

Many methods aim to improve the quality of learned shapes.
MetaSDF~\cite{sitzmann2020metasdf} was the first method to use gradient-based meta-learning (Section~\ref{sec:generalization_map_to_nf_param}) for neural fields, using it to infer 3D signed distance functions from point clouds.
Yang et al.~\cite{yang2021deep} show that optimizing not only the embeddings but also the network weights regularized to be close to the original weights can better resolve ambiguities in unobserved regions. 
IGR~\cite{gropp2020igr} uses a neural field to parameterize an SDF, but instead of requiring ground-truth signed distance values in a identity forward map, they leverage a partial differential equation forward map via the Eikonal equation to learn SDFs from a point cloud. 
Atzmon et al.~\cite{atzmon2019controlling} derive analytical gradients of the 3D position of points on a level set with respect to the neural field parameters.
SIREN~\cite{sitzmann2020siren} similarly learns an SDF with the Eikonal equation from oriented point clouds, but enables encoding of high-frequency detail by leveraging sinusoidal activations.  
Deep Medial Fields~\cite{rebain2021dmf} proposes to parameterize geometry via its medial field, the local thickness of a 3D shape, enabling faster level-set finding and other applications.
NGLoD~\cite{takikawa2021nglod} learns multiple level of details for SDFs with a compact hierarchical sparse octree of embeddings.
SAL~\cite{atzmon2020sal} uses a neural field to parameterize SDF, and show that the SDF can be learned by optimizing against the unsigned distance function of the point cloud given suitable initialization. SALD~\cite{atzmon2020sald} extends this with supervision of normals.
Neural Splines~\cite{williams2021neuralsplines} use as input oriented point clouds and use kernel regression of the neural field to optimize for normal alignment. 
PIFu~\cite{yu2021pixelnerf} performs prior-based reconstruction of geometry and appearance reconstruction by extracting features from images with a fully convolutional CNN, and, when querying a 3D point, projecting it on the image plane to use image features for local conditioning-via-concatenation.
Concurrently, DISN~\cite{xu2019disn} enhances single-view reconstruction by using a camera pose estimation that allows to project 3D coordinates onto the image plane and gather local CNN features. Combined with the global feature, the result is a more accurate SDF.

\parahead{Differentiable Rendering}
\label{sec:recon_appearance}
A major breakthrough in 3D reconstruction was the adoption of differentiable rendering (Section~\ref{sec:forward_maps}), which allowed reconstruction of 3D neural fields representing shape and/or appearance given only 2D images, instead of 3D supervision. 
This has significant implications since 3D data is often expensive to obtain, while 2D images are omni-present. 
A particularly important social implication is that non-experts can become 3D content creators, without the barrier of specialized hardware or capture rigs. 
SRNs~\cite{sitzmann2019srn} proposed using a differentiable sphere-tracing based renderer to reconstruct 3D geometry and appearance from only 2D observations. It leveraged global conditioning via a hypernetwork and inference via auto-decoding to enable reconstruction of a 3D neural field of geometry and appearance from only a single image for the first time.
Concurrently, Liu et al.~\cite{liu2019learning} use a CNN to predict an embedding to parameterize occupancy, and a form of differentiable volume rendering to produce a silhouette. 
Similar to DeepSDF, Occupancy Networks, and IM-Net, SRNs were designed to generalize, although they can be overfit given lots of 2D observations of a single 3D scene.

SDF-SRN~\cite{lin2020sdfsrn} enables learning an SRN from only a single observation per object at training time by enforcing a loss on the 2D projection of the 3D signed distance function. 
Kohli et al.~\cite{kohli2020semantic} leverages SRNs as a representation learning backbone for self-supervised semantic segmentation.
Liu et al.~\cite{liu2020dist} similarly reconstruct 3D geometry from 2D images with differentiable sphere-tracing.
DVR~\cite{niemeyer2020dvr} represents geometry via an occupancy function, and finds the zero-level set via ray-marching, subsequently querying a texture network for RGB color per ray.
Importantly, they derive an analytical gradient of the ray-marcher, which significantly reduces memory consumption at training time, but also requires ground-truth foreground-background masks.
They demonstrate the learning of appearance and shape priors across scenes via encoder-based conditioning to enable reconstruction from a single image, as well as overfitting on single scenes for high-quality, watertight 3D reconstruction.

\begin{figure}
    \centering
    \includegraphics[width=0.8\linewidth]{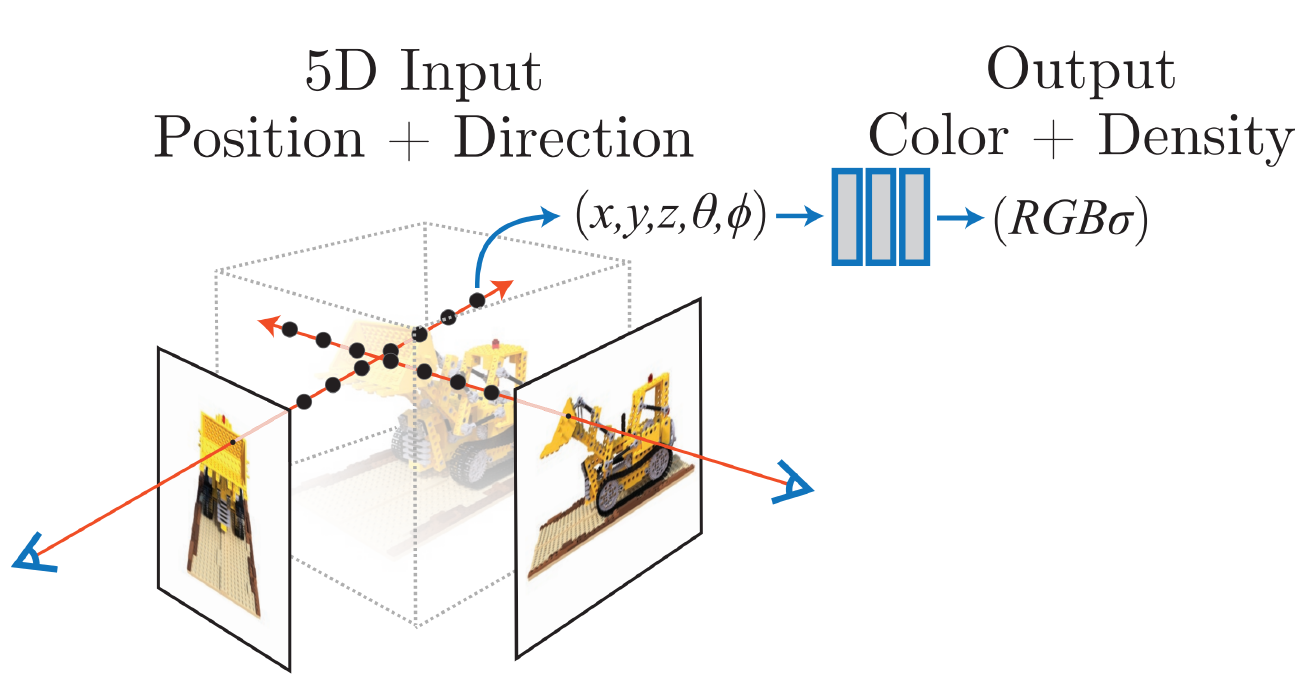}
	\caption{NeRF parameterizes a 3D scene as a 3D neural field mapping 3D coordinates to radiance and density, and can be rendered via volume rendering. Figure adapted from~\cite{mildenhall2020nerf}.}
	\label{fig:nerf}
	\vspace{-15pt}
\end{figure}
NeuralVolumes~\cite{lombardi2019nv} first proposed differentiable volume rendering, but leverages linearly interpolated voxel grids of color and density as a representation.
Though the paper mentions parameterization of color and density functions as neural fields, this was only a part of the ablation studies, and reportedly under-performed a 3D CNN decoder in terms of resolution.
cryoDRGN~\cite{zhong2020cryodrgn} implemented a differentiable volume renderer for the cryo-electron microscopy forward model to reconstruct protein structure from cryo-electron microscopy images, and proposed the positional encoding to allow fitting of high-frequency detail.
Here, the neural field is parameterized in the Fourier domain, and the forward model is implemented via the Fourier slice theorem.
NeRF~\cite{mildenhall2020nerf,tancik2020ffn} combined volume rendering with a single ReLU MLP, parameterizing a monolithic neural field, and added positional encodings (Figure~\ref{fig:nerf}). By fitting a single neural field to a large number of images of a single 3D scene, this achieved photo-realistic novel view synthesis from only 2D images of arbitrary scenes for the first time.
The visual quality and elegant formulation of NeRF has since inspired a large collection of follow-up work.
Nerf++~\cite{zhang2020nerf++} improves representation of unbounded 3D scenes via an inverted-sphere background parameterization.
Reizenstein et al.~\cite{reizenstein2021co3d} and Arandjelović et al. ~\cite{arandjelovic2021nerf} propose attention-based accumulation of  samples along a ray.
DoNeRF~\cite{neff2021donerf} proposes to jointly train a NeRF and a ray depth estimator for fewer samples and faster rendering at test time.
\cite{garbin2021fastnerf,yu2021plenoctrees,reiser2021kilonerf,hedman2021snerg,liu2020nsvf} propose various variants of local conditioning (without generalization, for overfitting a single scene) to speed up the rendering of NeRFs.
Mip-NeRF~\cite{barron2021mipnerf} proposes to control the frequency of the positional encoding for multi-scale resolution control. 
NeRF\textminus\textminus, BARF and iNeRF~\cite{wang2021nerf--,lin2021barf,yen-chen2021inerf} propose to back-propagate into camera parameters to enable camera pose estimation given a reasonable initialization.

PixelNeRF~\cite{yu2021pixelnerf} and GRF~\cite{trevithick2021grf} perform prior-based reconstruction by extracting features from images with a fully convolutional CNN, and, when querying a 3D point, projecting it on the image plane to use image features for local conditioning-via-concatenation, similar to PIFu and DISN~\cite{saito2019pifu,xu2019disn}.
With more context views, a similar approach can be used for multi-view-stereo-like 3D reconstruction~\cite{chen2021mvsnerf,chibane2021srf}.
NeRF-VAE embeds a globally-conditioned NeRF with encoder-based inference in a VAE-like framework~\cite{kosiorek2021nerfvae}.
While volume rendering has better convergence properties than surface rendering and enables photorealistic novel view synthesis, the quality of the reconstructed geometry is worse, due to the lack of an implicit, watertight surface representation. 
IDR~\cite{yariv2020idr} leverages an SDF parameterization of geometry, a sphere-tracing based surface-renderer, and positional encodings to enable high-quality geometry reconstruction. 
Kellnhofer et al.~\cite{kellnhofer2021nlr} distill an IDR model into a Lumigraph after rendering to enable fast novel view synthesis at test time.
Concurrently, UNISURF~\cite{oechsle2021unisurf}, NeuS~\cite{wang2021neus}, 
VolSDF~\cite{yariv2021volsdf} propose to relate the occupancy function of a volume to its volume density, thereby combining volume rendering and surface rendering, leading to improved rendering results and better geometry reconstruction. 
Ray marching requires many samples along a ray to faithfully render complex 3D scenes.
Even for relatively simple scenes, rendering requires hundreds or even thousands of evaluations of the neural scene representation per ray.

\begin{figure}[t]
	\centering
    \includegraphics[width=\linewidth]{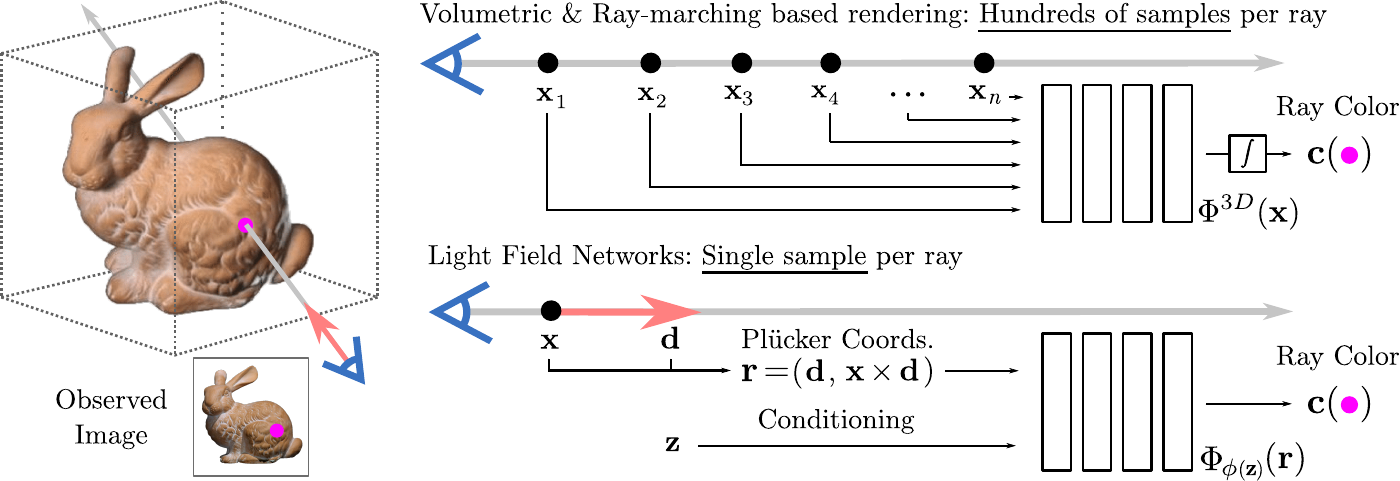}
	\caption{Instead of encoding a 3D scene with a 3D neural field that requires hundreds of samples along a ray to render, appearance \& geometry may be encoded as a neural light field directly mapping an oriented ray to a color, enabling rendering with a single sample per ray, but requiring a multi-view consistency prior. Figure adapted from ~\cite{sitzmann2021lfns}.}
	\label{fig:light_fields}
	\vspace{-15pt}
\end{figure}

To overcome this limitation, we can parameterize the \emph{light field} of a scene, which maps every ray to a radiance value.
This enables real-time novel view synthesis with a single neural field sample per ray, and geometry extraction from the neural light field without ray-marching.
Figure~\ref{fig:light_fields} displays the difference between ray-marching and rendering in light fields.

To prevent overfitting to the input views and allow view synthesis, we can learn multi-view consistency via global conditioning, hypernetworks, and inference via auto-decoding~\cite{sitzmann2021lfns}, or through ray embedding spaces~\cite{attal2022learning}.
Alternatively, \emph{densely} sampled rays of a single scene allow easier overfitting for view synthesis.
NeuLF~\cite{liu2021neulf} parameterizes forward-facing scenes via light fields, overfitting on single scenes, and addresses multi-view consistency via enforcing similarity of randomly sampled views with the context views in the Fourier domain.
Similar to light fields, NeX~\cite{wizadwongsa2021nex} parameterizes a set of multi-plane images as a 2D neural field, where the network inputs are pixel locations. Rendering is computationally efficient without ray marching.

\subsection{Reconstruction of Scene Material and Lighting}
\label{sec:mat_light}
The goal of \emph{material reconstruction} is to estimate the material properties of a surface or participating media from sparse measurements such as images.
For opaque surfaces, this may be the parameters of a bidirection scattering distribution function (BSDF). 
For participating media, this may be phase functions. This is a difficult problem, because materials are diverse and create
complex light transport effects. Even just for BSDFs, there is a whole taxonomy~\cite{mcguire2020taxonomy} of characteristic properties. 
In addition, to accurately estimating materials, we must also perform \emph{lighting reconstruction} or have prior knowledge of the lighting.

\begin{figure}
    \centering
    \includegraphics[width=1.0\linewidth]{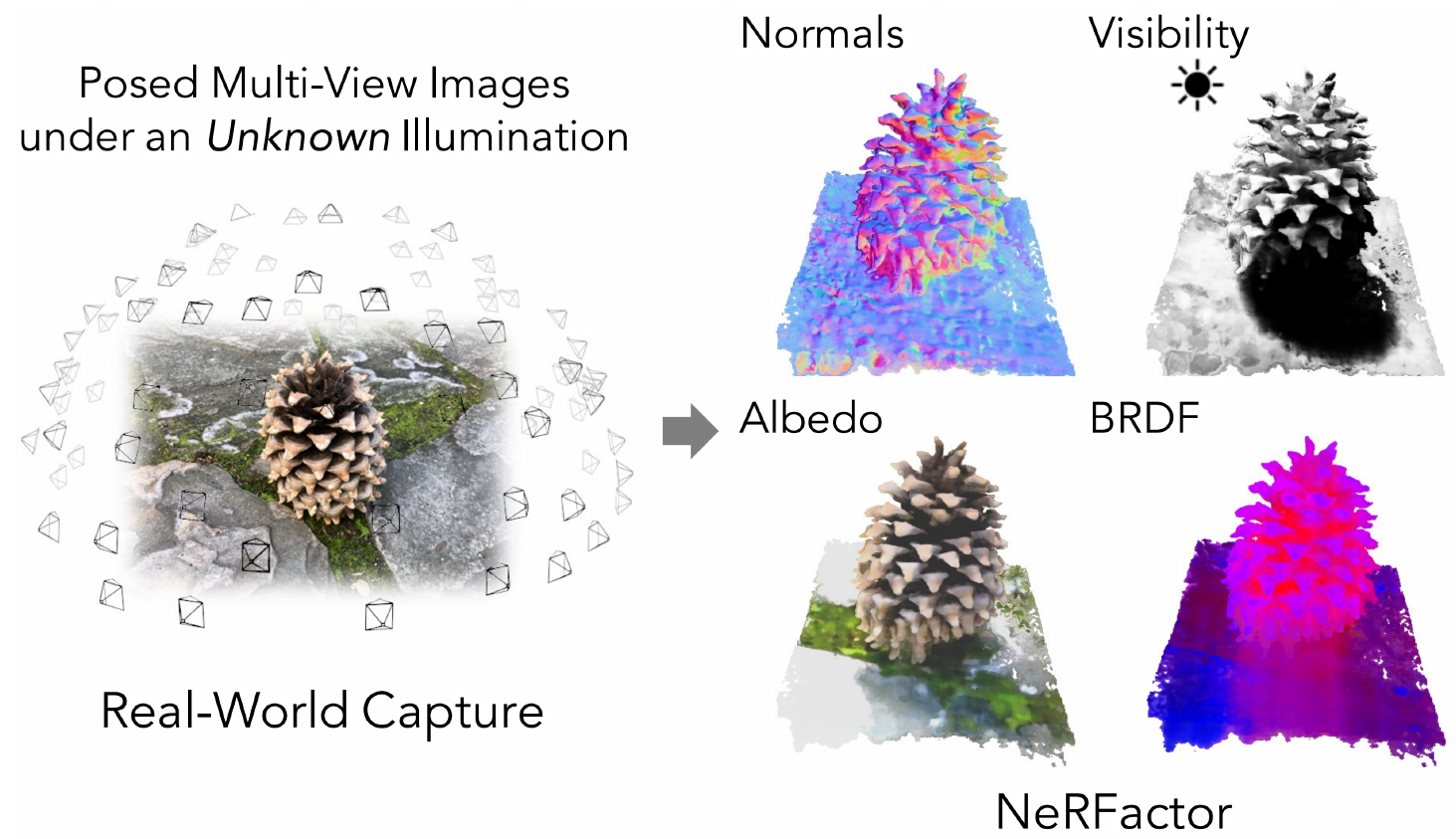}
	\caption{NeRFactor proposes neural fields for light visibility, BRDF, albedo, and surface normals, enabling re-rendering of 3D scenes under varying illumination and lighting following reconstruction. Figure adapted from ~\cite{zhang2021nerfactor}.}
	\label{fig:relighting}
	\vspace{-10pt}
\end{figure}

The forward modeling of light transport involves surface rendering~\cite{kajiya1986rendering} 
and volume rendering~\cite{kajiya1984ray} for participating media, which both equations having recursive integrals with no closed form solution for forward modeling. 
Approaches to inversely solve these equations differ in the degree of approximation they make. Because neural networks
are general function approximators, they can be useful for estimating arbitrary functions and integrals that are hard to solve.
Some methods reconstruct \emph{appearance} as approximate incident radiance, and other methods attempt to separate appearance into \emph{materials} via explicit scattering distribution functions and \emph{lighting},
enabling applications such as relighting.

\parahead{Neural Fields of Material Parameters} Many papers use a neural field to parameterize the parameter space of existing material models. Neural Reflectance Fields~\cite{bi2020neural} can reconstruct both the SVBRDF and geometry by assuming a known point light source, and use a neural field which parameterizes density, normals, and parameters of a microfacet BRDF~\cite{walter2007microfacet} with a volume rendering forward map with one bounce direct illumination.
NeRV~\cite{srinivasan2021nerv} extends this to handle more varied lighting setups with an environment map and a one bounce indirect illumination forward map, along with an additional neural field which parameterizes visibility, but assume the environment map is known a priori. 
NeRD~\cite{boss2021nerd} also use a neural field which paramterizes density, normals, and the parameters of a 
Disney BRDF model~\cite{cook1982reflectance, burley2012physically}, but remove any assumptions on the lighting by using spherical gaussians to represent lighting of which the parameters are directly optimized. They do not model visibility or indirect illumination.
PhySG~\cite{zhang2021physg} also directly optimize spherical gaussians, but uses a neural field to parameterize an SDF and a Ward BRDF~\cite{ward1992measuring} along with a hard-surface-based forward map to improve surface reconstruction accuracy. NeRFactor~\cite{zhang2021nerfactor} learns an embedding-based neural field of BRDF parameters as a prior on the material parameter space (Figure~\ref{fig:relighting}).

\posteg{

\parahead{Neural Materials} Other papers opt to directly model the material (BSDF, phase function) with a neural field. \cite{gargan1998approximating}, \tt{TODO: Probably many others}

\tt{Neural Radiosity and Neural Radiance Cache and the zhu RGBD transparent paper do not fit here but should go somewhere in the paper, probably post deadline. also need to fit neural raytracing in somewhere}

}

\begin{figure}
    \centering
    \includegraphics[width=0.75\linewidth]{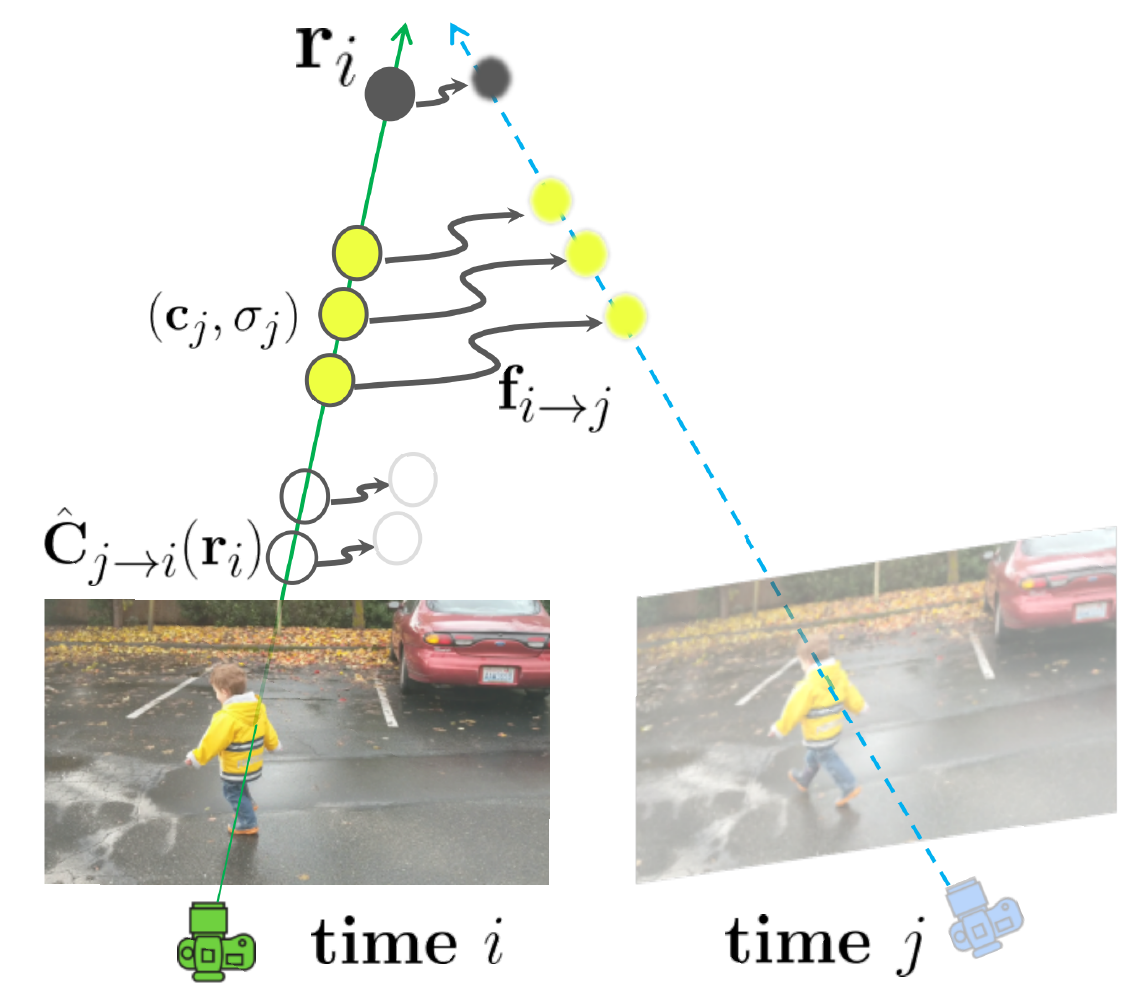}
	\caption{NSFF proposes novel time and view synthesis for dynamic scenes by reconstructing the 3D scene via a time dependent neural field. Figure adapted from \cite{li2021nsff}.}
	\label{fig:dynamic_scenes}
	\vspace{-15pt}
\end{figure}

\subsection{Dynamic Reconstruction}
\label{sec:recon_dynamic}

In addition to novel-view synthesis, dynamic scene reconstruction also allows measurement, mixed reality, and visual effects.
The challenges of modeling dynamic scenes are that the input data is even sparser in spacetime and often no 4D ground-truth data are available. With careful design choices, regularization loss terms, and the strong inductive bias of neural fields, several works have proposed solutions to this inverse problem. Figure~\ref{fig:dynamic_scenes} shows one of such approaches.
We review the existing approaches based on warp representation and how to embed temporal information.
For dynamic reconstruction of humans, please refer to Section~\ref{sec:digital_humans}.

\parahead{Embedding} 
Modeling temporally changing objects or scenes requires additional embedding that encodes frame information~\cite{rempe2020caspr}.
Occupancy Flow~\cite{niemeyer2019occupancyflow} models a dynamic component as a neural field conditioned by normalized temporal coordinates.
Similarly dynamic scene reconstruction methods based on differentiable rendering often condition neural fields with temporal coordinates~\cite{xian2021spacetime,li2021nsff,pumarola2021dnerf,du2021nerflow}.
Another approach is to jointly optimize per-frame latent code~\cite{park2021nerfies,li2021dynerf,tretschk2021nrnerf,park2021hypernerf,attal2021torf} as embedding.
While embedding based on temporal coordinates automatically incorporates temporal coherency as inductive bias, per-frame latent codes can enable the captures of more scene details.

\parahead{Warp Representation} 
To model dynamic scene from limited input data, we can split the problem into modeling a scene in the canonical space and warping it into each time frame. In Section \ref{subsubsec:st_warp}, 
we provide common warp representations and techniques to regularize the warp fields. While most approaches use the learned warping function as the final scene representation, several works use warp fields only for regularizing the predicted radiance fields~\cite{li2021nsff, gao2021dynamic}. Unlike other dynamic reconstruction approaches, where non-rigid complex warping is modeled by neural fields, STAR~\cite{yuan2021star} models motion as a global rigid transformation to primarily focus on tracking of a foreground object.

\section{Digital Humans}
\label{sec:digital_humans}
Human shape and appearance has received special attention in computer vision and graphics in the last decade, and is one of the most popular application areas of neural fields.
The adoption of neural fields has produced unprecedentedly high-quality synthesis and reconstruction of human faces, bodies, and hands. The state-of-the-art continues to evolve quickly.

\parahead{Face Modeling}
The data-driven parametric morphable model was introduced by Blanz and Vetter~\cite{blanz1999morphable}. However, these explicit representations lack realism and impose topological limitations making it difficult to model hair, teeth, etc. 
%
%
The expressiveness of fields, such as SDF and radiance fields, has made them an excellent candidate to address these limitations. 

Given a collection of high-quality 3D scans, i3DMM~\cite{yenamandra2021i3dmm} used coordinate-based neural networks that predict SDF and color to develop a 3D morphable head model with hair. Contrary to mesh-based representation, they learn an implicit reference shape as well as a deformation for each shape instance enabling editing abilities by disentangling color, identity, facial expressions, and hairstyle.
Ramon et al.~\cite{ramon2021h3dnet} use coordinate-based neural networks building upon IDR~\cite{yariv2020idr} to reconstruct a 3D head. To reduce the required amount of input images, a pre-trained DeepSDF-based latent space is used to regularize the test-time optimization.  
%
%
SIDER~\cite{chatziagapi2021sider} incorporates coarse geometric guidance via a fitted FLAME model~\cite{li2017learning} followed by implicit differentiation, enabling single-view optimization of facial geometry using SDF parameterized by a coordinate-based network.

While the aforementioned approaches focus on geometric accuracy assuming hard surfaces, NeRF have recently been adopted for photo-realistic view synthesis of human heads.
Nerfies~\cite{park2021nerfies} utilizes a casual video footage captured with a moving hand-held camera to learn radiance fields and deformation fields of a human head.
HyperNeRF~\cite{park2021hypernerf} extends Nerfies by incorporating auxiliary hyper dimensions to handle large topological change.

Another line of works enable the semantic control of radiance fields by conditioning on head pose and facial expression parameters obtained from a 3D morphable face model~\cite{gafni2021nerface, athar2021flameinnerf}. 
Supervised by multi-view video sequences, Wang et al.~\cite{wang2021hybridnerf} use a variational formulation to encode dynamic properties in spatially varying animation codes stored in voxels. 
%
Notably, the above methods require per-subject training, and reconstruction from limited observations remains challenging. 
Several works show promise in few-shot, generalizable reconstruction by exploiting test-time fine-tuning~\cite{gao2020portraitnerf} or pixel-aligned image features~\cite{raj2021pva}.


\parahead{Body and Hand Modeling}
Neural fields have demonstrated efficacy in 3D reconstruction of clothed humans from image inputs~\cite{saito2019pifu, saito2020pifuhd, he2020geopifu, huang2020arch, zheng2021pamir} or point clouds~\cite{chibane2020ifnet, bhatnagar2020ipnet}. Due to substantial variations in shape and appearance of clothed human bodies, a global latent embedding does not lead to plausible reconstruction. PIFu~\cite{saito2019pifu} addresses this by introducing a coordinate-base neural network conditioned on pixel-aligned local embeddings. Its followup works also employ the framework of PIFu for human digitization tasks from RGB inputs \cite{li2020monoport, he2020geopifu, suo2021neuralhumanfvv, yang2021s3, hong2021stereopifu} or RGB-D inputs \cite{li2020pifusion, yu2021function4d}.
To further improve the fidelity of reconstruction, multi-level feature representations are shown effective~\cite{saito2020pifuhd, chibane2020ifnet}. Yang et al.~\cite{yang2021s3} jointly predict skinning fields and skeletal joints for animating the reconstructed avatars. Also, several works explicitly leverage a parametric body model such as SMPL~\cite{loper2015smpl} to improve robustness under different poses~\cite{zheng2021pamir} and to enable an animation-ready avatar reconstruction from a single image~\cite{huang2020arch, he2021arch++}.

Human bodies are dynamic as they are both articulated and deformable. Several works show that providing the structure of human bodies significantly improves the learning of radiance fields~\cite{peng2021neuralbody, peng2021animatable, chen2021animatable, liu2021na}. 
Given a fitted body model to images, Neural Body~\cite{peng2021neuralbody} diffuses the latent embeddings on the body via sparse convolution, which are inputs to the neural field. Neural Actor~\cite{liu2021na} projects queried 3D points onto the closest point on the fitted body mesh, and retrieves latent embeddings on the UV texture map. A-NeRF~\cite{su2021anerf} explicitly incorporates joint articulations, and learns radiance fields in the normalized coordinate space. In addition, as shown for face modeling, pixel-aligned local embeddings are highly effective to support novel-view rendering of unseen subjects~\cite{shao2021doublefield, kwon2021neuralhumanperformer}.

Template-mesh registration is another important task in body modeling. IP-Net~\cite{bhatnagar2020ipnet} proposes to jointly infer inner body and clothing occupancy fields given input scans to aid registration.
3D-CODED~\cite{groueix20183dcoded} proposes Shape Deformation Networks to fit a template to a target shape and infer correspondences. 
Halimi et al.~\cite{halimi2020greater} show template-based shape completion by conditioning the deformation field on the part and whole encoding.
LoopReg~\cite{bhatnagar2020loopreg} presents a self-supervised learning of dense correspondence fields on the predicted implicit surface to a template human mesh, which improves the robustness of surface registration. 
A continuous local shape descriptor for dense correspondence is proposed by Yang et al.~\cite{yang2020continuous}.
Wang et al.~\cite{wang2021locally} model occupancy fields of input scans in an un-posed canonical space by predicting piece-wise transformation fields (PFT). Since shapes are modeled in the canonical space, we can perform template registration without self-intersection of different body parts. 

Lastly, several recent works model a parametric model of human bodies~\cite{mihajlovic2021leap, alldieck2021imghum}, clothed human~\cite{saito2021scanimate, tiwari2021neuralgif, palafox2021npms,  wang2021metaavatar}, hands~\cite{karunratanakul2020graspingfield}, or clothing~\cite{corona2021smplicit} as neural implicit surfaces. 
One unique property of human body is the articulation with non-rigid deformations. 
Occupancy flow~\cite{niemeyer2019occupancyflow} models the non-rigid deformation of human bodies by warp fields. 
NASA~\cite{deng2020nasa} presents articulated, per-body-part occupancy fields to model a pose-driven human body. 
However, articulated occupancy fields may suffer from artifacts around joints due to discontinuities. 
Another approach to handle articulation is jointly learning shapes in the un-posed canonical space and transformations from the canonical space to the posed space~\cite{saito2021scanimate, mihajlovic2021leap, palafox2021npms, chen2021snarf}, leading to continuous deformations around body joints. 
The transformation can be modeled as warp fields (Section \ref{subsubsec:st_warp}) in the form of displacements~\cite{palafox2021npms} or Linear Blend Skinning weights~\cite{saito2021scanimate, mihajlovic2021leap, chen2021snarf, tiwari2021neuralgif, wang2021metaavatar}. 
\section{Generative Modeling}
\label{sec:generativemodeling}
Assuming a dataset of samples drawn from a distribution $\mathbf{y} \sim \mathcal{D}$, 
generative modeling defines a latent distribution $\latentdistrib$, such that every sample $\mathbf{y}$ can be identified with a corresponding latent $\latent \sim \latentdistrib$. 
The mapping from $\latentdistrib$ to samples from $\mathcal{D}$, is performed via a learned generator $\mathbf{g}$, parameterized as a deep neural network,  $\mathbf{g}(\latent)=\mathbf{y}$.

Neural fields provide greater flexibility for generative modeling because their outputs can be sampled at arbitrary resolutions, and because input samples can be shifted by simply applying transforms to the input coordinates.
Each sample $\mathbf{y}$ is associated with a neural field $\Phi$ that can be densely queried across the coordinate domain to obtain the sample $\mathbf{y}$.
Consistent with the notation in Section~\ref{sec:generalization}, latent variables $\latent$ can either \emph{globally} or \emph{locally} condition the neural field $\Phi$, yielding a conditional neural field $\Phi(\coord, \latent)$.

\parahead{Generative Modeling of Images}
In generative modeling of images, samples $\mathbf{y}$ are images, and we assume that there exists a distribution $\mathcal{D}$ of ``natural'' images.
The neural field $\Phi$ maps 2D pixel coordinates to RGB colors, $\Phi:\mathbb{R}^2 \to \mathbb{R}^3$.
Bond-Taylor and Willcocks~\cite{bond2020gradient} adopt a single-step gradient-based meta-learning approach for variational inference of images with a SIREN~\cite{sitzmann2020siren} decoder.
INR-GAN~\cite{skorokhodov2021inrgan} leverage global conditioning and a hypernetwork predicting a low-rank decomposition of the weight matrices of $\Phi$.
CIPS~\cite{anokhin2021cips} and StyleGAN3~\cite{karras2021alias} leverage global conditioning with a FiLM-style conditioning.
The latter three approaches leverage Fourier Features \cite{tancik2020ffn} to allow generation of high-frequency content. At the time of writing this report, neural field-based image generative models are the state-of-the-art.

\parahead{Generative Modeling of 3D Shape}
IM-Net~\cite{chen2019imnet} generates 3D shapes by parameterizing a neural field of occupancy $\Phi:\mathbb{R}^3\to\mathbb{R}^1$ and globally conditioning on latents $\latent$ via concatenation. Similarly, Occupancy Networks~\cite{mescheder2019occupancynetworks} employ a variational autoencoder (VAE) to learn a generative model of 3D occupancy fields. DeepSDF, on the other hand, learns the distribution of SDFs by directly optimizing shape latent codes in the spirit of Generative Latent Optimization~\cite{bojanowski2017optimizing} without encoders.

\begin{figure}
    \centering
    \includegraphics[width=\linewidth]{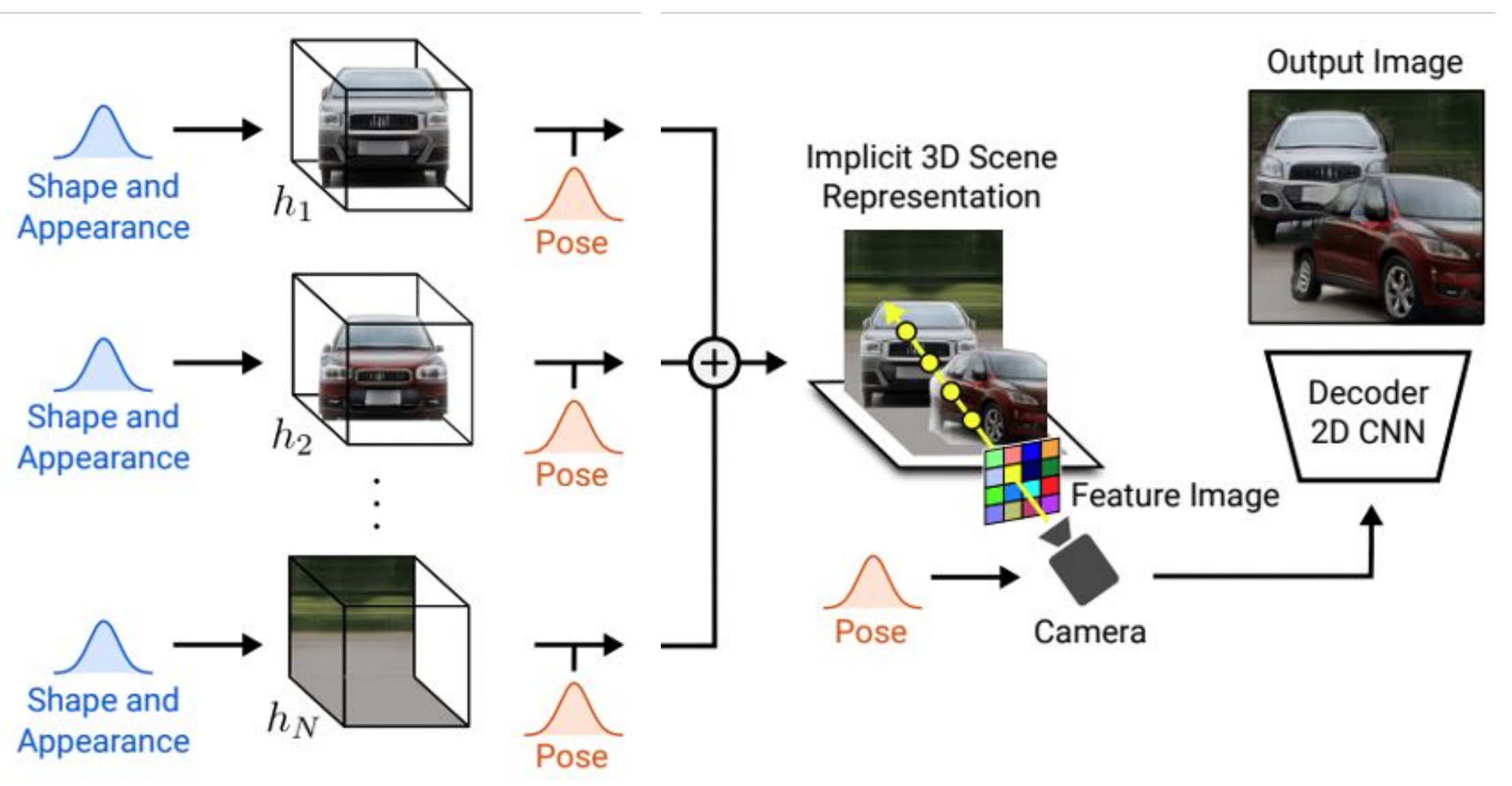}
	\caption{By leveraging object-centric local conditioning, Giraffe~\cite{niemeyer2021giraffe} learns a generative model that learns to disentangle a scene into separate 3D object representations.}
	\label{fig:giraffe}
	\vspace{-15pt}
\end{figure}

\parahead{Generative Modeling of 3D Shape and Appearance}
Instead of directly modeling the distribution of images, neural fields can parameterize distributions over 3D shape and appearance given only an image dataset.
The neural field $\Phi:\mathbb{R}^3\to \mathbb{R}^n$ then maps a \emph{3D} coordinate to a quantity that encodes shape and appearance. This is combined with a neural forward rendering model that renders an image given camera parameters.

%
GRAF~\cite{schwarz2020graf} first adapted this via a neural radiance field and volume rendering forward model, globally conditioned via concatenation.
Chan et al~\cite{chan2021pigan} improved upon GRAF by using sinusoidal activation functions~\cite{sitzmann2020siren} in the MLP, and global FiLM conditioning.
StyleNeRF~\cite{anonymous2022stylenerf} combine StyleGAN-style FiLM conditioning with accelerated rendering techniques such as low-resolution rendering and 2D upsampling.
By leveraging local conditioning via object-centric compositions (see Section~\ref{sec:hybrid_rep}), GIRAFFE~\cite{niemeyer2021giraffe} generates an output image as the composition of multiple neural fields, which allows control over shape, appearance and scene layout (Figure~\ref{fig:giraffe}).
CAMPARI~\cite{niemeyer2021campari} trains an additional camera generator along with the 3D volume generator to generalizes to complex pose distributions.

\parahead{Multi-modal Generative Modeling} An advantage of neural fields is that they are, in principle, agnostic to the signal they parameterize. Du et al.~\cite{du2021gem} combine global conditioning, the auto-decoder framework, and latent space regularization to learn multi-modal (such as audio-visual) manifolds. 

\section{2D Image Processing}
\label{sec:2D_neural_fields}

A compelling feature of 2D neural fields is the ability to represent continuous images. The first neural network to parameterize an image was demonstrated by Stanley et al.~\cite{stanley2007compositional}. These networks were not fit via gradient descent, instead relying on architecture search in a genetic algorithm framework, and could thus not represent images with fine detail. Fitting natural images with neural fields in a modern deep-learning framework was first demonstrated by SIREN~\cite{sitzmann2020siren} and FFN~\cite{tancik2020ffn}, and by \cite{sbai2019unsupervised} who decomposed the image into continuous vector layers. Unlike grid-based convolutional architectures, continuous images can be sampled at any resolution.
As a result, they have been used for a variety of image processing tasks described below.

%
%

\parahead{Image-to-image Translation}
These techniques take an input image and map it to another image that preserves some representation of the  content.   
%
Common tasks include image enhancement, super-resolution, denoising, inpainting, semantic mapping and generative modeling~\cite{isola2017image}.
%
Since this task requires learning a prior from data, an encoder-decoder architecture is often used, where the encoder is a convolutional neural network, and the decoder is a locally conditioned 2D neural field (Section~\ref{sec:generalization}). 
%

\begin{figure}[t]
	\centering
    \includegraphics[width=\linewidth]{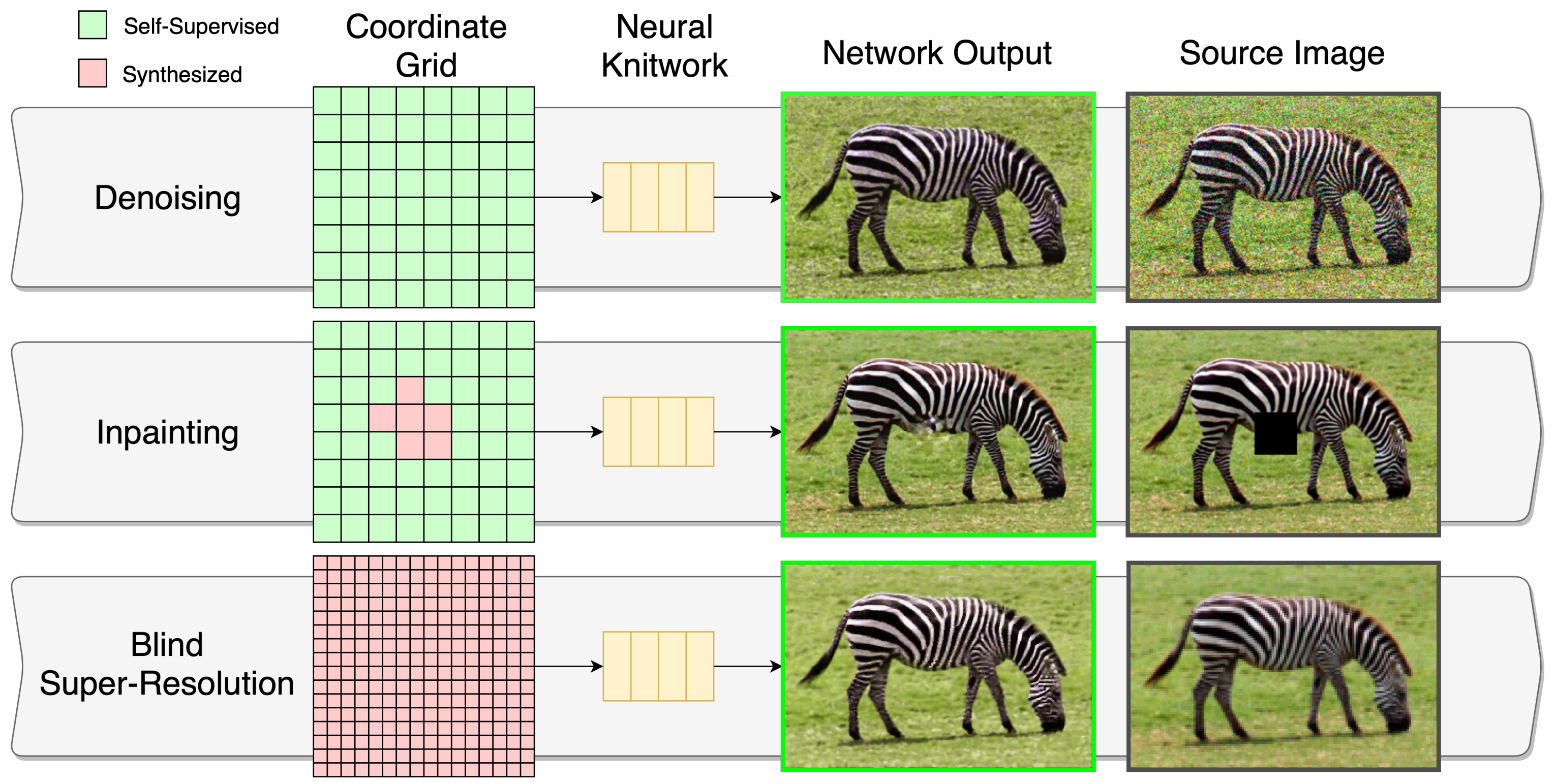}
	\caption{Neural Knitworks~\cite{czerkawski2021neuralknitworks} utilizes patch based 2D neural fields for image processing tasks: denoising, blind super-resolution and inpainting.}
	\label{fig:img_processing}
	\vspace{-15pt}
\end{figure}

Chen et al.~\cite{chen2021liif} propose such a locally conditioned neural field with a convolutional encoder for the task of image super-resolution, naturally leveraging the resolution independence of the neural field decoder.
Shaham et al.~\cite{shaham2021asapnet} leverage a convolutional encoder to produce a low-resolution, 2D feature map from an input image, which is then upsampled with nearest-neighbor interpolation to locally condition a decoder neural field. They demonstrate speedups for a variety of image-to-image translation tasks, such as segmentation and segmentation-to-RGB image as compared to the fully convolutional baseline.
Neural Knitworks~\cite{czerkawski2021neuralknitworks} discretize the 2D space into patches to introduce the appropriate receptive field, for inpainting, super-resolution, and denoising (Figure~\ref{fig:img_processing}).
CIPS~\cite{anokhin2021cips} proposes an image synthesis architecture whose input pixels coordinates are conditionally-independent given latent vector $\textbf{z}$.
Global information is given by $\textbf{z}$, which is used by a hypernetwork to modulate the neural field weights.
INR-GAN \cite{skorokhodov2021inrgan} similarly uses a hypernetwork and latent code $\textbf{z}$ to modulate the linear layer weights and biases in the neural field, for image generation.
Henzler et al. \cite{henzler2020neuraltexture} learns 2D texture exemplars and maps them into 3D by sampling random noise fields at desired positions as the neural texture field input.
PiCA \cite{ma2021pica} applies a lightweight SIREN to predict human facial texture over a guide mesh, where the 2D coordinate input lie in the UV space.
Li et al. \cite{li2021unsupervised} use a 2D neural field to parameterize deformation, for recovering images distorted by turbulent refractive media.

\parahead{Image Reconstruction} X-Fields \cite{bemana2020xfields} parameterizes the image as a 2D neural field conditioned on time and illumination to enable time and illumination interpolation.
Alternatively, Nam et al.~\cite{nam2021neural} represents dynamic images (video) with 2D neural fields, with additional homography warp, optical flow, or occlusion operations to model dynamic changes.
Kasten et al. \cite{kasten2021layered} represent dynamic images as time-dependent 2D neural fields, where individual foreground components are segmented as atlas, and alpha composited to obtain the final rendering.


\section{Robotics}
\label{sec:robotics}
Robotics requires complex perception systems that allow agents to efficiently infer, reason about, and manipulate representations of real-world scenes. 
Many robotics problems share similarities with visual computing, such as using 3D reconstruction for robot navigation, and so neural fields have recently been used in robotics too.
%
%
We discuss neural fields for robot perception, planning, and control.

\subsection{Localization via Camera Parameter Estimation}
%
Cameras observe the 3D world via 2D images. The projection of a world location onto the image plane is obtained through the extrinsic and intrinsic matrices, where the extrinsic matrix $[ \mathbf{R}|\mathbf{t} ]$ defines the 6DoF transformation between the world coordinate frame and the camera coordinate frame, while the intrinsics matrix $\mathbf{K}$ describes the projection of a 3D point onto the 2D image plane. Commonly, these camera parameters are obtained via SFM and keypoint matching with off-the-shelf tools like COLMAP \cite{schoenberger2016sfm, schoenberger2016mvs}.
Since neural rendering is end-to-end differentiable, camera parameters can be jointly estimated with the neural field making them useful for Simultaneous Localization and Mapping (SLAM) \cite{sucar2021imap} and Absolute Pose Regression (APR) \cite{chen2021directposenet}.

There exist numerous ways to parameterize camera intrinsic and extrinsic matrices. Representing the extrinsic matrix, particularly rotation, however, is a long-standing challenge. Since the 3-by-3 rotation matrix lies on $SO(3)$, continuity is not guaranteed \cite{zhou2019continuity, levinson2020analysis}. Consequently, alternative parameterizations were used in neural field literature: exponential coordinates \cite{yen-chen2021inerf}, Rodriguez formula \cite{wang2021nerf--}, continuous 6D representation \cite{meng2021gnerf, zhou2019continuity}, Euler angle\cite{azinovic2021neural}, and homography warp (for planar scenes) \cite{meng2021gnerf, yen-chen2021inerf}.

Furthermore, joint reconstruction and registration is a long-standing chicken-and-egg problem: camera parameters are needed to reconstruct the scene, and a reconstruction is needed to estimate camera parameters.
One simplification is to assume known reconstruction and solve only the registration problem. iNeRF \cite{yen-chen2021inerf} estimates camera poses given pre-trained neural fields, thus \textit{inverting} the problem statement. Due to the strong assumption, iNeRF has a limited use case of registering new, un-posed images to an already-reconstructed scene.
Furthermore, the optimization problem is known to be non-convex. A coarse initialization of camera parameters alleviates this challenge. As such, several works \textit{refine} camera parameters during reconstruction \cite{azinovic2021neural, lin2021barf, jeong2021selfcalibrating, wang2021nerf--, chen2021directposenet}. The chicken-and-egg nature of the problem is reflected by NeRF-- \cite{wang2021nerf--}, in which the authors jointly optimizes the scene and cameras, but retrain scene reconstruction with the optimized camera parameters for better reconstruction quality.

The more challenging problem of estimating unknown camera parameters, while jointly reconstructing the scene has been looked into by several works~\cite{lin2021barf,wang2021nerf--}. However, the scene is assumed to be forward-facing and the camera pose initialization relies on hand-crafted rules (\eg~cameras centered at origin, facing the -z axis, with focal lengths equal to the reference image size \cite{wang2021nerf--}). GNeRF \cite{meng2021gnerf} further relaxes these assumptions, and supports inward-facing $360^\circ$ scenes. The optimization between reconstruction and registration proceeds iteratively, rather than jointly, and are softly coupled via a discriminator.
Approaches to overcome local minima in optimization and preserve details include coarse-to-fine training~\cite{lin2021barf, jeong2021selfcalibrating,chen2021directposenet}, high-frequency positional encoding weights \cite{chen2021directposenet,lin2021barf}, and curriculum learning~\cite{jeong2021selfcalibrating} for optimizing focal lengths followed by intrinsics and lens distortion.


In the examples above, supervision is provided via appearance. Additional signals such as depth map from RGB-D sensor could provide strong signals for camera parameter optimization \cite{sucar2021imap}. Similarly, time-of-flight image may also reduce the problem complexity~\cite{attal2021torf}. 
A closely related topic is object pose estimation. While object pose can be modelled explicitly, a different approach is to predict a \emph{probability distribution} of pose over $SO(3)$~\cite{murphy2021implicitpdf}. The probability distribution is itself a neural field, mapping rotation to probability. The approach is especially useful for symmetric objects with multiple valid solutions.



\subsection{Planning}
Planning in robotics is the problem of identifying a sequence of valid robot configuration states to achieve an objective.
This includes path planning for navigation, trajectory planning for grasping or manipulation, or planning for interactive perception~\cite{bohg2017interactiveperception}.
Given the spatial nature of planning problems, neural fields have been used as a representation in many solutions.

Among the first attempts to use neural networks for robot path planning was by Lemmon~\cite{lemmon1991bneuralfieldspathplanning,lemmon1991aneuralfieldspathplanning}.
While this work refers to a 2D grid of neurons as a ``neural field,'' the neurons are mapped one-to-one to a 2D map and is thus a (non-continuous) coordinate-based network.
Their method finds the variational solution of Bellman's dynamic programming equation~\cite{bellman1954theory} used in path planning.
Other coordinate-based representations have been used for planning, including estimating affordance maps~\cite{kim2015_affordancemap,qi2020learning}.
Zhang et al. parameterize a Fisher Information Field~\cite{zhang2020fisher} via neural network. 
World constraints can be specified during planning, for example, by specifying constraints as a level set in a high-dimensional space~\cite{sutanto2020learning} which is learned as a neural field.


Grasping and manipulation problems require knowing 2D or 3D position of a robot gripper relative to the surface of objects.
Some approaches model the shape of the object as a collection of points~\cite{boularias2011_robot_grasping}, and learn potentially good points for grasping or manipulation using a Markov Random Field (MRF).
This approach can handle points in any continuous 3D position as long as an MRF can be built.
Similarly, other data-driven grasping methods often use coordinate-based representations~\cite{bohg2014graspsynthesisreview}.
ContactNets~\cite{pfrommer2020contactnets} models contact between objects via learned neural fields.
Neural fields can also synthesize human grasps~\cite{karunratanakul2020graspingfield}. 
GIGA~\cite{jiang2021giga} uses locally-conditioned neural fields encoding quality, orientation, width, and occupancy for grasp selection.

%

\subsection{Control}
Controllers are responsible for realizing plans, while ensuring that physical constraints and mechanical integrity are preserved. 
Control can be achieved either by relying on a planner or directly from observations.
Neural fields have been used for this task by learning an obstacle barrier function approximated by an SDF ~\cite{long2021learning}.
Similarly, Bhardwaj et al.~\cite{bhardwaj2021fast} solve the robot arm self-collision avoidance task by using neural fields to predict the closest distance between robot links, given its joint configuration.
In visuomotor control, control is driven directly by visual observations.
Li et al.~\cite{li20213d} use NeRF to facilitate view-invariant visuomotor control to achieve robot goal states specified via a 2D goal image. This is achieved by auto-decoding a dynamics model to support future prediction and novel view synthesis.

%

\begin{figure}
    \centering
    \includegraphics[width=\linewidth]{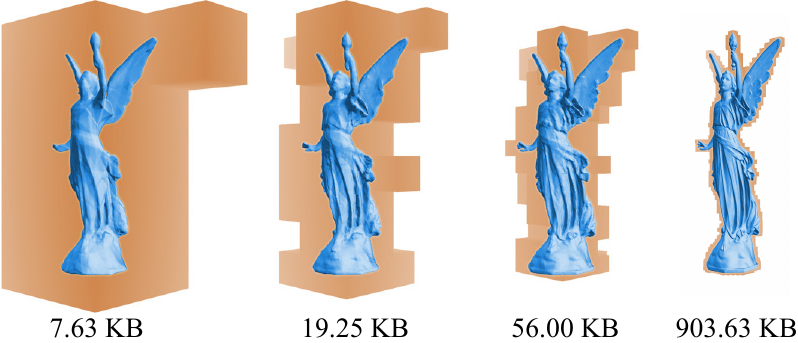}
	\caption{Neural Level of Detail (NGLoD) combines level-of-detail with neural fields as a hybrid continuous-discrete shape representation, demonstrating compression of 3D geometry. Figure adapted from~\cite{takikawa2021nglod}.}
	\label{fig:compression}
	\vspace{-15pt}
\end{figure}

\section{Lossy Compression}
\label{sec:compression}
The goal of lossy data compression is to approximate a signal as best as possible with as few bits as possible. These opposing forces naturally form a tradeoff which can be characterized as a Pareto frontier: the rate-distortion curve.
%
In practice, signals are often stored as discrete sequences of data which are transformed into an alternate basis such as the discrete cosine transform which help to decorrelate the signal (making downstream tasks like quantization and entropy coding more effective). Many standards exist, such as JPEG \cite{wallace1992jpeg} for images or HEVC \cite{sullivan2012hevc} for videos.
Recent work has explored the potential of neural fields as an alternate signal storage format which directly represents the continuous signal with a parameteric, continuous function.
Compression may be achieved in one of two ways. First, by leveraging the inductive bias of the network architecture itself, and simply overfitting a neural field to a signal, i.e., finding parameters $\Theta$ of a neural field $\Phi$. Only the parameters and architecture of the neural field need to be stored.
Second, \emph{prior-based} compression schemes achieve compression via learning a space of low-dimensional latent code vectors $\mathbf{z}$ that may be decoded into neural field parameters, where the storage cost of the decoder is amortized over many latent codes.


Encoding 3D geometry with neural fields presents an alternative to conventional mesh representations that may enable a significantly reduced memory footprint.
Subsequently, SIREN~\cite{sitzmann2020siren} demonstrated fitting a wide array of signals with neural fields: audio, video, images, and 3D geometry (including large-scale scenes).
Lu et al.~\cite{lu2021compressive} show that SIREN in conjunction with scalar weight quantization can compress dense volumetric data better than state-of-the-art approaches at the cost of higher encoding and decoding latency.
Davies et al.~\cite{davies2021on} compare neural field 3D geometry with decimated meshes and found better reconstruction quality with the same memory footprint. 
Takikawa et al.~\cite{takikawa2021nglod} show that a hierarchical tree structure can be used to learn multiresolution signals that can perform level-of-detail more effectively than decimated meshes.
Dupont et al.~\cite{dupont2021coin} compared the memory use of neural fields parameterizing images, and found neural fields can outperform JPEG~\cite{wallace1992jpeg} but not state-of-the-art image compression techniques.
ACORN \cite{martel2021acorn} use an adaptive quadtree data structure to fit high resolution images with neural fields, but do not outperform traditional image compression methods.
For dynamic 3D scenes, DyNeRF~\cite{li2021dynerf} uses 28 MB of memory for a 10-second 30 FPS \emph{3D} video sequence.
Light Field Networks~\cite{sitzmann2021lfns} allow a large reduction in memory used over a classical discrete light field.
General network compression and quantization techniques can further reduce network size~\cite{isik2021neural, bird2021cnerf}.
Bird et al.~apply entropy penalization to NeRF~\cite{bird2021cnerf}, and obtain higher compression rates compared to standard HEVC video encoding~\cite{sullivan2012hevc} and LLFF~\cite{mildenhall2019llff} for forward-facing scenes. 

The above methods study a variety of continuous signal modalities, and use different datasets, architectures, and metrics, making comparisons difficult.
Few rigorous comparisons exist between neural field and conventional compression schemes.
Many works lack ablation of design choices and the ability to make direct queries to the without decoding.
While neural fields for compression remains in its infancy at the time of writing, it is nonetheless a valuable perspective to consider signals as functions and neural networks as a data format.

\section{Beyond Visual Computing}
\label{sec:beyond}
Visual computing problems are a subset of all inverse problems which can be parameterized by neural fields. These problems often share the same challenges involving incomplete observations and the need for a flexible parameterization. In this section we will highlight some of the emerging research directions in neural fields beyond visual computing.

\label{sec:alternative_imaging}
\subsection{Alternative Signal Modalities}

Most of the works surveyed so far have been concerned with modeling the imaging process of consumer cameras, which measure the visible electromagnetic radiation via optical lenses, using sensors that digitize irradiance into intensity over a 2D raster grid. 
Nonetheless, neural fields can also model alternative signal modalities such as non-line-of-sight imaging \cite{shen2021nonlineofsight}, non-visible x-rays for computed tomography (CT) \cite{shen2021nerp, zang2021intratomo, 9606601}, magnetic resonance imaging (MRI) \cite{shen2021nerp}, pressure waves for audio~\cite{reed2021implicit}, chemiluminescence \cite{pan2021adaptive}, time-of-flight imaging \cite{attal2021torf}, as well as volumetric light displays~\cite{zheng2020neural}.
%
%

\parahead{Medical Imaging}
\label{sec:ct_mri}
\begin{figure}
    \centering
    \includegraphics[width=0.75\linewidth]{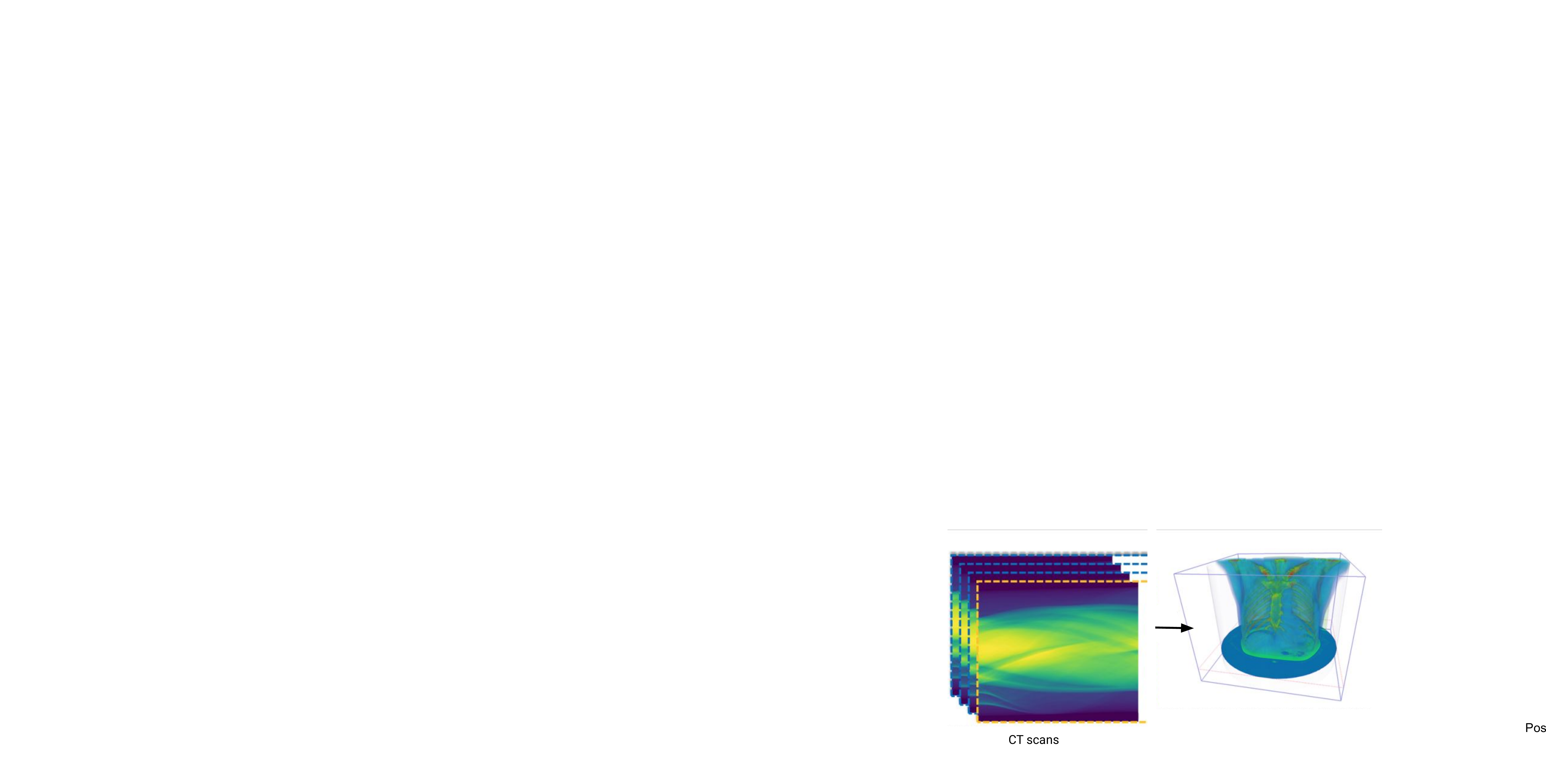}
	\caption{Reconstruction of 3D density field (right) from limited-angle CT measurements (left). Figure adapted from \cite{zang2021intratomo}.}
	\label{fig:ct}
	\vspace{-15pt}
\end{figure}
In CT and MRI, raw sensor measurements are the Radon and Fourier transformation of spatially-varying density, respectively \cite{shen2021nerp}. The sensor domains are not human-readable, while the reconstructed density volume (whose 2D slices are called the image domain) is. Reconstructing the density volume is an ill-posed problem \cite{shen2021nerp}, and classical techniques are sensitive to measurement noise \cite{zang2021intratomo}. Reconstruction and NVS in medical imaging is also limited by capture constraints such as scene movement, finite sensor resolution, limited viewing angle, and sparse views (to reduce X-ray exposure, and speed up procedure).

Neural fields can either parameterize the sensor domain \cite{9606601}, or the density domain directly \cite{shen2021nerp, zang2021intratomo}. In the former, the neural field maps sensor coordinates to predicted sensor activations, and can augment real measurement data, before applying classical reconstruction technique (filtered back-projection \cite{kak2001principles}) \cite{9606601}. In the latter the neural field directly predicts the density value at a 3D spatial coordinate, and is supervised by mapping its output to the sensor domain via Radon (CT) or Fourier (MRI) transform \cite{shen2021nerp, zang2021intratomo} (Figure~\ref{fig:ct}).

Similarly, in cryo-electron microscopy (cryo-EM), the 2D sensor measures the convolution between a point spread function and electron density. In CryoGRGN \cite{zhong2020cryodrgn}, the neural field maps a 3D coordinate to the deconvolved electron density.

\parahead{Audio}
\label{sec:audio}
Audio signals can be represented as raw waveforms or spectrograms, and each can be parameterized by neural fields \cite{sitzmann2020siren, gao2021objectfolder}. Raw waveform is a continuous 1D field function that maps temporal coordinates to amplitude, often stored as a collection of discrete samples. A spectrogram is the Fourier transform of a waveform, which is a 2D field function mapping time and frequency to amplitude. Neural field methods fit all time-dependent waveforms, which means that the techniques are applicable to all waves mechanical (e.g., seismic waves, ocean waves, acoustic waves) or electromagnetic (e.g., wave optics, radio waves).

Synthetic aperture sonar (SAS) also exhibits the common problems in inverse problems: noisy sensor domain and ill-posed problem. Reed et al.~\cite{reed2021implicit} use a neural field mapping a 2D position to a distribution of point scatter (reconstruction domain), and obtain supervision by mapping to the sensor domain via convolution.





\subsection{Physics-informed Problems}
\label{sec:PINN}

Physics-informed problems have solutions that are restricted to a set of partial differential equations (PDEs) based on laws of physics. The solutions are often continuous in spatio-temporal coordinates. Neural fields are therefore a natural parameterization of the solution space, given that neural networks are continuous, differentiable, and universal function approximators. These neural fields are also referred to as physics-informed neural networks (PINNs), whose use was first popularized by \cite{raissi2019physics}. Problems constrained by nonlinear PDEs often require arbitrarily-small step sizes for traditional methods, which often require prohibitive computation resources. Parameterizing the solution via neural networks re-formulates these problems as \textit{optimization}, rather than \textit{simulation}, which is more data-efficient \cite{raissi2019physics}. Since many physical processes in nature are governed by PDEs (boundary conditions), supervising (or regularizing) the neural network via its gradients is a common technique. In fact, the Eikonal regularization for SDF is an example in visual computing. Physics-informed problems include topology optimization \cite{zehnder2021ntopo}, geodesy estimation \cite{izzo2021geodesynets}, collision dynamics estimation \cite{pfrommer2020contactnets}, solving the Schrodinger Equation \cite{raissi2019physics}, Navier-Stokes equation \cite{raissi2019physics}, and Eikonal equation \cite{smith2020eikonet}.


\chapter{Discussion \& Conclusion}
\label{sec:discussion}
\vspace{-0.15in}
\section{Discussion}
As we have seen, neural fields have a rich but evolving technical `toolbox', and a rapidly increasing space of applications both within and outside of visual computing.
In our view, there are several factors that have resulted in this progress.

First, the idea of parameterizing a continuous \emph{field} using an MLP without the need to use more complex neural network architectures has simplified the training of fields and reduced the entry barrier~\cite{gargan1998approximating}.
Neural fields provide a fundamentally different approach to signal processing that is no longer discrete and more faithful to the original \emph{continuous} signal.
%
Second, techniques such as positional encoding and sinusoidal activations have significantly improved the quality of neural fields leading to large leaps in applications focused on quality.
Finally, applications in novel view synthesis and 3D reconstruction have been particularly important in popularizing neural fields because of the visually appealing nature of these applications~\cite{mildenhall2020nerf,sitzmann2019srn,groueix2018atlasnet,park2019deepsdf}.
Third, researchers have realized that differentiable volume and voxel rendering commonly used in novel view synthesis methods~\cite{mildenhall2019llff,yu2021plenoxels} can be useful in solving others tasks like 3D reconstruction~\cite{yariv2021volsdf} and even semantic segmentation~\cite{zhi2021semanticnerf,vora2021nesf}.

\parahead{Future Directions}
Despite the progress, we believe that neural fields have only started to scratch the surface and there remains great potential for continued progress in both techniques and applications.
In terms of \textbf{techniques}, a common limitation of many neural fields is their inability to generalize well to unseen data.
We believe that integration of stronger priors can enable these methods to generalize better and be data efficient.
Other inductive biases in the form of task-specific heuristics, laws of physics, or network architecture can further help generalization.
Building a \textbf{common framework} for incorporating these priors is a fruitful direction for future work.
Furthermore, the rapid progress has been at the expense of methodical evaluation and analysis of common techniques.
We need shared \textbf{datasets} and \textbf{benchmarks} on which different techniques can be evaluated and compared.

In terms of \textbf{applications}, a majority of neural field methods have thus far been used to solve ``low-level'' (\eg~image synthesis) and ``mid-level'' (\eg~3D reconstruction) tasks.
The application of neural fields for ``high-level'' \textbf{semantic tasks} remains an open problem.
Examples of these problems include understanding 3D scene layout~\cite{wang2021sceneformer}, 3D scene interaction~\cite{li2021learning}, and grouping of data into more meaningful entities~\cite{tschernezki2021neuraldiff}.
Furthermore, neural fields have focused on a single data modality, but exploring the fusion of \textbf{multiple modalities} could be a fruitful topic of research.
For example, synthesizing fields based on language input~\cite{jain2021dreamfields}, or joint modeling of image and text or audio input could foster closer connections with the NLP community.
Finally, future work should consider moving beyond supervised learning of fields and consider \textbf{weakly- or self-supervised learning} as an alternative.
Building upon advances in 3D deep learning such as transformation equivariance~\cite{sajnani2022condor,spezialetti2020learning,sun2021canonical} could allow neural fields to be data efficient and generalize better.

Finally, the neural fields community must become more self-aware to build a \textbf{culture} that promotes scholarship, sustainable growth, inclusivity, and diversity.
We must pay careful attention to past work and avoid repetition of work through ``\emph{selective amnesia}''~\cite{su2021affective}.
We must avoid becoming a ``\emph{scientific bandwagon}''~\cite{shannon1956bandwagon} by acknowledging limitations of neural fields and by collaborating with domain experts to identify shortcomings.
While the explosion of work in neural fields helps us make important advances, we must be aware of its impact on the mental health and well-being of researchers~\cite{su2021affective}.


\parahead{Societal Impact}
As the applications of neural fields increase, so will their societal impact, especially in domains where they make a significant difference.
In generative modeling of audio, imagery, 3D scenes, etc., neural fields have enabled the generation of realistic content for the purpose of deception, for instance, impersonating the image and voice of actors without their consent.
Recent work explores the detection of digitally-altered imagery~\cite{tolosana2020deepfakes}. 
Similarly, improving the capabilities of inverse-problem solvers, such as denoising and blur-removal algorithms, has implications for privacy.
For instance, this may enable de-anonymizing recordings that were previously assumed to not contain sufficient information to allow identification of participating actors.
It may also decrease the cost of surveillance, making it more widely available as the hardware requirements may decrease.  
Finally, neural fields could also have negative environmental impact as significant computational resources are spent optimizing neural networks with GPUs.

Neural fields can also positively impact society.
Democratizing the generation of photo-realistic imagery helps more artists and content creators to tell stories.
The ability to aggregate information from low-dimensional supervision (\eg~2D images) relaxes hardware constraints for 3D content creation.
Neural fields for computer vision may help build robotic automation that helps people.

\section{Conclusion}
This report provides an overview of the flourishing research direction of neural fields.
From a review of over \numpapers papers, we have summarized five classes of techniques using a shared mathematical formulation, including prior learning and conditioning, hybrid representations, forward maps, network architectures, and manipulation methods.
Then, we have surveyed applications across graphics, vision, robotics, medical imaging, and computational physics. 
Neural fields research will continue to grow. 
To aid in continued understanding, we have established a living report as a \href{https://neuralfields.cs.brown.edu/}{community-driven website}, where authors can submit their own papers and classify them using the taxonomy developed in this report. 
Neural fields will be a key enabler for progress across many areas of computer science, and we look forward to the research yet to come.
\chapter{Authors}

\textbf{Yiheng Xie} (\href{https://yxie20.github.io/}{https://yxie20.github.io/}) is a final-year undergraduate student at Brown University and a researcher at Unity Technologies. His research interests include 3D reconstruction, physically-based rendering, and robotics. He is a member of Brown Visual Computing (BVC), advised by Professor Srinath Sridhar, and Brown Humanity Centered Robotics Initiative (HCRI), advised by Professor Michael Littman. 

\textbf{Towaki Takikawa} (\href{https://tovacinni.github.io/}{https://tovacinni.github.io/}) is a Ph.D.~student at the University of Toronto with Prof.~Sanja Fidler and Prof.~Alec Jacobson. He is also a Research Scientist at NVIDIA in the Hyperscale Graphics Systems group. He received his bachelors in Computer Science at the University of Waterloo.

\textbf{Shunsuke Saito} (\href{http://www-scf.usc.edu/~saitos/}{http://www-scf.usc.edu/{\textasciitilde}saitos/}) is a Research Scientist at Meta Reality Labs in Pittsburgh. He finished his Ph.D.~at University of Southern California, where he worked with Prof.~Hao Li. Prior to USC, he spent one year at University of Pennsylvania as Visiting Researcher. He obtained B.Eng. and M.Eng.~in Applied Physics at Waseda University in 2013 and 2014.

\textbf{Or Litany} (\href{https://orlitany.github.io/}{https://orlitany.github.io/}) is a Research Scientist at NVIDIA. Prior, he was a postdoc at Stanford University working under Prof. Leonidas Guibas, and a postdoc at FAIR working under Prof. Jitendra Malik. Previously, he was a postdoc at the Technion and a research intern at Microsoft, Intel and Google. He received his PhD from Tel-Aviv University, advised by Prof. Alex Bronstein. He received my B.Sc. in Physics and Mathematics from the Hebrew University under the auspices of “Talpiot.”

\textbf{Shiqin Yan} (\href{https://player-eric.com/about}{https://player-eric.com/about}) is a masters student in Computer Science at Brown University. He is passionate about solving real-world problems with large-scale web-based systems.

\textbf{Numair Khan} (\href{http://cs.brown.edu/~nkhan6/}{http://cs.brown.edu/{\textasciitilde}nkhan6/}) is a PhD student at Brown University where his research focuses on methods and representations for scene reconstruction, differentiable rendering, and novel-view synthesis.

\textbf{Federico Tombari} (\href{https://federicotombari.github.io/}{https://federicotombari.github.io/}) is a Research Scientist and Manager at Google Zurich (Switzerland), where he leads an applied research team in Computer Vision and Machine Learning. He is also affiliated to the Faculty of Computer Science at TU Munich (Germany) as lecturer (Privatdozent).

\textbf{James Tompkin} (\href{https://jamestompkin.com/}{https://jamestompkin.com/}) is an assistant professor in visual computing at Brown University. He was a PhD student at UCL, with postdocs at the Max Planck Institute and Harvard. His lab develops visual understanding techniques for camera-captured media to remove barriers from image and video creation, editing, and interaction. This requires image and scene reconstruction techniques, especially from multi-camera systems. 

\textbf{Vincent Sitzmann} (\href{https://vsitzmann.github.io/}{https://vsitzmann.github.io/}) is an incoming assistant professor at MIT EECS. Currently, he is a Postdoc at MIT's CSAIL with Joshua Tenenbaum, William Freeman, and Frédo Durand. Previously, he finished his Ph.D. at Stanford University. His research interest lies in neural scene representations --- the way neural networks learn to represent information on our world. His goal is to allow independent agents to reason about our world given visual observations, such as inferring a complete model of a scene with information on geometry, material, lighting etc. from only few observations, a task that is simple for humans, but currently impossible for AI.

\textbf{Srinath Sridhar} (\href{https://srinathsridhar.com/}{https://srinathsridhar.com/}) is an assistant professor of computer science at Brown University. His research interests are in 3D computer vision and machine learning. Specifically, he focuses on visual understanding of 3D human physical interactions with applications ranging from robotics to mixed reality. He has won several fellowships (e.g., Google Research Scholar) and awards (e.g., Eurographics Best Paper Honorable Mention) for his work, and has previously spent time at Stanford, Max Planck Institute for Informatics, Microsoft Research Redmond, and Honda Research Institute.

\chapter{Acknowledgements}
This work was supported by NSF CNS-2038897 and the Google Research Scholar Program. We thank Sunny Li for their help in designing the website, Jayden Yi for a conceptual readthrough, and Alexander Rush and Hendrik Strobelt for the \href{https://github.com/Mini-Conf/Mini-Conf}{Mini-Conf} project.
\nocite{*}
\bibliographystyle{eg-alpha-doi} 
\bibliography{more_ref,references}

\newcommand{\etalchar}[1]{$^{#1}$}
\begin{thebibliography}{\uppercase{PCPMMN21}}

\bibitem[ACC{\etalchar{*}}21]{adamkiewicz2022visiononly}
\textsc{Adamkiewicz M., Chen T., Caccavale A., Gardner R., Culbertson P., Bohg
  J., Schwager M.}:
\newblock Vision-only robot navigation in a neural radiance world.
\newblock URL: \url{http://arxiv.org/abs/2110.00168v1}.

\bibitem[ADK{\etalchar{*}}21]{anokhin2021cips}
\textsc{Anokhin I., Demochkin K., Khakhulin T., Sterkin G., Lempitsky V.,
  Korzhenkov D.}:
\newblock Image generators with conditionally-independent pixel synthesis.
\newblock In \emph{Proceedings of the IEEE/CVF Conference on Computer Vision
  and Pattern Recognition (CVPR)} (2021).
\newblock URL: \url{http://arxiv.org/abs/2011.13775v1}.

\bibitem[AHY{\etalchar{*}}19]{atzmon2019controlling}
\textsc{Atzmon M., Haim N., Yariv L., Israelov O., Maron H., Lipman Y.}:
\newblock Controlling neural level sets.
\newblock In \emph{Advances in Neural Information Processing Systems (NeurIPS)}
  (2019), Curran Associates, Inc.
\newblock URL: \url{http://arxiv.org/abs/1905.11911v2}.

\bibitem[AHZ{\etalchar{*}}22]{attal2022learning}
\textsc{Attal B., Huang J.-B., Zollhoefer M., Kopf J., Kim C.}:
\newblock Learning neural light fields with ray-space embedding networks.
\newblock \emph{Proceedings of the IEEE/CVF Conference on Computer Vision and
  Pattern Recognition} (2022).

\bibitem[AL20a]{atzmon2020sal}
\textsc{Atzmon M., Lipman Y.}:
\newblock Sal: Sign agnostic learning of shapes from raw data.
\newblock In \emph{Proceedings of the IEEE/CVF Conference on Computer Vision
  and Pattern Recognition (CVPR)} (2020).
\newblock URL: \url{http://arxiv.org/abs/1911.10414v2}.

\bibitem[AL20b]{atzmon2020sald}
\textsc{Atzmon M., Lipman Y.}:
\newblock Sald: Sign agnostic learning with derivatives.
\newblock \emph{arXiv preprint arXiv:2006.05400} (2020).

\bibitem[ALG{\etalchar{*}}21]{attal2021torf}
\textsc{Attal B., Laidlaw E., Gokaslan A., Kim C., Richardt C., Tompkin J.,
  O'Toole M.}:
\newblock Torf: Time-of-flight radiance fields for dynamic scene view
  synthesis.
\newblock In \emph{Advances in Neural Information Processing Systems (NeurIPS)}
  (2021), Curran Associates, Inc.
\newblock URL: \url{http://arxiv.org/abs/2109.15271v1}.

\bibitem[AMBG{\etalchar{*}}21]{azinovic2021neural}
\textsc{Azinovic D., Martin-Brualla R., Goldman D.~B., Niessner M., Thies J.}:
\newblock Neural rgb-d surface reconstruction.
\newblock \emph{arXiv preprint arXiv:2104.04532} (2021).
\newblock URL: \url{http://arxiv.org/abs/2104.04532v1}.

\bibitem[AMK{\etalchar{*}}20]{antonova2020analytic}
\textsc{Antonova R., Maydanskiy M., Kragic D., Devlin S., Hofmann K.}:
\newblock Analytic manifold learning: Unifying and evaluating representations
  for continuous control.
\newblock \emph{arXiv preprint arXiv:2006.08718} (2020).

\bibitem[Ano22]{anonymous2022stylenerf}
\textsc{Anonymous}:
\newblock Stylene{RF}: A style-based 3d aware generator for high-resolution
  image synthesis.
\newblock In \emph{Submitted to The Tenth International Conference on Learning
  Representations} (2022).
\newblock URL: \url{https://openreview.net/forum?id=iUuzzTMUw9K}.

\bibitem[ANVL21]{atzmon2021augmenting}
\textsc{Atzmon M., Novotny D., Vedaldi A., Lipman Y.}:
\newblock Augmenting implicit neural shape representations with explicit
  deformation fields.
\newblock \emph{arXiv preprint arXiv:2108.08931} (2021).
\newblock URL: \url{http://arxiv.org/abs/2108.08931v1}.

\bibitem[ASS21]{athar2021flameinnerf}
\textsc{Athar S., Shu Z., Samaras D.}:
\newblock Flame-in-nerf : Neural control of radiance fields for free view face
  animation.
\newblock \emph{arXiv preprint arXiv:2108.04913} (2021).
\newblock URL: \url{http://arxiv.org/abs/2108.04913v1}.

\bibitem[AXS21]{alldieck2021imghum}
\textsc{Alldieck T., Xu H., Sminchisescu C.}:
\newblock {imGHUM}: Implicit generative models of 3d human shape and
  articulated pose.
\newblock In \emph{Proceedings of the IEEE/CVF International Conference on
  Computer Vision} (2021), pp.~5461--5470.

\bibitem[AZ21]{arandjelovic2021nerf}
\textsc{Arandjelovi{\'c} R., Zisserman A.}:
\newblock Nerf in detail: Learning to sample for view synthesis.
\newblock \emph{arXiv preprint arXiv:2106.05264} (2021).

\bibitem[BBJ{\etalchar{*}}21]{boss2021nerd}
\textsc{Boss M., Braun R., Jampani V., Barron J.~T., Liu C., Lensch H. P.~A.}:
\newblock Nerd: Neural reflectance decomposition from image collections.
\newblock In \emph{Proceedings of the IEEE International Conference on Computer
  Vision (ICCV)} (2021).
\newblock URL: \url{http://arxiv.org/abs/2012.03918v4}.

\bibitem[BBSC21]{bird2021cnerf}
\textsc{Bird T., Balle J., Singh S., Chou P.~A.}:
\newblock 3d scene compression through entropy penalized neural representation
  functions.
\newblock \emph{arXiv preprint arXiv:2104.12456} (2021).
\newblock URL: \url{http://arxiv.org/abs/2104.12456v1}.

\bibitem[Bel54]{bellman1954theory}
\textsc{Bellman R.}:
\newblock The theory of dynamic programming.
\newblock \emph{Bulletin of the American Mathematical Society 60}, 6 (1954),
  503--515.

\bibitem[BGP{\etalchar{*}}21]{baatz2021nerftex}
\textsc{Baatz H., Granskog J., Papas M., Rousselle F., Nov{\'a}k J.}:
\newblock Nerf-tex: Neural reflectance field textures.
\newblock \emph{Computer Graphics Forum} (2021).

\bibitem[BHS{\etalchar{*}}17]{bohg2017interactiveperception}
\textsc{Bohg J., Hausman K., Sankaran B., Brock O., Kragic D., Schaal S.,
  Sukhatme G.~S.}:
\newblock Interactive perception: Leveraging action in perception and
  perception in action.
\newblock \emph{IEEE Transactions on Robotics 33}, 6 (2017), 1273--1291.
\newblock \href {https://doi.org/10.1109/TRO.2017.2721939}
  {\path{doi:10.1109/TRO.2017.2721939}}.

\bibitem[BHumRZ21]{benbarka2021seeing}
\textsc{Benbarka N., Hofer T., ul-moqeet Riaz H., Zell A.}:
\newblock Seeing implicit neural representations as fourier series.
\newblock \emph{arXiv preprint arXiv:2109.00249} (2021).
\newblock URL: \url{http://arxiv.org/abs/2109.00249v1}.

\bibitem[BJB{\etalchar{*}}21]{boss2021neuralpil}
\textsc{Boss M., Jampani V., Braun R., Liu C., Barron J.~T., Lensch H. P.~A.}:
\newblock Neural-pil: Neural pre-integrated lighting for reflectance
  decomposition.
\newblock URL: \url{http://arxiv.org/abs/2110.14373v1}.

\bibitem[BJLPS17]{bojanowski2017optimizing}
\textsc{Bojanowski P., Joulin A., Lopez-Paz D., Szlam A.}:
\newblock Optimizing the latent space of generative networks.
\newblock \emph{arXiv preprint arXiv:1707.05776} (2017).

\bibitem[BKP11]{boularias2011_robot_grasping}
\textsc{Boularias A., Kroemer O., Peters J.}:
\newblock Learning robot grasping from 3-d images with markov random fields.
\newblock In \emph{2011 IEEE/RSJ International Conference on Intelligent Robots
  and Systems} (2011), pp.~1548--1553.
\newblock \href {https://doi.org/10.1109/IROS.2011.6094888}
  {\path{doi:10.1109/IROS.2011.6094888}}.

\bibitem[BKW21]{bergman2021metanlr++}
\textsc{Bergman A.~W., Kellnhofer P., Wetzstein G.}:
\newblock Fast training of neural lumigraph representations using meta
  learning.
\newblock \emph{arXiv preprint arXiv:2106.14942} (2021).
\newblock URL: \url{http://arxiv.org/abs/2106.14942v1}.

\bibitem[BMAK14]{bohg2014graspsynthesisreview}
\textsc{Bohg J., Morales A., Asfour T., Kragic D.}:
\newblock Data-driven grasp synthesis—a survey.
\newblock \emph{IEEE Transactions on Robotics 30}, 2 (2014), 289--309.
\newblock \href {https://doi.org/10.1109/TRO.2013.2289018}
  {\path{doi:10.1109/TRO.2013.2289018}}.

\bibitem[BMSR20]{bemana2020xfields}
\textsc{Bemana M., Myszkowski K., Seidel H.-P., Ritschel T.}:
\newblock X-fields: Implicit neural view-, light- and time-image interpolation.
\newblock \emph{ACM Transactions on Graphics (TOG)} (2020).
\newblock URL: \url{http://arxiv.org/abs/2010.00450v1}.

\bibitem[BMT{\etalchar{*}}21]{barron2021mipnerf}
\textsc{Barron J.~T., Mildenhall B., Tancik M., Hedman P., Martin-Brualla R.,
  Srinivasan P.~P.}:
\newblock Mip-nerf: A multiscale representation for anti-aliasing neural
  radiance fields.
\newblock In \emph{Proceedings of the IEEE International Conference on Computer
  Vision (ICCV)} (2021).
\newblock URL: \url{http://arxiv.org/abs/2103.13415v3}.

\bibitem[BPRS18]{baydin2018automatic}
\textsc{Baydin A.~G., Pearlmutter B.~A., Radul A.~A., Siskind J.~M.}:
\newblock Automatic differentiation in machine learning: a survey.
\newblock \emph{Journal of machine learning research 18} (2018).

\bibitem[BPZ{\etalchar{*}}21]{bozic2021neuraldeformationgraphs}
\textsc{Bozic A., Palafox P., Zollhofer M., Thies J., Dai A., Niessner M.}:
\newblock Neural deformation graphs for globally-consistent non-rigid
  reconstruction.
\newblock In \emph{Proceedings of the IEEE/CVF Conference on Computer Vision
  and Pattern Recognition (CVPR)} (2021).
\newblock URL: \url{http://arxiv.org/abs/2012.01451v1}.

\bibitem[BS12]{burley2012physically}
\textsc{Burley B., Studios W. D.~A.}:
\newblock Physically-based shading at disney.
\newblock In \emph{ACM SIGGRAPH} (2012), vol.~2012, vol. 2012, pp.~1--7.

\bibitem[BSM{\etalchar{*}}21a]{bhardwaj2021fast}
\textsc{Bhardwaj M., Sundaralingam B., Mousavian A., Ratliff N., Fox D., Ramos
  F., Boots B.}:
\newblock Fast joint space model-predictive control for reactive manipulation.
\newblock \emph{arXiv preprint arXiv:2104.13542} (2021).

\bibitem[BSM{\etalchar{*}}21b]{bhardwaj2021storm}
\textsc{Bhardwaj M., Sundaralingam B., Mousavian A., Ratliff N., Fox D., Ramos
  F., Boots B.}:
\newblock Storm: An integrated framework for fast joint-space model-predictive
  control for reactive manipulation.
\newblock In \emph{Advances in Neural Information Processing Systems (NeurIPS)}
  (2021), Curran Associates, Inc.
\newblock URL: \url{http://arxiv.org/abs/2104.13542v2}.

\bibitem[BSTPM20a]{bhatnagar2020ipnet}
\textsc{Bhatnagar B.~L., Sminchisescu C., Theobalt C., Pons-Moll G.}:
\newblock Combining implicit function learning and parametric models for 3d
  human reconstruction.
\newblock In \emph{Proceedings of the European Conference on Computer Vision
  (ECCV)} (2020).
\newblock URL: \url{http://arxiv.org/abs/2007.11432v1}.

\bibitem[BSTPM20b]{bhatnagar2020loopreg}
\textsc{Bhatnagar B.~L., Sminchisescu C., Theobalt C., Pons-Moll G.}:
\newblock Loopreg: Self-supervised learning of implicit surface
  correspondences, pose and shape for 3d human mesh registration.
\newblock In \emph{Advances in Neural Information Processing Systems (NeurIPS)}
  (2020), Curran Associates, Inc.
\newblock URL: \url{http://arxiv.org/abs/2010.12447v1}.

\bibitem[BTW21]{bond2020gradient}
\textsc{Bond-Taylor S., Willcocks C.~G.}:
\newblock Gradient origin networks.
\newblock In \emph{International Conference on Learning Representations}
  (2021).
\newblock URL: \url{https://openreview.net/pdf?id=0O_cQfw6uEh}.

\bibitem[BV99]{blanz1999morphable}
\textsc{Blanz V., Vetter T.}:
\newblock A morphable model for the synthesis of 3d faces.
\newblock In \emph{Proceedings of the 26th annual conference on Computer
  graphics and interactive techniques} (1999), pp.~187--194.

\bibitem[BXS{\etalchar{*}}20a]{bi2020neural}
\textsc{Bi S., Xu Z., Srinivasan P., Mildenhall B., Sunkavalli K., Hasan M.,
  Hold-Geoffroy Y., Kriegman D., Ramamoorthi R.}:
\newblock Neural reflectance fields for appearance acquisition.
\newblock \emph{arXiv preprint arXiv:2008.03824} (2020).
\newblock URL: \url{http://arxiv.org/abs/2008.03824v2}.

\bibitem[BXS{\etalchar{*}}20b]{bi2020deep}
\textsc{Bi S., Xu Z., Sunkavalli K., Hasan M., Hold-Geoffroy Y., Kriegman D.,
  Ramamoorthi R.}:
\newblock Deep reflectance volumes: Relightable reconstructions from multi-view
  photometric images.
\newblock In \emph{Proceedings of the European Conference on Computer Vision
  (ECCV)} (2020).
\newblock URL: \url{http://arxiv.org/abs/2007.09892v1}.

\bibitem[BZYsM21]{NeuralPull}
\textsc{Baorui M., Zhizhong H., Yu-shen L., Matthias Z.}:
\newblock Neural-pull: Learning signed distance functions from point clouds by
  learning to pull space onto surfaces.
\newblock In \emph{International Conference on Machine Learning (ICML)} (2021).

\bibitem[CAMNS21]{chatziagapi2021sider}
\textsc{Chatziagapi A., Athar S., Moreno-Noguer F., Samaras D.}:
\newblock Sider: Single-image neural optimization for facial geometric detail
  recovery.
\newblock \emph{arXiv preprint arXiv:2108.05465} (2021).
\newblock URL: \url{http://arxiv.org/abs/2108.05465v1}.

\bibitem[CAPM20]{chibane2020ifnet}
\textsc{Chibane J., Alldieck T., Pons-Moll G.}:
\newblock Implicit functions in feature space for 3d shape reconstruction and
  completion.
\newblock In \emph{Proceedings of the IEEE/CVF Conference on Computer Vision
  and Pattern Recognition (CVPR)} (2020).
\newblock URL: \url{http://arxiv.org/abs/2003.01456v2}.

\bibitem[CbGPW14]{coombes2014neural}
\textsc{Coombes S., beim Graben P., Potthast R., Wright J.}:
\newblock \emph{Neural fields: theory and applications}.
\newblock Springer, 2014.

\bibitem[CBLPM21]{chibane2021srf}
\textsc{Chibane J., Bansal A., Lazova V., Pons-Moll G.}:
\newblock Stereo radiance fields (srf): Learning view synthesis for sparse
  views of novel scenes.
\newblock In \emph{Proceedings of the IEEE/CVF Conference on Computer Vision
  and Pattern Recognition (CVPR)} (2021).
\newblock URL: \url{http://arxiv.org/abs/2104.06935v1}.

\bibitem[CCA{\etalchar{*}}21]{czerkawski2021neuralknitworks}
\textsc{Czerkawski M., Cardona J., Atkinson R., Michie C., Andonovic I.,
  Clemente C., Tachtatzis C.}:
\newblock Neural knitworks: Patched neural implicit representation networks.
\newblock \emph{arXiv preprint arXiv:2109.14406} (2021).
\newblock URL: \url{http://arxiv.org/abs/2109.14406v1}.

\bibitem[CCGC21]{chen2021neuraldefmap}
\textsc{Chen P.~Y., Chiaramonte M., Grinspun E., Carlberg K.}:
\newblock Model reduction for the material point method via learning the
  deformation map and its spatial-temporal gradients.
\newblock URL: \url{http://arxiv.org/abs/2109.12390v1}.

\bibitem[CFG{\etalchar{*}}15]{shapenet2015}
\textsc{Chang A.~X., Funkhouser T., Guibas L., Hanrahan P., Huang Q., Li Z.,
  Savarese S., Savva M., Song S., Su H., Xiao J., Yi L., Yu F.}:
\newblock \emph{{ShapeNet: An Information-Rich 3D Model Repository}}.
\newblock Tech. Rep. arXiv:1512.03012 [cs.GR], Stanford University ---
  Princeton University --- Toyota Technological Institute at Chicago, 2015.

\bibitem[Cha13]{chandrasekhar2013radiative}
\textsc{Chandrasekhar S.}:
\newblock \emph{Radiative transfer}.
\newblock Courier Corporation, 2013.

\bibitem[CKVL21]{chen2021visualselfmodelingrobot}
\textsc{Chen B., Kwiatkowski R., Vondrick C., Lipson H.}:
\newblock Full-body visual self-modeling of robot morphologies.
\newblock URL: \url{http://arxiv.org/abs/2111.06389v2}.

\bibitem[CLCO{\etalchar{*}}21]{chen2021mocoflow}
\textsc{Chen X., Li W., Cohen-Or D., Mitra N.~J., Chen B.}:
\newblock Moco-flow: Neural motion consensus flow for dynamic humans in
  stationary monocular cameras.
\newblock \emph{arXiv preprint arXiv:2106.04477} (2021).
\newblock URL: \url{http://arxiv.org/abs/2106.04477v1}.

\bibitem[CLI{\etalchar{*}}20]{chabra2020deepls}
\textsc{Chabra R., Lenssen J.~E., Ilg E., Schmidt T., Straub J., Lovegrove S.,
  Newcombe R.}:
\newblock Deep local shapes: Learning local sdf priors for detailed 3d
  reconstruction.
\newblock In \emph{Proceedings of the European Conference on Computer Vision
  (ECCV)} (2020).
\newblock URL: \url{http://arxiv.org/abs/2003.10983v3}.

\bibitem[CLW21a]{chen2021liif}
\textsc{Chen Y., Liu S., Wang X.}:
\newblock Learning continuous image representation with local implicit image
  function.
\newblock In \emph{Proceedings of the IEEE/CVF Conference on Computer Vision
  and Pattern Recognition (CVPR)} (2021).
\newblock URL: \url{http://arxiv.org/abs/2012.09161v2}.

\bibitem[CLW21b]{chen2021learning}
\textsc{Chen Y., Liu S., Wang X.}:
\newblock Learning continuous image representation with local implicit image
  function.
\newblock In \emph{Proc. CVPR} (2021), pp.~8628--8638.

\bibitem[CMK05]{chaimowicz2005_robotswarm}
\textsc{Chaimowicz L., Michael N., Kumar V.}:
\newblock Controlling swarms of robots using interpolated implicit functions.
\newblock In \emph{Proceedings of the 2005 IEEE International Conference on
  Robotics and Automation} (2005), pp.~2487--2492.
\newblock \href {https://doi.org/10.1109/ROBOT.2005.1570486}
  {\path{doi:10.1109/ROBOT.2005.1570486}}.

\bibitem[CMK{\etalchar{*}}21]{chan2021pigan}
\textsc{Chan E.~R., Monteiro M., Kellnhofer P., Wu J., Wetzstein G.}:
\newblock pi-gan: Periodic implicit generative adversarial networks for
  3d-aware image synthesis.
\newblock In \emph{Proceedings of the IEEE/CVF Conference on Computer Vision
  and Pattern Recognition (CVPR)} (2021).
\newblock URL: \url{http://arxiv.org/abs/2012.00926v2}.

\bibitem[CMPM20]{chibane2020ndf}
\textsc{Chibane J., Mir A., Pons-Moll G.}:
\newblock Neural unsigned distance fields for implicit function learning.
\newblock In \emph{Advances in Neural Information Processing Systems (NeurIPS)}
  (2020), Curran Associates, Inc.
\newblock URL: \url{http://arxiv.org/abs/2010.13938v1}.

\bibitem[CP21]{costain2021towards}
\textsc{Costain T.~W., Prisacariu V.~A.}:
\newblock Towards generalising neural implicit representations.
\newblock \emph{arXiv preprint arXiv:2101.12690} (2021).
\newblock URL: \url{http://arxiv.org/abs/2101.12690v2}.

\bibitem[CPA{\etalchar{*}}21]{corona2021smplicit}
\textsc{Corona E., Pumarola A., Alenya G., Pons-Moll G., Moreno-Noguer F.}:
\newblock Smplicit: Topology-aware generative model for clothed people.
\newblock In \emph{Proceedings of the IEEE/CVF Conference on Computer Vision
  and Pattern Recognition (CVPR)} (2021).
\newblock URL: \url{http://arxiv.org/abs/2103.06871v2}.

\bibitem[CPG21]{chitta2021neat}
\textsc{Chitta K., Prakash A., Geiger A.}:
\newblock Neat: Neural attention fields for end-to-end autonomous driving.
\newblock In \emph{Proceedings of the IEEE International Conference on Computer
  Vision (ICCV)} (2021).
\newblock URL: \url{http://arxiv.org/abs/2109.04456v1}.

\bibitem[CPM20]{chibane2020ifnettexture}
\textsc{Chibane J., Pons-Moll G.}:
\newblock Implicit feature networks for texture completion from partial 3d
  data.
\newblock In \emph{Proceedings of the European Conference on Computer Vision
  (ECCV)} (2020).
\newblock URL: \url{http://arxiv.org/abs/2009.09458v1}.

\bibitem[CRW{\etalchar{*}}20]{chaudhuri2020learning}
\textsc{Chaudhuri S., Ritchie D., Wu J., Xu K., Zhang H.}:
\newblock Learning generative models of 3d structures.
\newblock In \emph{Computer Graphics Forum} (2020), vol.~39, Wiley Online
  Library, pp.~643--666.

\bibitem[CT82]{cook1982reflectance}
\textsc{Cook R.~L., Torrance K.~E.}:
\newblock A reflectance model for computer graphics.
\newblock \emph{ACM Transactions on Graphics (ToG) 1}, 1 (1982), 7--24.

\bibitem[CTT{\etalchar{*}}21]{chiang2021stylizing}
\textsc{Chiang P.-Z., Tsai M.-S., Tseng H.-Y., sheng Lai W., Chiu W.-C.}:
\newblock Stylizing 3d scene via implicit representation and
  hypernetwork/meta-learning.
\newblock \emph{arXiv preprint arXiv:2105.13016} (2021).
\newblock URL: \url{http://arxiv.org/abs/2105.13016v2}.

\bibitem[CWP21]{chen2021directposenet}
\textsc{Chen S., Wang Z., Prisacariu V.}:
\newblock Direct-posenet: Absolute pose regression with photometric
  consistency.
\newblock \emph{arXiv preprint arXiv:2104.04073} (2021).
\newblock URL: \url{http://arxiv.org/abs/2104.04073v1}.

\bibitem[CXZ{\etalchar{*}}21]{chen2021mvsnerf}
\textsc{Chen A., Xu Z., Zhao F., Zhang X., Xiang F., Yu J., Su H.}:
\newblock Mvsnerf: Fast generalizable radiance field reconstruction from
  multi-view stereo.
\newblock In \emph{Proceedings of the IEEE International Conference on Computer
  Vision (ICCV)} (2021).
\newblock URL: \url{http://arxiv.org/abs/2103.15595v2}.

\bibitem[CZ19]{chen2019imnet}
\textsc{Chen Z., Zhang H.}:
\newblock Learning implicit fields for generative shape modeling.
\newblock In \emph{Proceedings of the IEEE/CVF Conference on Computer Vision
  and Pattern Recognition (CVPR)} (2019).
\newblock URL: \url{http://arxiv.org/abs/1812.02822v5}.

\bibitem[CZB{\etalchar{*}}21]{chen2021snarf}
\textsc{Chen X., Zheng Y., Black M.~J., Hilliges O., Geiger A.}:
\newblock Snarf: Differentiable forward skinning for animating non-rigid neural
  implicit shapes.
\newblock In \emph{Proceedings of the IEEE International Conference on Computer
  Vision (ICCV)} (2021).
\newblock URL: \url{http://arxiv.org/abs/2104.03953v1}.

\bibitem[CZG{\etalchar{*}}21]{chen2021mdif}
\textsc{Chen Z., Zhang Y., Genova K., Fanello S., Bouaziz S., Haene C., Du R.,
  Keskin C., Funkhouser T., Tang D.}:
\newblock Multiresolution deep implicit functions for 3d shape representation.
\newblock In \emph{Proceedings of the IEEE International Conference on Computer
  Vision (ICCV)} (2021).
\newblock URL: \url{http://arxiv.org/abs/2109.05591v2}.

\bibitem[CZK{\etalchar{*}}21]{chen2021animatable}
\textsc{Chen J., Zhang Y., Kang D., Zhe X., Bao L., Jia X., Lu H.}:
\newblock Animatable neural radiance fields from monocular rgb videos.
\newblock In \emph{Proceedings of the IEEE International Conference on Computer
  Vision (ICCV)} (2021).
\newblock URL: \url{http://arxiv.org/abs/2106.13629v2}.

\bibitem[CZL{\etalchar{*}}21]{chen2021hanerf}
\textsc{Chen X., Zhang Q., Li X., Chen Y., Feng Y., Wang X., Wang J.}:
\newblock Hallucinated neural radiance fields in the wild.
\newblock URL: \url{http://arxiv.org/abs/2111.15246v2}.

\bibitem[CZY{\etalchar{*}}21]{chen2021nefnet}
\textsc{Chen J., Zheng X., Yu H., Chen D.~Z., Wu J.}:
\newblock Electrocardio panorama: Synthesizing new ecg views with
  self-supervision.
\newblock In \emph{Proceedings of the Thirtieth International Joint Conference
  on Artificial Intelligence (IJCAI)} (2021), International Joint Conferences
  on Artificial Intelligence Organization.
\newblock URL: \url{http://arxiv.org/abs/2105.06293v1}.

\bibitem[DBS{\etalchar{*}}21]{devries2021unconstrained}
\textsc{DeVries T., Bautista M.~A., Srivastava N., Taylor G.~W., Susskind
  J.~M.}:
\newblock Unconstrained scene generation with locally conditioned radiance
  fields.
\newblock In \emph{Proceedings of the IEEE International Conference on Computer
  Vision (ICCV)} (2021).
\newblock URL: \url{http://arxiv.org/abs/2104.00670v1}.

\bibitem[DCTS21]{du2021gem}
\textsc{Du Y., Collins M.~K., Tenenbaum B.~J., Sitzmann V.}:
\newblock Learning signal-agnostic manifolds of neural fields.
\newblock In \emph{Proc. NeurIPS} (2021).

\bibitem[DGA{\etalchar{*}}21]{dupont2021coin}
\textsc{Dupont E., Golinski A., Alizadeh M., Teh Y.~W., Doucet A.}:
\newblock Coin: Compression with implicit neural representations.
\newblock In \emph{International Conference on Learning Representations}
  (2021).
\newblock URL: \url{http://arxiv.org/abs/2103.03123v2}.

\bibitem[DGF{\etalchar{*}}19]{deprelle2019learning}
\textsc{Deprelle T., Groueix T., Fisher M., Kim V.~G., Russell B.~C., Aubry
  M.}:
\newblock Learning elementary structures for 3d shape generation and matching.
\newblock In \emph{Proceedings of the IEEE International Conference on Computer
  Vision (ICCV)} (2019).
\newblock URL: \url{http://arxiv.org/abs/1908.04725v2}.

\bibitem[DGY{\etalchar{*}}20]{deng2020cvxnet}
\textsc{Deng B., Genova K., Yazdani S., Bouaziz S., Hinton G., Tagliasacchi
  A.}:
\newblock Cvxnet: Learnable convex decomposition.
\newblock In \emph{Proc. CVPR} (2020), pp.~31--44.

\bibitem[DHY{\etalchar{*}}21]{deng2021foveated}
\textsc{Deng N., He Z., Ye J., Chakravarthula P., Yang X., Sun Q.}:
\newblock Foveated neural radiance fields for real-time and egocentric virtual
  reality.
\newblock \emph{arXiv preprint arXiv:2103.16365} (2021).
\newblock URL: \url{http://arxiv.org/abs/2103.16365v1}.

\bibitem[DI21]{derksen2021snerf}
\textsc{Derksen D., Izzo D.}:
\newblock Shadow neural radiance fields for multi-view satellite
  photogrammetry.
\newblock In \emph{Proceedings of the IEEE/CVF Conference on Computer Vision
  and Pattern Recognition (CVPR)} (2021).
\newblock URL: \url{http://arxiv.org/abs/2104.09877v1}.

\bibitem[DLD{\etalchar{*}}21]{deng2021vectorneurons}
\textsc{Deng C., Litany O., Duan Y., Poulenard A., Tagliasacchi A., Guibas L.}:
\newblock Vector neurons: A general framework for so(3)-equivariant networks.
\newblock \emph{arXiv preprint arXiv:2104.12229} (2021).
\newblock URL: \url{http://arxiv.org/abs/2104.12229v1}.

\bibitem[DLJ{\etalchar{*}}20]{deng2020nasa}
\textsc{Deng B., Lewis J., Jeruzalski T., Pons-Moll G., Hinton G., Norouzi M.,
  Tagliasacchi A.}:
\newblock Nasa: Neural articulated shape approximation.
\newblock In \emph{Proceedings of the European Conference on Computer Vision
  (ECCV)} (2020).
\newblock URL: \url{http://arxiv.org/abs/1912.03207v4}.

\bibitem[DLZR21]{deng2021dsnerf}
\textsc{Deng K., Liu A., Zhu J.-Y., Ramanan D.}:
\newblock Depth-supervised nerf: Fewer views and faster training for free.
\newblock \emph{arXiv preprint arXiv:2107.02791} (2021).
\newblock URL: \url{http://arxiv.org/abs/2107.02791v1}.

\bibitem[DNJ21]{davies2021on}
\textsc{Davies T., Nowrouzezahrai D., Jacobson A.}:
\newblock On the effectiveness of weight-encoded neural implicit 3d shapes.
\newblock In \emph{International Conference on Machine Learning (ICML)} (2021),
  PMLR.
\newblock URL: \url{http://arxiv.org/abs/2009.09808v3}.

\bibitem[DPS{\etalchar{*}}18]{dumoulin2018feature}
\textsc{Dumoulin V., Perez E., Schucher N., Strub F., de~Vries H., Courville
  A., Bengio Y.}:
\newblock Feature-wise transformations.
\newblock \emph{Distill 3}, 7 (2018), e11.

\bibitem[DVPL21]{dehesa2021gfnn}
\textsc{Dehesa J., Vidler A., Padget J., Lutteroth C.}:
\newblock Grid-functioned neural networks.
\newblock In \emph{Proceedings of the 38th International Conference on Machine
  Learning} (18--24 Jul 2021), Meila M., Zhang T., (Eds.), vol.~139 of
  \emph{Proceedings of Machine Learning Research}, PMLR, pp.~2559--2567.
\newblock URL: \url{https://proceedings.mlr.press/v139/dehesa21a.html}.

\bibitem[DW85]{dippe1985antialiasing}
\textsc{Dipp{\'e} M.~A., Wold E.~H.}:
\newblock Antialiasing through stochastic sampling.
\newblock In \emph{Proceedings of the 12th annual conference on Computer
  graphics and interactive techniques} (1985), pp.~69--78.

\bibitem[DZY{\etalchar{*}}21]{du2021nerflow}
\textsc{Du Y., Zhang Y., Yu H.-X., Tenenbaum J.~B., Wu J.}:
\newblock Neural radiance flow for 4d view synthesis and video processing.
\newblock In \emph{Proceedings of the IEEE International Conference on Computer
  Vision (ICCV)} (2021).
\newblock URL: \url{http://arxiv.org/abs/2012.09790v2}.

\bibitem[EPF14]{eigen2014depth}
\textsc{Eigen D., Puhrsch C., Fergus R.}:
\newblock Depth map prediction from a single image using a multi-scale deep
  network.
\newblock \emph{arXiv preprint arXiv:1406.2283} (2014).

\bibitem[ERR{\etalchar{*}}17]{engel2017nsynth}
\textsc{Engel J., Resnick C., Roberts A., Dieleman S., Norouzi M., Eck D.,
  Simonyan K.}:
\newblock Neural audio synthesis of musical notes with wavenet autoencoders.
\newblock In \emph{Proceedings of the 34th International Conference on Machine
  Learning - Volume 70} (2017), ICML'17, JMLR.org, p.~1068–1077.

\bibitem[Eva15]{evans2015learning}
\textsc{Evans A.}:
\newblock Learning from failure: a survey of promising, unconventional and
  mostly abandoned renderers for ‘dreams ps4’, a geometrically dense,
  painterly ugc game.
\newblock \emph{Advances in Real-Time Rendering in Games. MediaMolecule,
  SIGGRAPH} (2015).

\bibitem[FAL17]{finn2017model}
\textsc{Finn C., Abbeel P., Levine S.}:
\newblock Model-agnostic meta-learning for fast adaptation of deep networks.
\newblock In \emph{International Conference on Machine Learning} (2017), PMLR,
  pp.~1126--1135.

\bibitem[FC19]{frankle2018lottery}
\textsc{Frankle J., Carbin M.}:
\newblock The lottery ticket hypothesis: Finding sparse, trainable neural
  networks.
\newblock \emph{Proc. ICLR} (2019).

\bibitem[Flo03]{floater2003mean}
\textsc{Floater M.~S.}:
\newblock Mean value coordinates.
\newblock \emph{Computer aided geometric design 20}, 1 (2003), 19--27.

\bibitem[FLS65]{feynman1965feynman}
\textsc{Feynman R.~P., Leighton R.~B., Sands M.}:
\newblock The feynman lectures on physics; vol. i.
\newblock \emph{American Journal of Physics 33}, 9 (1965), 750--752.

\bibitem[FSSV07]{florez2007efficient}
\textsc{Fl{\'o}rez J., Sbert M., Sainz M.~A., Veh{\'\i} J.}:
\newblock Efficient ray tracing using interval analysis.
\newblock In \emph{International Conference on Parallel Processing and Applied
  Mathematics} (2007), Springer, pp.~1351--1360.

\bibitem[FW21]{fu2021multiscene}
\textsc{Fu B., Wang Z.}:
\newblock Multi-scene representation learning with neural radiance fields.
\newblock \emph{Journal of Physics: Conference Series 1880}, 1 (2021), 012034.
\newblock URL: \url{https://doi.org/10.1088/1742-6596/1880/1/012034}, \href
  {https://doi.org/10.1088/1742-6596/1880/1/012034}
  {\path{doi:10.1088/1742-6596/1880/1/012034}}.

\bibitem[G{\etalchar{*}}89]{griewank1989automatic}
\textsc{Griewank A., et~al.}:
\newblock On automatic differentiation.
\newblock \emph{Mathematical Programming: recent developments and applications
  6}, 6 (1989), 83--107.

\bibitem[GCL{\etalchar{*}}21]{guo2021adnerf}
\textsc{Guo Y., Chen K., Liang S., Liu Y.-J., Bao H., Zhang J.}:
\newblock Ad-nerf: Audio driven neural radiance fields for talking head
  synthesis.
\newblock In \emph{Proceedings of the IEEE International Conference on Computer
  Vision (ICCV)} (2021).
\newblock URL: \url{http://arxiv.org/abs/2103.11078v3}.

\bibitem[GCM{\etalchar{*}}21]{gao2021objectfolder}
\textsc{Gao R., Chang Y.-Y., Mall S., Fei-Fei L., Wu J.}:
\newblock Objectfolder: A dataset of objects with implicit visual, auditory,
  and tactile representations.
\newblock In \emph{Proceedings of the Conference on Robot Learning (CoRL)}
  (2021).
\newblock URL: \url{http://arxiv.org/abs/2109.07991v2}.

\bibitem[GCS{\etalchar{*}}20]{genova2020ldif}
\textsc{Genova K., Cole F., Sud A., Sarna A., Funkhouser T.}:
\newblock Local deep implicit functions for 3d shape.
\newblock In \emph{Proceedings of the IEEE/CVF Conference on Computer Vision
  and Pattern Recognition (CVPR)} (2020).
\newblock URL: \url{http://arxiv.org/abs/1912.06126v2}.

\bibitem[GFK{\etalchar{*}}18a]{groueix20183dcoded}
\textsc{Groueix T., Fisher M., Kim V.~G., Russell B.~C., Aubry M.}:
\newblock 3d-coded : 3d correspondences by deep deformation, 2018.

\bibitem[GFK{\etalchar{*}}18b]{groueix2018atlasnet}
\textsc{Groueix T., Fisher M., Kim V.~G., Russell B.~C., Aubry M.}:
\newblock Atlasnet: A papier-mache approach to learning 3d surface generation.
\newblock In \emph{Proceedings of the IEEE/CVF Conference on Computer Vision
  and Pattern Recognition (CVPR)} (2018).
\newblock URL: \url{http://arxiv.org/abs/1802.05384v3}.

\bibitem[GFWF20]{guo2020osfs}
\textsc{Guo M., Fathi A., Wu J., Funkhouser T.}:
\newblock Object-centric neural scene rendering.
\newblock \emph{arXiv preprint arXiv:2012.08503} (2020).
\newblock URL: \url{http://arxiv.org/abs/2012.08503v1}.

\bibitem[GG21]{giebenhain2021airnets}
\textsc{Giebenhain S., Goldluecke B.}:
\newblock Air-nets: An attention-based framework for locally conditioned
  implicit representations.
\newblock In \emph{2021 International Conference on 3D Vision (3DV)} (2021),
  IEEE.

\bibitem[GGH02]{gu2002geometry}
\textsc{Gu X., Gortler S.~J., Hoppe H.}:
\newblock Geometry images.
\newblock In \emph{Proceedings of the 29th annual conference on Computer
  graphics and interactive techniques} (2002), pp.~355--361.

\bibitem[GGPP20]{galin2020segment}
\textsc{Galin E., Gu{\'e}rin E., Paris A., Peytavie A.}:
\newblock Segment tracing using local lipschitz bounds.
\newblock In \emph{Computer Graphics Forum} (2020), vol.~39, Wiley Online
  Library, pp.~545--554.

\bibitem[GKJ{\etalchar{*}}21]{garbin2021fastnerf}
\textsc{Garbin S.~J., Kowalski M., Johnson M., Shotton J., Valentin J.}:
\newblock Fastnerf: High-fidelity neural rendering at 200fps.
\newblock \emph{arXiv preprint arXiv:2103.10380} (2021).
\newblock URL: \url{http://arxiv.org/abs/2103.10380v2}.

\bibitem[GN98]{gargan1998approximating}
\textsc{Gargan D., Neelamkavil F.}:
\newblock Approximating reflectance functions using neural networks.
\newblock \emph{Computer Graphics Forum} (1998), 23--34.

\bibitem[GPAM{\etalchar{*}}14]{goodfellow2014generative}
\textsc{Goodfellow I., Pouget-Abadie J., Mirza M., Xu B., Warde-Farley D.,
  Ozair S., Courville A., Bengio Y.}:
\newblock Generative adversarial nets.
\newblock \emph{Advances in neural information processing systems 27} (2014).

\bibitem[Gre86]{greene1986environment}
\textsc{Greene N.}:
\newblock Environment mapping and other applications of world projections.
\newblock \emph{IEEE computer graphics and Applications 6}, 11 (1986), 21--29.

\bibitem[Gre03]{green2003spherical}
\textsc{Green R.}:
\newblock Spherical harmonic lighting: The gritty details.
\newblock In \emph{Archives of the game developers conference} (2003), vol.~56,
  p.~4.

\bibitem[GSKH21]{gao2021dynamic}
\textsc{Gao C., Saraf A., Kopf J., Huang J.-B.}:
\newblock Dynamic view synthesis from dynamic monocular video.
\newblock \emph{arXiv preprint arXiv:2105.06468} (2021).
\newblock URL: \url{http://arxiv.org/abs/2105.06468v1}.

\bibitem[GSL{\etalchar{*}}20]{gao2020portraitnerf}
\textsc{Gao C., Shih Y., Lai W.-S., Liang C.-K., Huang J.-B.}:
\newblock Portrait neural radiance fields from a single image.
\newblock \emph{arXiv preprint arXiv:2012.05903} (2020).
\newblock URL: \url{http://arxiv.org/abs/2012.05903v2}.

\bibitem[GSRN21]{granskog2021neural}
\textsc{Granskog J., Schnabel T.~N., Rousselle F., Nov{\'a}k J.}:
\newblock Neural scene graph rendering.
\newblock \emph{ACM Transactions on Graphics (TOG) 40}, 4 (2021), 1--11.

\bibitem[GTZN21]{gafni2021nerface}
\textsc{Gafni G., Thies J., Zollhofer M., Niessner M.}:
\newblock Dynamic neural radiance fields for monocular 4d facial avatar
  reconstruction.
\newblock In \emph{Proceedings of the IEEE/CVF Conference on Computer Vision
  and Pattern Recognition (CVPR)} (2021).
\newblock URL: \url{http://arxiv.org/abs/2012.03065v1}.

\bibitem[GW20]{galanti2020modularity}
\textsc{Galanti T., Wolf L.}:
\newblock On the modularity of hypernetworks.
\newblock \emph{Proc. NeurIPS} (2020).

\bibitem[GYH{\etalchar{*}}20]{gropp2020igr}
\textsc{Gropp A., Yariv L., Haim N., Atzmon M., Lipman Y.}:
\newblock Implicit geometric regularization for learning shapes.
\newblock In \emph{Proceedings of the IEEE/CVF Conference on Computer Vision
  and Pattern Recognition (CVPR)} (2020).
\newblock URL: \url{http://arxiv.org/abs/2002.10099v2}.

\bibitem[HAESB20]{hao2020dualsdf}
\textsc{Hao Z., Averbuch-Elor H., Snavely N., Belongie S.}:
\newblock Dualsdf: Semantic shape manipulation using a two-level
  representation.
\newblock In \emph{Proceedings of the IEEE/CVF Conference on Computer Vision
  and Pattern Recognition (CVPR)} (2020).
\newblock URL: \url{http://arxiv.org/abs/2004.02869v1}.

\bibitem[Har96]{hart1996sphere}
\textsc{Hart J.~C.}:
\newblock Sphere tracing: A geometric method for the antialiased ray tracing of
  implicit surfaces.
\newblock \emph{The Visual Computer 12}, 10 (1996), 527--545.

\bibitem[HAS21]{huang2021a}
\textsc{Huang X., Alkhalifah T., Song C.}:
\newblock \emph{A modified physics-informed neural network with positional
  encoding}.
\newblock 2021, pp.~2480--2484.
\newblock URL:
  \url{https://library.seg.org/doi/abs/10.1190/segam2021-3584127.1}, \href
  {https://doi.org/10.1190/segam2021-3584127.1}
  {\path{doi:10.1190/segam2021-3584127.1}}.

\bibitem[HB17]{huang2017arbitrary}
\textsc{Huang X., Belongie S.}:
\newblock Arbitrary style transfer in real-time with adaptive instance
  normalization.
\newblock In \emph{Proc. ICCV} (2017), pp.~1501--1510.

\bibitem[HCJS20]{he2020geopifu}
\textsc{He T., Collomosse J., Jin H., Soatto S.}:
\newblock Geo-pifu: Geometry and pixel aligned implicit functions for
  single-view human reconstruction.
\newblock In \emph{Advances in Neural Information Processing Systems (NeurIPS)}
  (2020), Curran Associates, Inc.
\newblock URL: \url{http://arxiv.org/abs/2006.08072v2}.

\bibitem[HCZ21]{hadadan2021neural}
\textsc{Hadadan S., Chen S., Zwicker M.}:
\newblock Neural radiosity.
\newblock \emph{arXiv preprint arXiv:2105.12319} (2021).
\newblock URL: \url{http://arxiv.org/abs/2105.12319v1}.

\bibitem[HDL16]{ha2016hypernetworks}
\textsc{Ha D., Dai A., Le Q.}:
\newblock Hypernetworks.
\newblock In \emph{International Conference on Learning Representations}
  (2016).

\bibitem[HECI20]{hani2020corn}
\textsc{Hani N., Engin S., Chao J.-J., Isler V.}:
\newblock Continuous object representation networks: Novel view synthesis
  without target view supervision.
\newblock In \emph{Advances in Neural Information Processing Systems (NeurIPS)}
  (2020), Curran Associates, Inc.
\newblock URL: \url{http://arxiv.org/abs/2007.15627v2}.

\bibitem[HIL{\etalchar{*}}20]{halimi2020greater}
\textsc{Halimi O., Imanuel I., Litany O., Trappolini G., Rodolà E., Guibas L.,
  Kimmel R.}:
\newblock The whole is greater than the sum of its nonrigid parts, 2020.

\bibitem[HJKK21]{hu2021when}
\textsc{Hu Z., Jagtap A.~D., Karniadakis G.~E., Kawaguchi K.}:
\newblock When do extended physics-informed neural networks (xpinns) improve
  generalization?
\newblock \emph{arXiv preprint arXiv:2109.09444} (2021).
\newblock URL: \url{http://arxiv.org/abs/2109.09444v2}.

\bibitem[HLB21]{henderson2021unsupervised}
\textsc{Henderson P., Lampert C.~H., Bickel B.}:
\newblock Unsupervised video prediction from a single frame by estimating 3d
  dynamic scene structure.
\newblock \emph{arXiv preprint arXiv:2106.09051} (2021).
\newblock URL: \url{http://arxiv.org/abs/2106.09051v1}.

\bibitem[HMBL21]{hao2021gancraft}
\textsc{Hao Z., Mallya A., Belongie S., Liu M.-Y.}:
\newblock Gancraft: Unsupervised 3d neural rendering of minecraft worlds.
\newblock In \emph{Proceedings of the IEEE International Conference on Computer
  Vision (ICCV)} (2021).
\newblock URL: \url{http://arxiv.org/abs/2104.07659v1}.

\bibitem[HMR20]{henzler2020neuraltexture}
\textsc{Henzler P., Mitra N.~J., Ritschel T.}:
\newblock Learning a neural 3d texture space from 2d exemplars.
\newblock In \emph{Proceedings of the IEEE/CVF Conference on Computer Vision
  and Pattern Recognition (CVPR)} (2020).
\newblock URL: \url{http://arxiv.org/abs/1912.04158v2}.

\bibitem[HMZ{\etalchar{*}}21]{huh2021low}
\textsc{Huh M., Mobahi H., Zhang R., Cheung B., Agrawal P., Isola P.}:
\newblock The low-rank simplicity bias in deep networks.
\newblock \emph{arXiv preprint arXiv:2103.10427} (2021).

\bibitem[Hor70]{horn1970shape}
\textsc{Horn B.~K.}:
\newblock Shape from shading: A method for obtaining the shape of a smooth
  opaque object from one view.

\bibitem[HPG{\etalchar{*}}21]{hertz2021sape}
\textsc{Hertz A., Perel O., Giryes R., Sorkine-Hornung O., Cohen-Or D.}:
\newblock Sape: Spatially-adaptive progressive encoding for neural
  optimization.
\newblock \emph{arXiv preprint arXiv:2104.09125} (2021).
\newblock URL: \url{http://arxiv.org/abs/2104.09125v2}.

\bibitem[HRL{\etalchar{*}}21]{henzler2021unsupervised}
\textsc{Henzler P., Reizenstein J., Labatut P., Shapovalov R., Ritschel T.,
  Vedaldi A., Novotny D.}:
\newblock Unsupervised learning of 3d object categories from videos in the
  wild.
\newblock In \emph{Proceedings of the IEEE/CVF Conference on Computer Vision
  and Pattern Recognition (CVPR)} (2021).
\newblock URL: \url{http://arxiv.org/abs/2103.16552v1}.

\bibitem[HSC21]{hsu2021omninerf}
\textsc{Hsu C.-Y., Sun C., Chen H.-T.}:
\newblock Moving in a 360 world: Synthesizing panoramic parallaxes from a
  single panorama.
\newblock \emph{arXiv preprint arXiv:2106.10859} (2021).
\newblock URL: \url{http://arxiv.org/abs/2106.10859v1}.

\bibitem[HSM{\etalchar{*}}21]{hedman2021snerg}
\textsc{Hedman P., Srinivasan P.~P., Mildenhall B., Barron J.~T., Debevec P.}:
\newblock Baking neural radiance fields for real-time view synthesis.
\newblock \emph{arXiv preprint arXiv:2103.14645} (2021).
\newblock URL: \url{http://arxiv.org/abs/2103.14645v1}.

\bibitem[HXL{\etalchar{*}}20]{huang2020arch}
\textsc{Huang Z., Xu Y., Lassner C., Li H., Tung T.}:
\newblock Arch: Animatable reconstruction of clothed humans.
\newblock In \emph{Proceedings of the IEEE/CVF Conference on Computer Vision
  and Pattern Recognition (CVPR)} (2020).
\newblock URL: \url{http://arxiv.org/abs/2004.04572v2}.

\bibitem[HXS{\etalchar{*}}21]{he2021arch++}
\textsc{He T., Xu Y., Saito S., Soatto S., Tung T.}:
\newblock Arch++: Animation-ready clothed human reconstruction revisited.
\newblock In \emph{Proceedings of the IEEE International Conference on Computer
  Vision (ICCV)} (2021).
\newblock URL: \url{http://arxiv.org/abs/2108.07845v1}.

\bibitem[HZ03]{hartley2003mvg}
\textsc{Hartley R., Zisserman A.}:
\newblock \emph{Multiple View Geometry in Computer Vision}, 2~ed.
\newblock Cambridge University Press, New York, NY, USA, 2003.

\bibitem[HZJ{\etalchar{*}}21]{hong2021stereopifu}
\textsc{Hong Y., Zhang J., Jiang B., Guo Y., Liu L., Bao H.}:
\newblock Stereopifu: Depth aware clothed human digitization via stereo vision.
\newblock In \emph{Proceedings of the IEEE/CVF Conference on Computer Vision
  and Pattern Recognition (CVPR)} (2021).
\newblock URL: \url{http://arxiv.org/abs/2104.05289v2}.

\bibitem[HZRS16]{he2016resnet}
\textsc{He K., Zhang X., Ren S., Sun J.}:
\newblock Deep residual learning for image recognition.
\newblock In \emph{Proceedings of the IEEE Conference on Computer Vision and
  Pattern Recognition (CVPR)} (2016), pp.~770--778.

\bibitem[IAKG21]{ichnowski2021dexnerf}
\textsc{Ichnowski J., Avigal Y., Kerr J., Goldberg K.}:
\newblock Dex-nerf: Using a neural radiance field to grasp transparent objects.
\newblock URL: \url{http://arxiv.org/abs/2110.14217v1}.

\bibitem[IG21]{izzo2021geodesynets}
\textsc{Izzo D., Gomez P.}:
\newblock Geodesy of irregular small bodies via neural density fields:
  geodesynets.
\newblock \emph{arXiv preprint arXiv:2105.13031} (2021).
\newblock URL: \url{http://arxiv.org/abs/2105.13031v1}.

\bibitem[IH81]{ikeuchi1981numerical}
\textsc{Ikeuchi K., Horn B.~K.}:
\newblock Numerical shape from shading and occluding boundaries.
\newblock \emph{Artificial intelligence 17}, 1-3 (1981), 141--184.

\bibitem[IJC{\etalchar{*}}97]{iso1997virtual}
\textsc{ISO V., JTC I., Consortium V., et~al.}:
\newblock The virtual reality modeling language.
\newblock \emph{Part 2} (1997), 14772--2.

\bibitem[ILK21]{ibing20213d}
\textsc{Ibing M., Lim I., Kobbelt L.}:
\newblock 3d shape generation with grid-based implicit functions.
\newblock In \emph{Proceedings of the IEEE/CVF Conference on Computer Vision
  and Pattern Recognition (CVPR)} (June 2021), pp.~13559--13568.

\bibitem[Isi21]{isik2021neural}
\textsc{Isik B.}:
\newblock Neural 3d scene compression via model compression.
\newblock \emph{arXiv preprint arXiv:2105.03120} (2021).
\newblock URL: \url{http://arxiv.org/abs/2105.03120v1}.

\bibitem[IZZE17]{isola2017image}
\textsc{Isola P., Zhu J.-Y., Zhou T., Efros A.~A.}:
\newblock Image-to-image translation with conditional adversarial networks.
\newblock In \emph{Proc. CVPR} (2017), pp.~1125--1134.

\bibitem[JA21]{jang2021codenerf}
\textsc{Jang W., Agapito L.}:
\newblock Codenerf: Disentangled neural radiance fields for object categories.
\newblock In \emph{Proceedings of the IEEE International Conference on Computer
  Vision (ICCV)} (2021).
\newblock URL: \url{http://arxiv.org/abs/2109.01750v1}.

\bibitem[JAC{\etalchar{*}}21]{jeong2021selfcalibrating}
\textsc{Jeong Y., Ahn S., Choy C., Anandkumar A., Cho M., Park J.}:
\newblock Self-calibrating neural radiance fields.
\newblock In \emph{Proceedings of the IEEE International Conference on Computer
  Vision (ICCV)} (2021).

\bibitem[JEA{\etalchar{*}}20]{jiang2020meshfreeflownet}
\textsc{Jiang C.~M., Esmaeilzadeh S., Azizzadenesheli K., Kashinath K., Mustafa
  M., Tchelepi H.~A., Marcus P., Prabhat, Anandkumar A.}:
\newblock Meshfreeflownet: A physics-constrained deep continuous space-time
  super-resolution framework.
\newblock URL: \url{http://arxiv.org/abs/2005.01463v2}.

\bibitem[JGH18]{jacot2018neural}
\textsc{Jacot A., Gabriel F., Hongler C.}:
\newblock Neural tangent kernel: Convergence and generalization in neural
  networks.
\newblock \emph{arXiv preprint arXiv:1806.07572} (2018).

\bibitem[JK20]{jagtap2020extended}
\textsc{Jagtap A.~D., Karniadakis G.~E.}:
\newblock Extended physics-informed neural networks (xpinns): A generalized
  space-time domain decomposition based deep learning framework for nonlinear
  partial differential equations.
\newblock \emph{Communications in Computational Physics 28}, 5 (2020),
  2002--2041.

\bibitem[JMB{\etalchar{*}}21]{jain2021dreamfields}
\textsc{Jain A., Mildenhall B., Barron J.~T., Abbeel P., Poole B.}:
\newblock Zero-shot text-guided object generation with dream fields.

\bibitem[JMD{\etalchar{*}}07]{joshi2007harmonic}
\textsc{Joshi P., Meyer M., DeRose T., Green B., Sanocki T.}:
\newblock Harmonic coordinates for character articulation.
\newblock \emph{ACM Transactions on Graphics (TOG) 26}, 3 (2007), 71--es.

\bibitem[JMdB{\etalchar{*}}21]{juhl2021implicit}
\textsc{Juhl K.~A., Morales X., de~Backer O., Camara O., Paulsen R.~R.}:
\newblock Implicit neural distance representation for unsupervised and
  supervised classification of complex anatomies.
\newblock In \emph{International Conference on Medical Image Computing and
  Computer-Assisted Intervention} (2021), Springer, pp.~405--415.

\bibitem[JSM{\etalchar{*}}20]{jiang2020local}
\textsc{Jiang C., Sud A., Makadia A., Huang J., Nie{\ss}ner M., Funkhouser T.,
  et~al.}:
\newblock Local implicit grid representations for 3d scenes.
\newblock In \emph{Proc. CVPR} (2020), pp.~6001--6010.

\bibitem[JTA21]{jain2021dietnerf}
\textsc{Jain A., Tancik M., Abbeel P.}:
\newblock Putting nerf on a diet: Semantically consistent few-shot view
  synthesis.
\newblock \emph{arXiv preprint arXiv:2104.00677} (2021).
\newblock URL: \url{http://arxiv.org/abs/2104.00677v1}.

\bibitem[JZS{\etalchar{*}}21]{jiang2021giga}
\textsc{Jiang Z., Zhu Y., Svetlik M., Fang K., Zhu Y.}:
\newblock Synergies between affordance and geometry: 6-dof grasp detection via
  implicit representations.
\newblock In \emph{Proceedings of Robotics: Science and Systems} (2021).
\newblock URL: \url{http://arxiv.org/abs/2104.01542v2}.

\bibitem[KA03]{kim2003approximation}
\textsc{Kim T., Adal{\i} T.}:
\newblock Approximation by fully complex multilayer perceptrons.
\newblock \emph{Neural computation 15}, 7 (2003), 1641--1666.

\bibitem[KA13]{karras2013fast}
\textsc{Karras T., Aila T.}:
\newblock Fast parallel construction of high-quality bounding volume
  hierarchies.
\newblock In \emph{Proceedings of the 5th High-Performance Graphics Conference}
  (2013), pp.~89--99.

\bibitem[Kaj86]{kajiya1986rendering}
\textsc{Kajiya J.~T.}:
\newblock The rendering equation.
\newblock In \emph{Proceedings of the 13th annual conference on Computer
  graphics and interactive techniques} (1986), pp.~143--150.

\bibitem[KAL{\etalchar{*}}21]{karras2021alias}
\textsc{Karras T., Aittala M., Laine S., H{\"a}rk{\"o}nen E., Hellsten J.,
  Lehtinen J., Aila T.}:
\newblock Alias-free generative adversarial networks.
\newblock \emph{Proc. NeurIPS} (2021).

\bibitem[KAT{\etalchar{*}}21]{khot2021surrf}
\textsc{Khot T., Agrawal S., Tulsiani S., Mertz C., Lucey S., Hebert M.}:
\newblock Learning unsupervised multi-view stereopsis via robust photometric
  consistency.
\newblock \emph{arXiv preprint arXiv:1905.02706} (2021).
\newblock URL: \url{http://arxiv.org/abs/1905.02706v2}.

\bibitem[KBH21]{knodt2021neuralraytracing}
\textsc{Knodt J., Baek S.-H., Heide F.}:
\newblock Neural ray-tracing: Learning surfaces and reflectance for relighting
  and view synthesis.
\newblock \emph{arXiv preprint arXiv:2104.13562} (2021).
\newblock URL: \url{http://arxiv.org/abs/2104.13562v1}.

\bibitem[KH84]{kajiya1984ray}
\textsc{Kajiya J.~T., Herzen B. P.~V.}:
\newblock Ray tracing volume densities.
\newblock \emph{ACM SIGGRAPH computer graphics 18}, 3 (1984), 165--174.

\bibitem[KIT{\etalchar{*}}21]{kondo2021vaxnerf}
\textsc{Kondo N., Ikeda Y., Tagliasacchi A., Matsuo Y., Ochiai Y., Gu S.~S.}:
\newblock Vaxnerf: Revisiting the classic for voxel-accelerated neural radiance
  field.
\newblock URL: \url{http://arxiv.org/abs/2111.13112v1}.

\bibitem[KJJ{\etalchar{*}}21]{kellnhofer2021nlr}
\textsc{Kellnhofer P., Jebe L., Jones A., Spicer R., Pulli K., Wetzstein G.}:
\newblock Neural lumigraph rendering.
\newblock In \emph{Proceedings of the IEEE/CVF Conference on Computer Vision
  and Pattern Recognition (CVPR)} (2021).
\newblock URL: \url{http://arxiv.org/abs/2103.11571v1}.

\bibitem[KKCF21]{kwon2021neuralhumanperformer}
\textsc{Kwon Y., Kim D., Ceylan D., Fuchs H.}:
\newblock Neural human performer: Learning generalizable radiance fields for
  human performance rendering.
\newblock In \emph{Advances in Neural Information Processing Systems (NeurIPS)}
  (2021), Curran Associates, Inc.
\newblock URL: \url{http://arxiv.org/abs/2109.07448v1}.

\bibitem[KKL{\etalchar{*}}21]{karniadakis2021physics}
\textsc{Karniadakis G.~E., Kevrekidis I.~G., Lu L., Perdikaris P., Wang S.,
  Yang L.}:
\newblock Physics-informed machine learning.
\newblock \emph{Nature Reviews Physics 3}, 6 (2021), 422--440.

\bibitem[KOWD21]{kasten2021layered}
\textsc{Kasten Y., Ofri D., Wang O., Dekel T.}:
\newblock Layered neural atlases for consistent video editing.
\newblock \emph{ACM Transactions on Graphics (TOG)} (2021).
\newblock URL: \url{http://arxiv.org/abs/2109.11418v1}.

\bibitem[KS01]{kak2001principles}
\textsc{Kak A.~C., Slaney M.}:
\newblock \emph{Front Matter}.
\newblock SIAM, 2001, pp.~i--xiv.
\newblock URL: \url{https://epubs.siam.org/doi/abs/10.1137/1.9780898719277.fm},
  \href {https://doi.org/10.1137/1.9780898719277.fm}
  {\path{doi:10.1137/1.9780898719277.fm}}.

\bibitem[KS15]{kim2015_affordancemap}
\textsc{Kim D.~I., Sukhatme G.~S.}:
\newblock Interactive affordance map building for a robotic task.
\newblock In \emph{2015 IEEE/RSJ International Conference on Intelligent Robots
  and Systems (IROS)} (2015), pp.~4581--4586.
\newblock \href {https://doi.org/10.1109/IROS.2015.7354029}
  {\path{doi:10.1109/IROS.2015.7354029}}.

\bibitem[KSF{\etalchar{*}}21]{karunratanakul2021halo}
\textsc{Karunratanakul K., Spurr A., Fan Z., Hilliges O., Tang S.}:
\newblock A skeleton-driven neural occupancy representation for articulated
  hands.
\newblock \emph{arXiv preprint arXiv:2109.11399} (2021).
\newblock URL: \url{http://arxiv.org/abs/2109.11399v1}.

\bibitem[KSW20]{kohli2020semantic}
\textsc{Kohli A., Sitzmann V., Wetzstein G.}:
\newblock Semantic implicit neural scene representations with semi-supervised
  training.
\newblock In \emph{International Conference on 3D Vision (3DV)} (2020), IEEE.
\newblock URL: \url{http://arxiv.org/abs/2003.12673v2}.

\bibitem[KSZ{\etalchar{*}}21]{kosiorek2021nerfvae}
\textsc{Kosiorek A.~R., Strathmann H., Zoran D., Moreno P., Schneider R., Mokra
  S., Rezende D.~J.}:
\newblock Nerf-vae: A geometry aware 3d scene generative model.
\newblock \emph{arXiv preprint arXiv:2104.00587} (2021).
\newblock URL: \url{http://arxiv.org/abs/2104.00587v1}.

\bibitem[KV21]{khademi2021view}
\textsc{Khademi W., Ventura J.}:
\newblock View synthesis in casually captured scenes using a cylindrical neural
  radiance field with exposure compensation.
\newblock \emph{ACM Transactions on Graphics (TOG)} (2021).
\newblock URL: \url{https://doi.org/10.1145/3450618.3469147}, \href
  {https://doi.org/10.1145/3450618.3469147}
  {\path{doi:10.1145/3450618.3469147}}.

\bibitem[KYZ{\etalchar{*}}20]{karunratanakul2020graspingfield}
\textsc{Karunratanakul K., Yang J., Zhang Y., Black M., Muandet K., Tang S.}:
\newblock Grasping field: Learning implicit representations for human grasps.
\newblock In \emph{International Conference on 3D Vision (3DV)} (2020), IEEE.
\newblock URL: \url{http://arxiv.org/abs/2008.04451v3}.

\bibitem[LAM06]{larsson2006dynamic}
\textsc{Larsson T., Akenine-M{\"o}ller T.}:
\newblock A dynamic bounding volume hierarchy for generalized collision
  detection.
\newblock \emph{Computers \& Graphics 30}, 3 (2006), 450--459.

\bibitem[LBB{\etalchar{*}}17]{li2017learning}
\textsc{Li T., Bolkart T., Black M.~J., Li H., Romero J.}:
\newblock Learning a model of facial shape and expression from 4d scans.
\newblock \emph{ACM Trans. Graph. 36}, 6 (2017), 194--1.

\bibitem[LBH15]{lecun2015deep}
\textsc{LeCun Y., Bengio Y., Hinton G.}:
\newblock Deep learning.
\newblock \emph{nature 521}, 7553 (2015), 436--444.

\bibitem[LCZ{\etalchar{*}}21]{luo2021convolutional}
\textsc{Luo H., Chen A., Zhang Q., Pang B., Wu M., Xu L., Yu J.}:
\newblock Convolutional neural opacity radiance fields.
\newblock URL: \url{http://arxiv.org/abs/2104.01772v1}.

\bibitem[LEC{\etalchar{*}}20]{levinson2020analysis}
\textsc{Levinson J., Esteves C., Chen K., Snavely N., Kanazawa A., Rostamizadeh
  A., Makadia A.}:
\newblock An analysis of svd for deep rotation estimation.
\newblock \emph{arXiv preprint arXiv:2006.14616} (2020).

\bibitem[Lem91]{lemmon1991bneuralfieldspathplanning}
\textsc{Lemmon M.}:
\newblock 2-degree-of-freedom robot path planning using cooperative neural
  fields.
\newblock \emph{Neural Computation 3}, 3 (1991), 350--362.

\bibitem[Lem92]{lemmon1991aneuralfieldspathplanning}
\textsc{Lemmon M.}:
\newblock Oscillatory neural fields for globally optimal path planning.
\newblock In \emph{Advances in Neural Information Processing Systems} (1992),
  Moody J., Hanson S., Lippmann R.~P., (Eds.), vol.~4, Morgan-Kaufmann.
\newblock URL:
  \url{https://proceedings.neurips.cc/paper/1991/file/c9892a989183de32e976c6f04e700201-Paper.pdf}.

\bibitem[LFS{\etalchar{*}}21]{li2021mine}
\textsc{Li J., Feng Z., She Q., Ding H., Wang C., Lee G.~H.}:
\newblock Mine: Towards continuous depth mpi with nerf for novel view
  synthesis.
\newblock In \emph{Proceedings of the IEEE International Conference on Computer
  Vision (ICCV)} (2021).
\newblock URL: \url{http://arxiv.org/abs/2103.14910v3}.

\bibitem[LGL{\etalchar{*}}20]{liu2020nsvf}
\textsc{Liu L., Gu J., Lin K.~Z., Chua T.-S., Theobalt C.}:
\newblock Neural sparse voxel fields.
\newblock In \emph{Proceedings of the European Conference on Computer Vision
  (ECCV)} (2020).
\newblock URL: \url{http://arxiv.org/abs/2007.11571v2}.

\bibitem[LHR{\etalchar{*}}21]{liu2021na}
\textsc{Liu L., Habermann M., Rudnev V., Sarkar K., Gu J., Theobalt C.}:
\newblock Neural actor: Neural free-view synthesis of human actors with pose
  control.
\newblock \emph{ACM Transactions on Graphics (TOG)} (2021).
\newblock URL: \url{http://arxiv.org/abs/2106.02019v1}.

\bibitem[LJLB21]{lu2021compressive}
\textsc{Lu Y., Jiang K., Levine J.~A., Berger M.}:
\newblock Compressive neural representations of volumetric scalar fields.
\newblock URL: \url{http://arxiv.org/abs/2104.04523v1}.

\bibitem[LLS{\etalchar{*}}21]{li20213d}
\textsc{Li Y., Li S., Sitzmann V., Agrawal P., Torralba A.}:
\newblock 3d neural scene representations for visuomotor control.
\newblock In \emph{Proceedings of Robotics: Science and Systems} (2021).
\newblock URL: \url{http://arxiv.org/abs/2107.04004v1}.

\bibitem[LLYX21]{liu2021neulf}
\textsc{Liu C., Li Z., Yuan J., Xu Y.}:
\newblock Neulf: Efficient novel view synthesis with neural 4d light field.
\newblock \emph{arXiv preprint arXiv:2105.07112} (2021).
\newblock URL: \url{http://arxiv.org/abs/2105.07112v4}.

\bibitem[LMR{\etalchar{*}}15]{loper2015smpl}
\textsc{Loper M., Mahmood N., Romero J., Pons-Moll G., Black M.~J.}:
\newblock Smpl: A skinned multi-person linear model.
\newblock \emph{ACM transactions on graphics (TOG) 34}, 6 (2015), 1--16.

\bibitem[LMTL21]{lin2021barf}
\textsc{Lin C.-H., Ma W.-C., Torralba A., Lucey S.}:
\newblock Barf: Bundle-adjusting neural radiance fields.
\newblock In \emph{Proceedings of the IEEE International Conference on Computer
  Vision (ICCV)} (2021).
\newblock URL: \url{http://arxiv.org/abs/2104.06405v2}.

\bibitem[LMW{\etalchar{*}}21a]{li2021learning}
\textsc{Li X., Mello S.~D., Wang X., Yang M.-H., Kautz J., Liu S.}:
\newblock Learning continuous environment fields via implicit functions.
\newblock \emph{arXiv preprint arXiv:2111.13997} (2021).

\bibitem[LMW21b]{lindell2021autoint}
\textsc{Lindell D.~B., Martel J. N.~P., Wetzstein G.}:
\newblock Autoint: Automatic integration for fast neural volume rendering.
\newblock In \emph{Proceedings of the IEEE/CVF Conference on Computer Vision
  and Pattern Recognition (CVPR)} (2021).
\newblock URL: \url{http://arxiv.org/abs/2012.01714v2}.

\bibitem[LNSW21]{li2021nsff}
\textsc{Li Z., Niklaus S., Snavely N., Wang O.}:
\newblock Neural scene flow fields for space-time view synthesis of dynamic
  scenes.
\newblock In \emph{Proceedings of the IEEE/CVF Conference on Computer Vision
  and Pattern Recognition (CVPR)} (2021).
\newblock URL: \url{http://arxiv.org/abs/2011.13084v3}.

\bibitem[LQCA20]{long2021learning}
\textsc{Long K., Qian C., Cortes J., Atanasov N.}:
\newblock Learning barrier functions with memory for robust safe navigation.
\newblock URL: \url{http://arxiv.org/abs/2011.01899v2}.

\bibitem[LS04]{lim20043d}
\textsc{Lim W.~P., Shamsuddin S.~M.}:
\newblock \emph{3D Object Reconstruction and Representation Using Neural
  Networks}.
\newblock Universiti Teknologi Malaysia, 2004.

\bibitem[LSCL19]{liu2019learning}
\textsc{Liu S., Saito S., Chen W., Li H.}:
\newblock Learning to infer implicit surfaces without 3d supervision.
\newblock In \emph{Advances in Neural Information Processing Systems (NeurIPS)}
  (2019), Curran Associates, Inc.
\newblock URL: \url{http://arxiv.org/abs/1911.00767v1}.

\bibitem[LSG{\etalchar{*}}20]{lei2020pix2surf}
\textsc{Lei J., Sridhar S., Guerrero P., Sung M., Mitra N., Guibas L.~J.}:
\newblock Pix2surf: Learning parametric 3d surface models of objects from
  images.
\newblock In \emph{Proceedings of the European Conference on Computer Vision
  (ECCV)} (2020).
\newblock URL: \url{http://arxiv.org/abs/2008.07760v1}.

\bibitem[LSS{\etalchar{*}}19]{lombardi2019nv}
\textsc{Lombardi S., Simon T., Saragih J., Schwartz G., Lehrmann A., Sheikh
  Y.}:
\newblock Neural volumes: Learning dynamic renderable volumes from images.
\newblock \emph{ACM Transactions on Graphics (TOG)} (2019).
\newblock URL: \url{http://arxiv.org/abs/1906.07751v1}, \href
  {https://doi.org/10.1145/3306346.3323020}
  {\path{doi:10.1145/3306346.3323020}}.

\bibitem[LSZ{\etalchar{*}}21]{li2021dynerf}
\textsc{Li T., Slavcheva M., Zollhoefer M., Green S., Lassner C., Kim C.,
  Schmidt T., Lovegrove S., Goesele M., Lv Z.}:
\newblock Neural 3d video synthesis.
\newblock \emph{arXiv preprint arXiv:2103.02597} (2021).
\newblock URL: \url{http://arxiv.org/abs/2103.02597v1}.

\bibitem[LTLS21]{lee2021metasparseinr}
\textsc{Lee J., Tack J., Lee N., Shin J.}:
\newblock Meta-learning sparse implicit neural representations.
\newblock URL: \url{http://arxiv.org/abs/2110.14678v2}.

\bibitem[LTW{\etalchar{*}}21]{li2021unsupervised}
\textsc{Li N., Thapa S., Whyte C., Reed A., Jayasuriya S., Ye J.}:
\newblock Unsupervised non-rigid image distortion removal via grid deformation.
\newblock In \emph{Proceedings of the IEEE International Conference on Computer
  Vision (ICCV)} (2021).

\bibitem[LW19]{littwin2019deep}
\textsc{Littwin G., Wolf L.}:
\newblock Deep meta functionals for shape representation.
\newblock In \emph{Proceedings of the IEEE International Conference on Computer
  Vision (ICCV)} (2019).
\newblock URL: \url{http://arxiv.org/abs/1908.06277v1}.

\bibitem[LWL20]{lin2020sdfsrn}
\textsc{Lin C.-H., Wang C., Lucey S.}:
\newblock Sdf-srn: Learning signed distance 3d object reconstruction from
  static images.
\newblock In \emph{Advances in Neural Information Processing Systems (NeurIPS)}
  (2020), Curran Associates, Inc.
\newblock URL: \url{http://arxiv.org/abs/2010.10505v1}.

\bibitem[LXS{\etalchar{*}}20]{li2020monoport}
\textsc{Li R., Xiu Y., Saito S., Huang Z., Olszewski K., Li H.}:
\newblock Monocular real-time volumetric performance capture.
\newblock In \emph{Proceedings of the European Conference on Computer Vision
  (ECCV)} (2020).
\newblock URL: \url{http://arxiv.org/abs/2007.13988v1}.

\bibitem[LYP{\etalchar{*}}20]{li2020pifusion}
\textsc{Li Z., Yu T., Pan C., Zheng Z., Liu Y.}:
\newblock Robust 3d self-portraits in seconds.
\newblock In \emph{Proceedings of the IEEE/CVF Conference on Computer Vision
  and Pattern Recognition (CVPR)} (2020).
\newblock URL: \url{http://arxiv.org/abs/2004.02460v1}.

\bibitem[LZP{\etalchar{*}}20]{liu2020dist}
\textsc{Liu S., Zhang Y., Peng S., Shi B., Pollefeys M., Cui Z.}:
\newblock Dist: Rendering deep implicit signed distance function with
  differentiable sphere tracing.
\newblock In \emph{Proceedings of the IEEE/CVF Conference on Computer Vision
  and Pattern Recognition (CVPR)} (2020).
\newblock URL: \url{http://arxiv.org/abs/1911.13225v2}.

\bibitem[LZZ{\etalchar{*}}21]{liu2021editing}
\textsc{Liu S., Zhang X., Zhang Z., Zhang R., Zhu J.-Y., Russell B.}:
\newblock Editing conditional radiance fields.
\newblock \emph{arXiv preprint arXiv:2105.06466} (2021).
\newblock URL: \url{http://arxiv.org/abs/2105.06466v2}.

\bibitem[MAKM21]{morreale2021neural}
\textsc{Morreale L., Aigerman N., Kim V., Mitra N.~J.}:
\newblock Neural surface maps.
\newblock In \emph{Proceedings of the IEEE/CVF Conference on Computer Vision
  and Pattern Recognition (CVPR)} (2021).
\newblock URL: \url{http://arxiv.org/abs/2103.16942v1}.

\bibitem[MBOL{\etalchar{*}}21]{michel2022text2mesh}
\textsc{Michel O., Bar-On R., Liu R., Benaim S., Hanocka R.}:
\newblock Text2mesh: Text-driven neural stylization for meshes.
\newblock URL: \url{http://arxiv.org/abs/2112.03221v1}.

\bibitem[MBRS{\etalchar{*}}21]{martin-brualla2021nerfw}
\textsc{Martin-Brualla R., Radwan N., Sajjadi M. S.~M., Barron J.~T.,
  Dosovitskiy A., Duckworth D.}:
\newblock Nerf in the wild: Neural radiance fields for unconstrained photo
  collections.
\newblock In \emph{Proceedings of the IEEE/CVF Conference on Computer Vision
  and Pattern Recognition (CVPR)} (2021).
\newblock URL: \url{http://arxiv.org/abs/2008.02268v3}.

\bibitem[MCL{\etalchar{*}}21]{meng2021gnerf}
\textsc{Meng Q., Chen A., Luo H., Wu M., Su H., Xu L., He X., Yu J.}:
\newblock Gnerf: Gan-based neural radiance field without posed camera.
\newblock URL: \url{http://arxiv.org/abs/2103.15606v3}.

\bibitem[McM02]{mcmullin2002origins}
\textsc{McMullin E.}:
\newblock The origins of the field concept in physics.
\newblock \emph{Physics in Perspective 4}, 1 (2002), 13--39.

\bibitem[MCN20]{mehta2020extreme}
\textsc{Mehta H., Cutkosky A., Neyshabur B.}:
\newblock Extreme memorization via scale of initialization.
\newblock \emph{arXiv preprint arXiv:2008.13363} (2020).

\bibitem[MDH{\etalchar{*}}20]{mcguire2020taxonomy}
\textsc{McGuire M., Dorsey J., Haines E., Hughes J.~F., Marschner S., Pharr M.,
  Shirley P.}:
\newblock A taxonomy of bidirectional scattering distribution function lobes
  for rendering engineers.
\newblock In \emph{MAM@ EGSR} (2020), pp.~25--28.

\bibitem[MEIL19]{mitchell2022hof}
\textsc{Mitchell E., Engin S., Isler V., Lee D.~D.}:
\newblock Higher-order function networks for learning composable 3d object
  representations.
\newblock URL: \url{http://arxiv.org/abs/1907.10388v2}.

\bibitem[MEJ{\etalchar{*}}21]{murphy2021implicitpdf}
\textsc{Murphy K., Esteves C., Jampani V., Ramalingam S., Makadia A.}:
\newblock Implicit-pdf: Non-parametric representation of probability
  distributions on the rotation manifold.
\newblock In \emph{International Conference on Machine Learning (ICML)} (2021),
  PMLR.
\newblock URL: \url{http://arxiv.org/abs/2106.05965v1}.

\bibitem[MGB{\etalchar{*}}21]{mehta2021modulated}
\textsc{Mehta I., Gharbi M., Barnes C., Shechtman E., Ramamoorthi R.,
  Chandraker M.}:
\newblock Modulated periodic activations for generalizable local functional
  representations.
\newblock In \emph{Proceedings of the IEEE International Conference on Computer
  Vision (ICCV)} (2021).
\newblock URL: \url{http://arxiv.org/abs/2104.03960v1}.

\bibitem[MLDI21]{mergy2021visionbased}
\textsc{Mergy A., Lecuyer G., Derksen D., Izzo D.}:
\newblock Vision-based neural scene representations for spacecraft.
\newblock In \emph{Proceedings of the IEEE/CVF Conference on Computer Vision
  and Pattern Recognition (CVPR)} (2021).
\newblock URL: \url{http://arxiv.org/abs/2105.06405v1}.

\bibitem[MLL{\etalchar{*}}21]{martel2021acorn}
\textsc{Martel J. N.~P., Lindell D.~B., Lin C.~Z., Chan E.~R., Monteiro M.,
  Wetzstein G.}:
\newblock Acorn: Adaptive coordinate networks for neural scene representation.
\newblock \emph{ACM Transactions on Graphics (TOG)} (2021).
\newblock URL: \url{http://arxiv.org/abs/2105.02788v1}.

\bibitem[MML{\etalchar{*}}21]{mandl2021neuralcameras}
\textsc{Mandl D., Mohr P., Langlotz T., Ebner C., Mori S., Zollmann S., Roth
  P.~M., Kalkofen D.}:
\newblock Neural cameras: Learning camera characteristics for coherent mixed
  reality rendering.

\bibitem[MMNM21a]{moseley2021fbpinns}
\textsc{Moseley B., Markham A., Nissen-Meyer T.}:
\newblock Finite basis physics-informed neural networks (fbpinns): a scalable
  domain decomposition approach for solving differential equations.
\newblock \emph{arXiv preprint arXiv:2107.07871} (2021).
\newblock URL: \url{http://arxiv.org/abs/2107.07871v1}.

\bibitem[MMNM21b]{moseley2021finite}
\textsc{Moseley B., Markham A., Nissen-Meyer T.}:
\newblock Finite basis physics-informed neural networks (fbpinns): a scalable
  domain decomposition approach for solving differential equations.
\newblock \emph{arXiv preprint arXiv:2107.07871} (2021).

\bibitem[MOB{\etalchar{*}}21]{meister2021survey}
\textsc{Meister D., Ogaki S., Benthin C., Doyle M.~J., Guthe M., Bittner J.}:
\newblock A survey on bounding volume hierarchies for ray tracing.
\newblock In \emph{Computer Graphics Forum} (2021), vol.~40, Wiley Online
  Library, pp.~683--712.

\bibitem[MON{\etalchar{*}}19]{mescheder2019occupancynetworks}
\textsc{Mescheder L., Oechsle M., Niemeyer M., Nowozin S., Geiger A.}:
\newblock Occupancy networks: Learning 3d reconstruction in function space.
\newblock In \emph{Proceedings of the IEEE/CVF Conference on Computer Vision
  and Pattern Recognition (CVPR)} (2019).
\newblock URL: \url{http://arxiv.org/abs/1812.03828v2}.

\bibitem[MQK{\etalchar{*}}21]{mu2021asdf}
\textsc{Mu J., Qiu W., Kortylewski A., Yuille A., Vasconcelos N., Wang X.}:
\newblock A-sdf: Learning disentangled signed distance functions for
  articulated shape representation.
\newblock \emph{arXiv preprint arXiv:2104.07645} (2021).
\newblock URL: \url{http://arxiv.org/abs/2104.07645v1}.

\bibitem[MRNK21]{muller2021realtime}
\textsc{Muller T., Rousselle F., Novak J., Keller A.}:
\newblock Real-time neural radiance caching for path tracing.
\newblock \emph{ACM Transactions on Graphics (TOG)} (2021).
\newblock URL: \url{http://arxiv.org/abs/2106.12372v2}, \href
  {https://doi.org/10.1145/3450626.3459812}
  {\path{doi:10.1145/3450626.3459812}}.

\bibitem[MS15]{maturana2015voxnet}
\textsc{Maturana D., Scherer S.}:
\newblock Voxnet: A 3d convolutional neural network for real-time object
  recognition.
\newblock In \emph{2015 IEEE/RSJ International Conference on Intelligent Robots
  and Systems (IROS)} (2015), pp.~922--928.
\newblock \href {https://doi.org/10.1109/IROS.2015.7353481}
  {\path{doi:10.1109/IROS.2015.7353481}}.

\bibitem[MSOC{\etalchar{*}}19]{mildenhall2019llff}
\textsc{Mildenhall B., Srinivasan P.~P., Ortiz-Cayon R., Kalantari N.~K.,
  Ramamoorthi R., Ng R., Kar A.}:
\newblock Local light field fusion: Practical view synthesis with prescriptive
  sampling guidelines.
\newblock \emph{ACM Transactions on Graphics (TOG)} (2019).

\bibitem[MSS{\etalchar{*}}21]{ma2021pica}
\textsc{Ma S., Simon T., Saragih J., Wang D., Li Y., Torre F. D.~L., Sheikh
  Y.}:
\newblock Pixel codec avatars.
\newblock In \emph{Proceedings of the IEEE/CVF Conference on Computer Vision
  and Pattern Recognition (CVPR)} (2021).
\newblock URL: \url{http://arxiv.org/abs/2104.04638v1}.

\bibitem[MST{\etalchar{*}}20]{mildenhall2020nerf}
\textsc{Mildenhall B., Srinivasan P.~P., Tancik M., Barron J.~T., Ramamoorthi
  R., Ng R.}:
\newblock Nerf: Representing scenes as neural radiance fields for view
  synthesis.
\newblock In \emph{Proceedings of the European Conference on Computer Vision
  (ECCV)} (2020).
\newblock URL: \url{http://arxiv.org/abs/2003.08934v2}.

\bibitem[MZBT21]{mihajlovic2021leap}
\textsc{Mihajlovic M., Zhang Y., Black M.~J., Tang S.}:
\newblock Leap: Learning articulated occupancy of people.
\newblock In \emph{Proceedings of the IEEE/CVF Conference on Computer Vision
  and Pattern Recognition (CVPR)} (2021).
\newblock URL: \url{http://arxiv.org/abs/2104.06849v1}.

\bibitem[NBB21]{nam2021neural}
\textsc{Nam S., Brubaker M.~A., Brown M.~S.}:
\newblock Neural image representations for multi-image fusion and layer
  separation.
\newblock \emph{arXiv preprint arXiv:2108.01199} (2021).
\newblock URL: \url{http://arxiv.org/abs/2108.01199v2}.

\bibitem[NeRa]{NeRFatIC64:online}
Nerf at iccv 2021 - frank dellaert.
\newblock \url{https://dellaert.github.io/NeRF21/}.

\bibitem[NeRb]{NeRFExpl90:online}
Nerf explosion 2020 - frank dellaert.
\newblock \url{https://dellaert.github.io/NeRF/}.

\bibitem[NeRc]{NeRFNeur57:online}
Nerf: Neural radiance fields - ai research graph - crossminds.
\newblock
  \url{https://crossminds.ai/graphlist/nerf-neural-radiance-fields-ai-research-graph-60708936c8663c4cfa875fc2/}.

\bibitem[NG21a]{niemeyer2021campari}
\textsc{Niemeyer M., Geiger A.}:
\newblock Campari: Camera-aware decomposed generative neural radiance fields.
\newblock \emph{arXiv preprint arXiv:2103.17269} (2021).
\newblock URL: \url{http://arxiv.org/abs/2103.17269v1}.

\bibitem[NG21b]{niemeyer2021giraffe}
\textsc{Niemeyer M., Geiger A.}:
\newblock Giraffe: Representing scenes as compositional generative neural
  feature fields.
\newblock In \emph{Proceedings of the IEEE/CVF Conference on Computer Vision
  and Pattern Recognition (CVPR)} (2021).
\newblock URL: \url{http://arxiv.org/abs/2011.12100v2}.

\bibitem[NGHJ18]{novak2018monte}
\textsc{Nov{\'a}k J., Georgiev I., Hanika J., Jarosz W.}:
\newblock Monte carlo methods for volumetric light transport simulation.
\newblock In \emph{Computer Graphics Forum} (2018), vol.~37, Wiley Online
  Library, pp.~551--576.

\bibitem[NIH{\etalchar{*}}11]{newcombe2011kinectfusion}
\textsc{Newcombe R.~A., Izadi S., Hilliges O., Molyneaux D., Kim D., Davison
  A.~J., Kohi P., Shotton J., Hodges S., Fitzgibbon A.}:
\newblock Kinectfusion: Real-time dense surface mapping and tracking.
\newblock In \emph{2011 10th IEEE international symposium on mixed and
  augmented reality} (2011), IEEE, pp.~127--136.

\bibitem[NMOG19]{niemeyer2019occupancyflow}
\textsc{Niemeyer M., Mescheder L., Oechsle M., Geiger A.}:
\newblock Occupancy flow: 4d reconstruction by learning particle dynamics.
\newblock In \emph{Proceedings of the IEEE International Conference on Computer
  Vision (ICCV)} (2019), pp.~5379--5389.

\bibitem[NMOG20]{niemeyer2020dvr}
\textsc{Niemeyer M., Mescheder L., Oechsle M., Geiger A.}:
\newblock Differentiable volumetric rendering: Learning implicit 3d
  representations without 3d supervision.
\newblock In \emph{Proceedings of the IEEE/CVF Conference on Computer Vision
  and Pattern Recognition (CVPR)} (2020).
\newblock URL: \url{http://arxiv.org/abs/1912.07372v2}.

\bibitem[NNB04]{nister2004visual}
\textsc{Nist{\'e}r D., Naroditsky O., Bergen J.}:
\newblock Visual odometry.
\newblock In \emph{Proceedings of the 2004 IEEE Computer Society Conference on
  Computer Vision and Pattern Recognition, 2004. CVPR 2004.} (2004), vol.~1,
  Ieee, pp.~I--I.

\bibitem[NPLT{\etalchar{*}}19]{nguyen2019hologan}
\textsc{Nguyen-Phuoc T., Li C., Theis L., Richardt C., Yang Y.-L.}:
\newblock Hologan: Unsupervised learning of 3d representations from natural
  images.
\newblock In \emph{Proc. ICCV} (2019), pp.~7588--7597.

\bibitem[NSLH21]{noguchi2021narf}
\textsc{Noguchi A., Sun X., Lin S., Harada T.}:
\newblock Neural articulated radiance field.
\newblock In \emph{Proceedings of the IEEE International Conference on Computer
  Vision (ICCV)} (2021).
\newblock URL: \url{http://arxiv.org/abs/2104.03110v2}.

\bibitem[NSP{\etalchar{*}}21]{neff2021donerf}
\textsc{Neff T., Stadlbauer P., Parger M., Kurz A., Mueller J.~H., Chaitanya C.
  R.~A., Kaplanyan A., Steinberger M.}:
\newblock Donerf: Towards real-time rendering of compact neural radiance fields
  using depth oracle networks.
\newblock \emph{Computer Graphics Forum} (2021).
\newblock URL: \url{http://arxiv.org/abs/2103.03231v4}, \href
  {https://doi.org/10.1111/cgf.14340} {\path{doi:10.1111/cgf.14340}}.

\bibitem[NWH21]{nirkin2021hyperseg}
\textsc{Nirkin Y., Wolf L., Hassner T.}:
\newblock {HyperSeg}: Patch-wise hypernetwork for real-time semantic
  segmentation.
\newblock In \emph{IEEE/CVF Conference on Computer Vision and Pattern
  Recognition (CVPR)} (June 2021).

\bibitem[OELS{\etalchar{*}}21]{or-el2021stylesdf}
\textsc{Or-El R., Luo X., Shan M., Shechtman E., Park J.~J.,
  Kemelmacher-Shlizerman I.}:
\newblock Stylesdf: High-resolution 3d-consistent image and geometry
  generation.
\newblock URL: \url{http://arxiv.org/abs/2112.11427v1}.

\bibitem[OF03]{osher2003signed}
\textsc{Osher S., Fedkiw R.}:
\newblock Signed distance functions.
\newblock In \emph{Level set methods and dynamic implicit surfaces}. Springer,
  2003, pp.~17--22.

\bibitem[OMN{\etalchar{*}}19]{oechsle2019texturefields}
\textsc{Oechsle M., Mescheder L., Niemeyer M., Strauss T., Geiger A.}:
\newblock Texture fields: Learning texture representations in function space.
\newblock In \emph{Proceedings of the IEEE International Conference on Computer
  Vision (ICCV)} (2019).
\newblock URL: \url{http://arxiv.org/abs/1905.07259v1}.

\bibitem[OMT{\etalchar{*}}21]{ost2021neural}
\textsc{Ost J., Mannan F., Thuerey N., Knodt J., Heide F.}:
\newblock Neural scene graphs for dynamic scenes.
\newblock In \emph{Proceedings of the IEEE/CVF Conference on Computer Vision
  and Pattern Recognition (CVPR)} (2021).
\newblock URL: \url{http://arxiv.org/abs/2011.10379v3}.

\bibitem[OPG21]{oechsle2021unisurf}
\textsc{Oechsle M., Peng S., Geiger A.}:
\newblock Unisurf: Unifying neural implicit surfaces and radiance fields for
  multi-view reconstruction.
\newblock In \emph{Proceedings of the IEEE International Conference on Computer
  Vision (ICCV)} (2021).
\newblock URL: \url{http://arxiv.org/abs/2104.10078v1}.

\bibitem[PBD{\etalchar{*}}10]{parker2010optix}
\textsc{Parker S.~G., Bigler J., Dietrich A., Friedrich H., Hoberock J., Luebke
  D., McAllister D., McGuire M., Morley K., Robison A., et~al.}:
\newblock Optix: a general purpose ray tracing engine.
\newblock \emph{Acm transactions on graphics (tog) 29}, 4 (2010), 1--13.

\bibitem[PBT{\etalchar{*}}21]{palafox2021npms}
\textsc{Palafox P., Bozic A., Thies J., Niessner M., Dai A.}:
\newblock Npms: Neural parametric models for 3d deformable shapes.
\newblock In \emph{Proceedings of the IEEE International Conference on Computer
  Vision (ICCV)} (2021).
\newblock URL: \url{http://arxiv.org/abs/2104.00702v2}.

\bibitem[PCPMMN21]{pumarola2021dnerf}
\textsc{Pumarola A., Corona E., Pons-Moll G., Moreno-Noguer F.}:
\newblock D-nerf: Neural radiance fields for dynamic scenes.
\newblock In \emph{Proceedings of the IEEE/CVF Conference on Computer Vision
  and Pattern Recognition (CVPR)} (2021).
\newblock URL: \url{http://arxiv.org/abs/2011.13961v1}.

\bibitem[PDW{\etalchar{*}}21]{peng2021animatable}
\textsc{Peng S., Dong J., Wang Q., Zhang S., Shuai Q., Bao H., Zhou X.}:
\newblock Animatable neural radiance fields for human body modeling.
\newblock In \emph{Proceedings of the IEEE International Conference on Computer
  Vision (ICCV)} (2021).
\newblock URL: \url{http://arxiv.org/abs/2105.02872v1}.

\bibitem[PFAK20]{poursaeed2020hybridnet}
\textsc{Poursaeed O., Fisher M., Aigerman N., Kim V.~G.}:
\newblock Coupling explicit and implicit surface representations for generative
  3d modeling.
\newblock In \emph{Proceedings of the European Conference on Computer Vision
  (ECCV)} (2020).
\newblock URL: \url{http://arxiv.org/abs/2007.10294v2}.

\bibitem[PFS{\etalchar{*}}19]{park2019deepsdf}
\textsc{Park J.~J., Florence P., Straub J., Newcombe R., Lovegrove S.}:
\newblock Deepsdf: Learning continuous signed distance functions for shape
  representation.
\newblock In \emph{Proceedings of the IEEE/CVF Conference on Computer Vision
  and Pattern Recognition (CVPR)} (2019).
\newblock URL: \url{http://arxiv.org/abs/1901.05103v1}.

\bibitem[PH89]{perlin1989hypertexture}
\textsc{Perlin K., Hoffert E.~M.}:
\newblock Hypertexture.
\newblock In \emph{Proceedings of the 16th annual conference on Computer
  graphics and interactive techniques} (1989), pp.~253--262.

\bibitem[PHP20]{pfrommer2020contactnets}
\textsc{Pfrommer S., Halm M., Posa M.}:
\newblock Contactnets: Learning discontinuous contact dynamics with smooth,
  implicit representations.
\newblock In \emph{Proceedings of the Conference on Robot Learning (CoRL)}
  (2020).
\newblock URL: \url{http://arxiv.org/abs/2009.11193v2}.

\bibitem[PJH16]{pharr2016physically}
\textsc{Pharr M., Jakob W., Humphreys G.}:
\newblock \emph{Physically based rendering: From theory to implementation}.
\newblock Morgan Kaufmann, 2016.

\bibitem[PJL{\etalchar{*}}21]{peng2021sap}
\textsc{Peng S., Jiang C.~M., Liao Y., Niemeyer M., Pollefeys M., Geiger A.}:
\newblock Shape as points: A differentiable poisson solver.
\newblock URL: \url{http://arxiv.org/abs/2106.03452v2}.

\bibitem[PNM{\etalchar{*}}20]{peng2020convolutional}
\textsc{Peng S., Niemeyer M., Mescheder L., Pollefeys M., Geiger A.}:
\newblock Convolutional occupancy networks.
\newblock In \emph{Proceedings of the European Conference on Computer Vision
  (ECCV)} (2020).
\newblock URL: \url{http://arxiv.org/abs/2003.04618v2}.

\bibitem[PSB{\etalchar{*}}21]{park2021nerfies}
\textsc{Park K., Sinha U., Barron J.~T., Bouaziz S., Goldman D.~B., Seitz
  S.~M., Martin-Brualla R.}:
\newblock Nerfies: Deformable neural radiance fields.
\newblock In \emph{Proceedings of the IEEE International Conference on Computer
  Vision (ICCV)} (2021).
\newblock URL: \url{http://arxiv.org/abs/2011.12948v5}.

\bibitem[PSH{\etalchar{*}}21]{park2021hypernerf}
\textsc{Park K., Sinha U., Hedman P., Barron J.~T., Bouaziz S., Goldman D.~B.,
  Martin-Brualla R., Seitz S.~M.}:
\newblock Hypernerf: A higher-dimensional representation for topologically
  varying neural radiance fields.
\newblock \emph{arXiv preprint arXiv:2106.13228} (2021).
\newblock URL: \url{http://arxiv.org/abs/2106.13228v2}.

\bibitem[PXL{\etalchar{*}}21]{pan2021shadegan}
\textsc{Pan X., Xu X., Loy C.~C., Theobalt C., Dai B.}:
\newblock A shading-guided generative implicit model for shape-accurate
  3d-aware image synthesis.
\newblock In \emph{Advances in Neural Information Processing Systems (NeurIPS)}
  (2021).

\bibitem[PXZ{\etalchar{*}}21]{pan2021adaptive}
\textsc{Pan H., Xiao D., Zhang F., Li X., Xu M.}:
\newblock Adaptive weight matrix and phantom intensity learning for computed
  tomography of chemiluminescence.
\newblock \emph{Opt. Express 29}, 15 (Jul 2021), 23682--23700.
\newblock URL:
  \url{http://www.opticsexpress.org/abstract.cfm?URI=oe-29-15-23682}, \href
  {https://doi.org/10.1364/OE.427459} {\path{doi:10.1364/OE.427459}}.

\bibitem[PZX{\etalchar{*}}21]{peng2021neuralbody}
\textsc{Peng S., Zhang Y., Xu Y., Wang Q., Shuai Q., Bao H., Zhou X.}:
\newblock Neural body: Implicit neural representations with structured latent
  codes for novel view synthesis of dynamic humans.
\newblock In \emph{Proceedings of the IEEE/CVF Conference on Computer Vision
  and Pattern Recognition (CVPR)} (2021).
\newblock URL: \url{http://arxiv.org/abs/2012.15838v2}.

\bibitem[QMGR20]{qi2020learning}
\textsc{Qi W., Mullapudi R.~T., Gupta S., Ramanan D.}:
\newblock Learning to move with affordance maps.
\newblock \emph{arXiv preprint arXiv:2001.02364} (2020).

\bibitem[QSMG17]{qi2017pointnet}
\textsc{Qi C.~R., Su H., Mo K., Guibas L.~J.}:
\newblock Pointnet: Deep learning on point sets for 3d classification and
  segmentation.
\newblock In \emph{Proceedings of the IEEE Conference on Computer Vision and
  Pattern Recognition (CVPR)} (2017), pp.~652--660.

\bibitem[RBA{\etalchar{*}}19]{rahaman2019spectral}
\textsc{Rahaman N., Baratin A., Arpit D., Draxler F., Lin M., Hamprecht F.,
  Bengio Y., Courville A.}:
\newblock On the spectral bias of neural networks.
\newblock In \emph{Proceedings of the 36th International Conference on Machine
  Learning} (09--15 Jun 2019), Chaudhuri K., Salakhutdinov R., (Eds.), vol.~97
  of \emph{Proceedings of Machine Learning Research}, PMLR, pp.~5301--5310.
\newblock URL: \url{https://proceedings.mlr.press/v97/rahaman19a.html}.

\bibitem[RBBJ]{reed2021implicit}
\textsc{Reed A., Blanford T., Brown D.~C., Jayasuriya S.}:
\newblock Implicit neural representations for deconvolving sas images.

\bibitem[RBZ{\etalchar{*}}20]{rempe2020caspr}
\textsc{Rempe D., Birdal T., Zhao Y., Gojcic Z., Sridhar S., Guibas L.~J.}:
\newblock Caspr: Learning canonical spatiotemporal point cloud representations.
\newblock \emph{arXiv preprint arXiv:2008.02792} (2020).

\bibitem[RJY{\etalchar{*}}21]{rebain2021derf}
\textsc{Rebain D., Jiang W., Yazdani S., Li K., Yi K.~M., Tagliasacchi A.}:
\newblock Derf: Decomposed radiance fields.
\newblock In \emph{Proceedings of the IEEE/CVF Conference on Computer Vision
  and Pattern Recognition (CVPR)} (2021).
\newblock URL: \url{http://arxiv.org/abs/2011.12490v1}.

\bibitem[RKA{\etalchar{*}}21]{reed2021inr}
\textsc{Reed A.~W., Kim H., Anirudh R., Mohan K.~A., Champley K., Kang J.,
  Jayasuriya S.}:
\newblock Dynamic ct reconstruction from limited views with implicit neural
  representations and parametric motion fields.
\newblock \emph{arXiv preprint arXiv:2104.11745} (2021).
\newblock URL: \url{http://arxiv.org/abs/2104.11745v1}.

\bibitem[RL21]{ramasinghe2021learning}
\textsc{Ramasinghe S., Lucey S.}:
\newblock Learning positional embeddings for coordinate-mlps.
\newblock URL: \url{http://arxiv.org/abs/2112.11577v1}.

\bibitem[RLS{\etalchar{*}}21]{rebain2021dmf}
\textsc{Rebain D., Li K., Sitzmann V., Yazdani S., Yi K.~M., Tagliasacchi A.}:
\newblock Deep medial fields.
\newblock \emph{arXiv preprint arXiv:2106.03804} (2021).
\newblock URL: \url{http://arxiv.org/abs/2106.03804v1}.

\bibitem[RMBF21]{rematas2021sharf}
\textsc{Rematas K., Martin-Brualla R., Ferrari V.}:
\newblock Sharf: Shape-conditioned radiance fields from a single view.
\newblock In \emph{International Conference on Machine Learning (ICML)} (2021),
  PMLR.
\newblock URL: \url{http://arxiv.org/abs/2102.08860v2}.

\bibitem[RPK19a]{raissi2019physics}
\textsc{Raissi M., Perdikaris P., Karniadakis G.~E.}:
\newblock Physics-informed neural networks: A deep learning framework for
  solving forward and inverse problems involving nonlinear partial differential
  equations.
\newblock \emph{Journal of Computational Physics 378} (2019), 686--707.

\bibitem[RPK19b]{raissi2019physicsinformed}
\textsc{Raissi M., Perdikaris P., Karniadakis G.~E.}:
\newblock Physics-informed neural networks: A deep learning framework for
  solving forward and inverse problems involving nonlinear partial differential
  equations.
\newblock \emph{Journal of Computational Physics 378} (2019), 686--707.

\bibitem[RPLG21]{reiser2021kilonerf}
\textsc{Reiser C., Peng S., Liao Y., Geiger A.}:
\newblock Kilonerf: Speeding up neural radiance fields with thousands of tiny
  mlps.
\newblock In \emph{Proceedings of the IEEE International Conference on Computer
  Vision (ICCV)} (2021).
\newblock URL: \url{http://arxiv.org/abs/2103.13744v2}.

\bibitem[RSH{\etalchar{*}}21]{reizenstein2021co3d}
\textsc{Reizenstein J., Shapovalov R., Henzler P., Sbordone L., Labatut P.,
  Novotny D.}:
\newblock Common objects in 3d: Large-scale learning and evaluation of
  real-life 3d category reconstruction.
\newblock In \emph{Proceedings of the IEEE/CVF Conference on Computer Vision
  and Pattern Recognition (CVPR)} (2021).
\newblock URL: \url{http://arxiv.org/abs/2109.00512v1}.

\bibitem[RTE{\etalchar{*}}21]{ramon2021h3dnet}
\textsc{Ramon E., Triginer G., Escur J., Pumarola A., Garcia J., i~Nieto X.~G.,
  Moreno-Noguer F.}:
\newblock H3d-net: Few-shot high-fidelity 3d head reconstruction.
\newblock \emph{arXiv preprint arXiv:2107.12512} (2021).
\newblock URL: \url{http://arxiv.org/abs/2107.12512v1}.

\bibitem[RW80]{rubin19803}
\textsc{Rubin S.~M., Whitted T.}:
\newblock A 3-dimensional representation for fast rendering of complex scenes.
\newblock In \emph{Proceedings of the 7th annual conference on Computer
  graphics and interactive techniques} (1980), pp.~110--116.

\bibitem[RWG{\etalchar{*}}13]{ren2013global}
\textsc{Ren P., Wang J., Gong M., Lin S., Tong X., Guo B.}:
\newblock Global illumination with radiance regression functions.
\newblock \emph{ACM Trans. Graph. 32}, 4 (jul 2013).
\newblock URL: \url{https://doi.org/10.1145/2461912.2462009}, \href
  {https://doi.org/10.1145/2461912.2462009}
  {\path{doi:10.1145/2461912.2462009}}.

\bibitem[RZS{\etalchar{*}}21]{raj2021pva}
\textsc{Raj A., Zollhoefer M., Simon T., Saragih J., Saito S., Hays J.,
  Lombardi S.}:
\newblock Pva: Pixel-aligned volumetric avatars.
\newblock In \emph{Proceedings of the IEEE/CVF Conference on Computer Vision
  and Pattern Recognition (CVPR)} (2021).
\newblock URL: \url{http://arxiv.org/abs/2101.02697v1}.

\bibitem[Sab88]{sabella1988rendering}
\textsc{Sabella P.}:
\newblock A rendering algorithm for visualizing 3d scalar fields.
\newblock In \emph{Proceedings of the 15th annual conference on Computer
  graphics and interactive techniques} (1988), pp.~51--58.

\bibitem[SAR20]{smith2020eikonet}
\textsc{Smith J.~D., Azizzadenesheli K., Ross Z.~E.}:
\newblock Eikonet: Solving the eikonal equation with deep neural networks.
\newblock URL: \url{http://arxiv.org/abs/2004.00361v3}.

\bibitem[SC21]{su2021affective}
\textsc{Su N.~M., Crandall D.~J.}:
\newblock The affective growth of computer vision.
\newblock In \emph{Proceedings of the IEEE/CVF Conference on Computer Vision
  and Pattern Recognition} (2021), pp.~9291--9300.

\bibitem[SCA19]{sbai2019unsupervised}
\textsc{Sbai O., Couprie C., Aubry M.}:
\newblock Unsupervised image decomposition in vector layers, 2019.

\bibitem[SCT{\etalchar{*}}20]{sitzmann2020metasdf}
\textsc{Sitzmann V., Chan E.~R., Tucker R., Snavely N., Wetzstein G.}:
\newblock Metasdf: Meta-learning signed distance functions.
\newblock In \emph{Advances in Neural Information Processing Systems (NeurIPS)}
  (2020), Curran Associates, Inc.
\newblock URL: \url{http://arxiv.org/abs/2006.09662v1}.

\bibitem[SDZ{\etalchar{*}}21]{srinivasan2021nerv}
\textsc{Srinivasan P.~P., Deng B., Zhang X., Tancik M., Mildenhall B., Barron
  J.~T.}:
\newblock Nerv: Neural reflectance and visibility fields for relighting and
  view synthesis.
\newblock In \emph{Proceedings of the IEEE/CVF Conference on Computer Vision
  and Pattern Recognition (CVPR)} (2021).
\newblock URL: \url{http://arxiv.org/abs/2012.03927v1}.

\bibitem[SF16]{schoenberger2016sfm}
\textsc{Sch\"{o}nberger J.~L., Frahm J.-M.}:
\newblock Structure-from-motion revisited.
\newblock In \emph{Conference on Computer Vision and Pattern Recognition
  (CVPR)} (2016).

\bibitem[SFE{\etalchar{*}}20]{sutanto2020learning}
\textsc{Sutanto G., Fernandez I. M.~R., Englert P., Ramachandran R.~K.,
  Sukhatme G.~S.}:
\newblock Learning equality constraints for motion planning on manifolds.
\newblock In \emph{Proceedings of the Conference on Robot Learning (CoRL)}
  (2020).
\newblock URL: \url{http://arxiv.org/abs/2009.11852v1}.

\bibitem[SGZ{\etalchar{*}}21]{shaham2021asapnet}
\textsc{Shaham T.~R., Gharbi M., Zhang R., Shechtman E., Michaeli T.}:
\newblock Spatially-adaptive pixelwise networks for fast image translation.
\newblock In \emph{Proceedings of the IEEE/CVF Conference on Computer Vision
  and Pattern Recognition (CVPR)} (2021).
\newblock URL: \url{http://arxiv.org/abs/2012.02992v1}.

\bibitem[Sha56]{shannon1956bandwagon}
\textsc{Shannon C.~E.}:
\newblock The bandwagon.
\newblock \emph{IRE transactions on Information Theory 2}, 1 (1956), 3.

\bibitem[SHN{\etalchar{*}}19]{saito2019pifu}
\textsc{Saito S., Huang Z., Natsume R., Morishima S., Kanazawa A., Li H.}:
\newblock Pifu: Pixel-aligned implicit function for high-resolution clothed
  human digitization.
\newblock In \emph{Proceedings of the IEEE International Conference on Computer
  Vision (ICCV)} (2019).
\newblock URL: \url{http://arxiv.org/abs/1905.05172v3}.

\bibitem[Sib81]{sibson1981brief}
\textsc{Sibson R.}:
\newblock A brief description of natural neighbour interpolation.
\newblock \emph{Interpreting multivariate data} (1981).

\bibitem[SIE21]{skorokhodov2021inrgan}
\textsc{Skorokhodov I., Ignatyev S., Elhoseiny M.}:
\newblock Adversarial generation of continuous images.
\newblock In \emph{Proceedings of the IEEE/CVF Conference on Computer Vision
  and Pattern Recognition (CVPR)} (2021).
\newblock URL: \url{http://arxiv.org/abs/2011.12026v2}.

\bibitem[Sit]{Sitzmann_Awesome_Implicit_Representations}
\textsc{Sitzmann V.}:
\newblock Awesome implicit representations - a curated list of resources on
  implicit neural representations.
\newblock URL:
  \url{https://github.com/vsitzmann/awesome-implicit-representations}.

\bibitem[SJ22]{sharp2022spelunkingthedeep}
\textsc{Sharp N., Jacobson A.}:
\newblock Spelunking the deep: Guaranteed queries for general neural implicit
  surfaces.
\newblock URL: \url{http://arxiv.org/abs/2202.02444v1}.

\bibitem[SJK21]{shukla2021parallel}
\textsc{Shukla K., Jagtap A.~D., Karniadakis G.~E.}:
\newblock Parallel physics-informed neural networks via domain decomposition.
\newblock \emph{arXiv preprint arXiv:2104.10013} (2021).
\newblock URL: \url{http://arxiv.org/abs/2104.10013v3}.

\bibitem[SJL{\etalchar{*}}21]{suo2021neuralhumanfvv}
\textsc{Suo X., Jiang Y., Lin P., Zhang Y., Wu M., Guo K., Xu L.}:
\newblock Neuralhumanfvv: Real-time neural volumetric human performance
  rendering using rgb cameras.
\newblock In \emph{Proceedings of the IEEE/CVF Conference on Computer Vision
  and Pattern Recognition} (2021), pp.~6226--6237.

\bibitem[SK97]{schluns1997local}
\textsc{Schl{\"u}ns K., Klette R.}:
\newblock Local and global integration of discrete vector fields.
\newblock In \emph{Advances in Computer Vision}. Springer, 1997, pp.~149--158.

\bibitem[SKK21]{stelzner2021obsurf}
\textsc{Stelzner K., Kersting K., Kosiorek A.~R.}:
\newblock Decomposing 3d scenes into objects via unsupervised volume
  segmentation.
\newblock \emph{arXiv preprint arXiv:2104.01148} (2021).
\newblock URL: \url{http://arxiv.org/abs/2104.01148v1}.

\bibitem[SLB{\etalchar{*}}21]{sun2021nelf}
\textsc{Sun T., Lin K.-E., Bi S., Xu Z., Ramamoorthi R.}:
\newblock Nelf: Neural light-transport field for portrait view synthesis and
  relighting.
\newblock \emph{Computer Graphics Forum} (2021).
\newblock URL: \url{http://arxiv.org/abs/2107.12351v1}.

\bibitem[SLNG20]{schwarz2020graf}
\textsc{Schwarz K., Liao Y., Niemeyer M., Geiger A.}:
\newblock Graf: Generative radiance fields for 3d-aware image synthesis.
\newblock In \emph{Advances in Neural Information Processing Systems (NeurIPS)}
  (2020), Curran Associates, Inc.
\newblock URL: \url{http://arxiv.org/abs/2007.02442v4}.

\bibitem[SLOD21]{sucar2021imap}
\textsc{Sucar E., Liu S., Ortiz J., Davison A.~J.}:
\newblock imap: Implicit mapping and positioning in real-time.
\newblock In \emph{Proceedings of the IEEE International Conference on Computer
  Vision (ICCV)} (2021).
\newblock URL: \url{http://arxiv.org/abs/2103.12352v2}.

\bibitem[SLX{\etalchar{*}}21]{9606601}
\textsc{Sun Y., Liu J., Xie M., Wohlberg B., Kamilov U.~S.}:
\newblock Coil: Coordinate-based internal learning for tomographic imaging.
\newblock \emph{IEEE Transactions on Computational Imaging 7} (2021),
  1400--1412.
\newblock \href {https://doi.org/10.1109/TCI.2021.3125564}
  {\path{doi:10.1109/TCI.2021.3125564}}.

\bibitem[SMB{\etalchar{*}}20]{sitzmann2020siren}
\textsc{Sitzmann V., Martel J. N.~P., Bergman A.~W., Lindell D.~B., Wetzstein
  G.}:
\newblock Implicit neural representations with periodic activation functions.
\newblock In \emph{Advances in Neural Information Processing Systems (NeurIPS)}
  (2020), Curran Associates, Inc.
\newblock URL: \url{http://arxiv.org/abs/2006.09661v1}.

\bibitem[SOHW12]{sullivan2012hevc}
\textsc{Sullivan G.~J., Ohm J.-R., Han W.-J., Wiegand T.}:
\newblock Overview of the high efficiency video coding (hevc) standard.
\newblock \emph{IEEE Transactions on circuits and systems for video technology
  22}, 12 (2012), 1649--1668.

\bibitem[SPB{\etalchar{*}}18]{schwarz2018emerging}
\textsc{Schwarz S., Preda M., Baroncini V., Budagavi M., Cesar P., Chou P.~A.,
  Cohen R.~A., Krivoku{\'c}a M., Lasserre S., Li Z., et~al.}:
\newblock Emerging mpeg standards for point cloud compression.
\newblock \emph{IEEE Journal on Emerging and Selected Topics in Circuits and
  Systems 9}, 1 (2018), 133--148.

\bibitem[SPJ{\etalchar{*}}22]{sajnani2022condor}
\textsc{Sajnani R., Poulenard A., Jain J., Dua R., Guibas L.~J., Sridhar S.}:
\newblock Condor: Self-supervised canonicalization of 3d pose for partial
  shapes, 2022.

\bibitem[SPX21]{shen2021nerp}
\textsc{Shen L., Pauly J., Xing L.}:
\newblock Nerp: Implicit neural representation learning with prior embedding
  for sparsely sampled image reconstruction.
\newblock \emph{arXiv preprint arXiv:ers/2108/2108.10991} (2021).
\newblock URL: \url{http://arxiv.org/abs/2108.10991v1}.

\bibitem[SRAMN21]{shen2021snerf}
\textsc{Shen J., Ruiz A., Agudo A., Moreno-Noguer F.}:
\newblock Stochastic neural radiance fields: Quantifying uncertainty in
  implicit 3d representations.
\newblock \emph{arXiv preprint arXiv:2109.02123} (2021).
\newblock URL: \url{http://arxiv.org/abs/2109.02123v3}.

\bibitem[SRF{\etalchar{*}}21]{sitzmann2021lfns}
\textsc{Sitzmann V., Rezchikov S., Freeman W.~T., Tenenbaum J.~B., Durand F.}:
\newblock Light field networks: Neural scene representations with
  single-evaluation rendering.
\newblock In \emph{Advances in Neural Information Processing (NeurIPS)} (2021).
\newblock URL: \url{http://arxiv.org/abs/2106.02634v1}.

\bibitem[SSM{\etalchar{*}}20]{spezialetti2020learning}
\textsc{Spezialetti R., Stella F., Marcon M., Silva L., Salti S., Stefano
  L.~D.}:
\newblock Learning to orient surfaces by self-supervised spherical cnns.
\newblock \emph{Advances in Neural Information Processing Systems 33} (2020).

\bibitem[SSSJ20]{saito2020pifuhd}
\textsc{Saito S., Simon T., Saragih J., Joo H.}:
\newblock Pifuhd: Multi-level pixel-aligned implicit function for
  high-resolution 3d human digitization.
\newblock In \emph{Proceedings of the IEEE/CVF Conference on Computer Vision
  and Pattern Recognition (CVPR)} (2020).
\newblock URL: \url{http://arxiv.org/abs/2004.00452v1}.

\bibitem[Sta07]{stanley2007compositional}
\textsc{Stanley K.~O.}:
\newblock Compositional pattern producing networks: A novel abstraction of
  development.
\newblock \emph{Genetic programming and evolvable machines 8}, 2 (2007),
  131--162.

\bibitem[STD{\etalchar{*}}21]{sun2021canonical}
\textsc{Sun W., Tagliasacchi A., Deng B., Sabour S., Yazdani S., Hinton G.~E.,
  Yi K.~M.}:
\newblock Canonical capsules: Self-supervised capsules in canonical pose.
\newblock \emph{Advances in Neural Information Processing Systems 34} (2021).

\bibitem[STE21]{skorokhodov2022styleganv}
\textsc{Skorokhodov I., Tulyakov S., Elhoseiny M.}:
\newblock Stylegan-v: A continuous video generator with the price, image
  quality and perks of stylegan2.
\newblock URL: \url{http://arxiv.org/abs/2112.14683v1}.

\bibitem[STH{\etalchar{*}}19]{sitzmann2019deepvoxels}
\textsc{Sitzmann V., Thies J., Heide F., Nie{\ss}ner M., Wetzstein G.,
  Zollh{\"o}fer M.}:
\newblock Deepvoxels: Learning persistent 3d feature embeddings.
\newblock In \emph{Proc. Computer Vision and Pattern Recognition (CVPR), IEEE}
  (2019).

\bibitem[SWL{\etalchar{*}}21]{shen2021nonlineofsight}
\textsc{Shen S., Wang Z., Liu P., Pan Z., Li R., Gao T., Li S., Yu J.}:
\newblock Non-line-of-sight imaging via neural transient fields.
\newblock \emph{IEEE Transactions on Pattern Analysis and Machine Intelligence}
  (2021).
\newblock URL: \url{http://arxiv.org/abs/2101.00373v3}.

\bibitem[SXC{\etalchar{*}}21]{sun2021neuralrecon}
\textsc{Sun J., Xie Y., Chen L., Zhou X., Bao H.}:
\newblock Neuralrecon: Real-time coherent 3d reconstruction from monocular
  video.
\newblock In \emph{Proceedings of the IEEE/CVF Conference on Computer Vision
  and Pattern Recognition (CVPR)} (2021).
\newblock URL: \url{http://arxiv.org/abs/2104.00681v1}.

\bibitem[SYMB21]{saito2021scanimate}
\textsc{Saito S., Yang J., Ma Q., Black M.~J.}:
\newblock Scanimate: Weakly supervised learning of skinned clothed avatar
  networks.
\newblock In \emph{Proceedings of the IEEE/CVF Conference on Computer Vision
  and Pattern Recognition (CVPR)} (2021).
\newblock URL: \url{http://arxiv.org/abs/2104.03313v2}.

\bibitem[SYZR21]{su2021anerf}
\textsc{Su S.-Y., Yu F., Zollhoefer M., Rhodin H.}:
\newblock A-nerf: Surface-free human 3d pose refinement via neural rendering.
\newblock \emph{arXiv preprint arXiv:2102.06199} (2021).
\newblock URL: \url{http://arxiv.org/abs/2102.06199v1}.

\bibitem[SZPF16]{schoenberger2016mvs}
\textsc{Sch\"{o}nberger J.~L., Zheng E., Pollefeys M., Frahm J.-M.}:
\newblock Pixelwise view selection for unstructured multi-view stereo.
\newblock In \emph{European Conference on Computer Vision (ECCV)} (2016).

\bibitem[SZW19]{sitzmann2019srn}
\textsc{Sitzmann V., Zollhofer M., Wetzstein G.}:
\newblock Scene representation networks: Continuous 3d-structure-aware neural
  scene representations.
\newblock In \emph{Advances in Neural Information Processing Systems (NeurIPS)}
  (2019), Curran Associates, Inc.
\newblock URL: \url{http://arxiv.org/abs/1906.01618v2}.

\bibitem[SZZ{\etalchar{*}}21]{shao2021doublefield}
\textsc{Shao R., Zhang H., Zhang H., Cao Y., Yu T., Liu Y.}:
\newblock Doublefield: Bridging the neural surface and radiance fields for
  high-fidelity human rendering.
\newblock \emph{arXiv preprint arXiv:2106.03798} (2021).
\newblock URL: \url{http://arxiv.org/abs/2106.03798v2}.

\bibitem[TFT{\etalchar{*}}20]{tewari2020state}
\textsc{Tewari A., Fried O., Thies J., Sitzmann V., Lombardi S., Sunkavalli K.,
  Martin-Brualla R., Simon T., Saragih J., Nie{\ss}ner M., et~al.}:
\newblock State of the art on neural rendering.
\newblock In \emph{Computer Graphics Forum} (2020), vol.~39, Wiley Online
  Library, pp.~701--727.

\bibitem[Thr02]{thrun2002probabilistic}
\textsc{Thrun S.}:
\newblock Probabilistic robotics.
\newblock \emph{Communications of the ACM 45}, 3 (2002), 52--57.

\bibitem[TLV21]{tschernezki2021neuraldiff}
\textsc{Tschernezki V., Larlus D., Vedaldi A.}:
\newblock Neuraldiff: Segmenting 3d objects that move in egocentric videos.
\newblock \emph{arXiv preprint arXiv:2110.09936} (2021).

\bibitem[TLY{\etalchar{*}}21]{takikawa2021nglod}
\textsc{Takikawa T., Litalien J., Yin K., Kreis K., Loop C., Nowrouzezahrai D.,
  Jacobson A., McGuire M., Fidler S.}:
\newblock Neural geometric level of detail: Real-time rendering with implicit
  3d shapes.
\newblock In \emph{Proceedings of the IEEE/CVF Conference on Computer Vision
  and Pattern Recognition (CVPR)} (2021).
\newblock URL: \url{http://arxiv.org/abs/2101.10994v1}.

\bibitem[TMW{\etalchar{*}}21]{tancik2021learned}
\textsc{Tancik M., Mildenhall B., Wang T., Schmidt D., Srinivasan P.~P., Barron
  J.~T., Ng R.}:
\newblock Learned initializations for optimizing coordinate-based neural
  representations.
\newblock In \emph{Proceedings of the IEEE/CVF Conference on Computer Vision
  and Pattern Recognition (CVPR)} (2021).
\newblock URL: \url{http://arxiv.org/abs/2012.02189v2}.

\bibitem[TSM{\etalchar{*}}20]{tancik2020ffn}
\textsc{Tancik M., Srinivasan P.~P., Mildenhall B., Fridovich-Keil S., Raghavan
  N., Singhal U., Ramamoorthi R., Barron J.~T., Ng R.}:
\newblock Fourier features let networks learn high frequency functions in low
  dimensional domains.
\newblock In \emph{Advances in Neural Information Processing Systems (NeurIPS)}
  (2020), Curran Associates, Inc.
\newblock URL: \url{http://arxiv.org/abs/2006.10739v1}.

\bibitem[TSTPM21]{tiwari2021neuralgif}
\textsc{Tiwari G., Sarafianos N., Tung T., Pons-Moll G.}:
\newblock Neural-gif: Neural generalized implicit functions for animating
  people in clothing.
\newblock In \emph{Proceedings of the IEEE International Conference on Computer
  Vision (ICCV)} (2021).
\newblock URL: \url{http://arxiv.org/abs/2108.08807v2}.

\bibitem[TSV19]{tesfaldet2019fcppn}
\textsc{Tesfaldet M., Snelgrove X., Vazquez D.}:
\newblock Fourier-cppns for image synthesis.
\newblock URL: \url{http://arxiv.org/abs/1909.09273v1}.

\bibitem[TTG{\etalchar{*}}20]{tretschk2020patchnets}
\textsc{Tretschk E., Tewari A., Golyanik V., Zollhofer M., Stoll C., Theobalt
  C.}:
\newblock Patchnets: Patch-based generalizable deep implicit 3d shape
  representations.
\newblock In \emph{Proceedings of the European Conference on Computer Vision
  (ECCV)} (2020).
\newblock URL: \url{http://arxiv.org/abs/2008.01639v2}.

\bibitem[TTG{\etalchar{*}}21]{tretschk2021nrnerf}
\textsc{Tretschk E., Tewari A., Golyanik V., Zollhofer M., Lassner C., Theobalt
  C.}:
\newblock Non-rigid neural radiance fields: Reconstruction and novel view
  synthesis of a dynamic scene from monocular video.
\newblock In \emph{Proceedings of the IEEE International Conference on Computer
  Vision (ICCV)} (2021).
\newblock URL: \url{http://arxiv.org/abs/2012.12247v4}.

\bibitem[TTM{\etalchar{*}}21]{tewari2021advances}
\textsc{Tewari A., Thies J., Mildenhall B., Srinivasan P., Tretschk E., Wang
  Y., Lassner C., Sitzmann V., Martin-Brualla R., Lombardi S., Simon T.,
  Theobalt C., Niessner M., Barron J.~T., Wetzstein G., Zollhoefer M., Golyanik
  V.}:
\newblock Advances in neural rendering.
\newblock URL: \url{http://arxiv.org/abs/2111.05849v1}.

\bibitem[TVRF{\etalchar{*}}20]{tolosana2020deepfakes}
\textsc{Tolosana R., Vera-Rodriguez R., Fierrez J., Morales A., Ortega-Garcia
  J.}:
\newblock Deepfakes and beyond: A survey of face manipulation and fake
  detection.
\newblock \emph{Information Fusion 64} (2020), 131--148.

\bibitem[TY21]{trevithick2021grf}
\textsc{Trevithick A., Yang B.}:
\newblock Grf: Learning a general radiance field for 3d representation and
  rendering.
\newblock In \emph{Proceedings of the IEEE International Conference on Computer
  Vision (ICCV)} (2021).
\newblock URL: \url{http://arxiv.org/abs/2010.04595v3}.

\bibitem[vdODZ{\etalchar{*}}16]{oord2016wavenet}
\textsc{van~den Oord A., Dieleman S., Zen H., Simonyan K., Vinyals O., Graves
  A., Kalchbrenner N., Senior A., Kavukcuoglu K.}:
\newblock Wavenet: A generative model for raw audio.
\newblock \emph{arXiv preprint arXiv:1609.03499} (2016).

\bibitem[VJK21]{vicini2021non}
\textsc{Vicini D., Jakob W., Kaplanyan A.}:
\newblock A non-exponential transmittance model for volumetric scene
  representations.
\newblock \emph{ACM Transactions on Graphics (TOG) 40}, 4 (2021), 1--16.

\bibitem[VKS{\etalchar{*}}21]{Venkatesh_2021_ICCV}
\textsc{Venkatesh R., Karmali T., Sharma S., Ghosh A., Babu R.~V., Jeni L.~A.,
  Singh M.}:
\newblock Deep implicit surface point prediction networks.
\newblock In \emph{Proceedings of the IEEE/CVF International Conference on
  Computer Vision (ICCV)} (October 2021), pp.~12653--12662.

\bibitem[VRG{\etalchar{*}}21]{vora2021nesf}
\textsc{Vora S., Radwan N., Greff K., Meyer H., Genova K., Sajjadi M.~S., Pot
  E., Tagliasacchi A., Duckworth D.}:
\newblock Nesf: Neural semantic fields for generalizable semantic segmentation
  of 3d scenes.
\newblock \emph{arXiv preprint arXiv:2111.13260} (2021).

\bibitem[Wal92]{wallace1992jpeg}
\textsc{Wallace G.}:
\newblock The jpeg still picture compression standard.
\newblock \emph{IEEE Transactions on Consumer Electronics 38}, 1 (1992),
  xviii--xxxiv.
\newblock \href {https://doi.org/10.1109/30.125072}
  {\path{doi:10.1109/30.125072}}.

\bibitem[War92]{ward1992measuring}
\textsc{Ward G.~J.}:
\newblock Measuring and modeling anisotropic reflection.
\newblock In \emph{Proceedings of the 19th annual conference on Computer
  graphics and interactive techniques} (1992), pp.~265--272.

\bibitem[WBL{\etalchar{*}}21]{wang2021hybridnerf}
\textsc{Wang Z., Bagautdinov T., Lombardi S., Simon T., Saragih J., Hodgins J.,
  Zollhofer M.}:
\newblock Learning compositional radiance fields of dynamic human heads.
\newblock In \emph{Proceedings of the IEEE/CVF Conference on Computer Vision
  and Pattern Recognition (CVPR)} (2021).
\newblock URL: \url{http://arxiv.org/abs/2012.09955v1}.

\bibitem[WELG21]{wang2021dctnerf}
\textsc{Wang C., Eckart B., Lucey S., Gallo O.}:
\newblock Neural trajectory fields for dynamic novel view synthesis.
\newblock \emph{arXiv preprint arXiv:2105.05994} (2021).
\newblock URL: \url{http://arxiv.org/abs/2105.05994v1}.

\bibitem[WGK{\etalchar{*}}21]{williams2021neuralkernelfields(nkf)}
\textsc{Williams F., Gojcic Z., Khamis S., Zorin D., Bruna J., Fidler S.,
  Litany O.}:
\newblock Neural fields as learnable kernels for 3d reconstruction.
\newblock URL: \url{http://arxiv.org/abs/2111.13674v1}.

\bibitem[WGT21]{wang2021locally}
\textsc{Wang S., Geiger A., Tang S.}:
\newblock Locally aware piecewise transformation fields for 3d human mesh
  registration.
\newblock In \emph{Proceedings of the IEEE/CVF Conference on Computer Vision
  and Pattern Recognition (CVPR)} (2021).
\newblock URL: \url{http://arxiv.org/abs/2104.08160v1}.

\bibitem[WLL{\etalchar{*}}21]{wang2021neus}
\textsc{Wang P., Liu L., Liu Y., Theobalt C., Komura T., Wang W.}:
\newblock Neus: Learning neural implicit surfaces by volume rendering for
  multi-view reconstruction.
\newblock In \emph{Proceedings of the Thirtieth International Joint Conference
  on Artificial Intelligence (IJCAI)} (2021), International Joint Conferences
  on Artificial Intelligence Organization.
\newblock URL: \url{http://arxiv.org/abs/2106.10689v1}.

\bibitem[WLR{\etalchar{*}}21]{wei2021nerfingmvs}
\textsc{Wei Y., Liu S., Rao Y., Zhao W., Lu J., Zhou J.}:
\newblock Nerfingmvs: Guided optimization of neural radiance fields for indoor
  multi-view stereo.
\newblock URL: \url{http://arxiv.org/abs/2109.01129v2}.

\bibitem[WLX{\etalchar{*}}21]{wu2021irem}
\textsc{Wu Q., Li Y., Xu L., Feng R., Wei H., Yang Q., Yu B., Liu X., Yu J.,
  Zhang Y.}:
\newblock Irem: High-resolution magnetic resonance (mr) image reconstruction
  via implicit neural representation.
\newblock URL: \url{http://arxiv.org/abs/2106.15097v1}.

\bibitem[WLYT21]{wang2021spline}
\textsc{Wang P.-S., Liu Y., Yang Y.-Q., Tong X.}:
\newblock Spline positional encoding for learning 3d implicit signed distance
  fields.
\newblock In \emph{Proceedings of the Thirtieth International Joint Conference
  on Artificial Intelligence (IJCAI)} (2021), International Joint Conferences
  on Artificial Intelligence Organization.
\newblock URL: \url{http://arxiv.org/abs/2106.01553v1}.

\bibitem[WMLT07]{walter2007microfacet}
\textsc{Walter B., Marschner S.~R., Li H., Torrance K.~E.}:
\newblock Microfacet models for refraction through rough surfaces.
\newblock \emph{Rendering techniques 2007} (2007), 18th.

\bibitem[WMM{\etalchar{*}}21]{wang2021metaavatar}
\textsc{Wang S., Mihajlovic M., Ma Q., Geiger A., Tang S.}:
\newblock Metaavatar: Learning animatable clothed human models from few depth
  images.
\newblock In \emph{Proceedings of the Thirtieth International Joint Conference
  on Artificial Intelligence (IJCAI)} (2021), International Joint Conferences
  on Artificial Intelligence Organization.
\newblock URL: \url{http://arxiv.org/abs/2106.11944v1}.

\bibitem[WPLN{\etalchar{*}}20]{williams2020voronoinet}
\textsc{Williams F., Parent-Levesque J., Nowrouzezahrai D., Panozzo D., Yi
  K.~M., Tagliasacchi A.}:
\newblock Voronoinet: General functional approximators with local support.
\newblock In \emph{Proceedings of the IEEE/CVF Conference on Computer Vision
  and Pattern Recognition Workshops} (2020), pp.~264--265.

\bibitem[WPYS21]{wizadwongsa2021nex}
\textsc{Wizadwongsa S., Phongthawee P., Yenphraphai J., Suwajanakorn S.}:
\newblock Nex: Real-time view synthesis with neural basis expansion.
\newblock In \emph{Proceedings of the IEEE/CVF Conference on Computer Vision
  and Pattern Recognition (CVPR)} (2021).
\newblock URL: \url{http://arxiv.org/abs/2103.05606v2}.

\bibitem[WSH{\etalchar{*}}19]{wang2019normalized}
\textsc{Wang H., Sridhar S., Huang J., Valentin J., Song S., Guibas L.~J.}:
\newblock Normalized object coordinate space for category-level 6d object pose
  and size estimation.
\newblock In \emph{Proceedings of the IEEE/CVF Conference on Computer Vision
  and Pattern Recognition} (2019), pp.~2642--2651.

\bibitem[WSS{\etalchar{*}}19]{williams2019deep}
\textsc{Williams F., Schneider T., Silva C., Zorin D., Bruna J., Panozzo D.}:
\newblock Deep geometric prior for surface reconstruction.
\newblock In \emph{Proceedings of the IEEE/CVF Conference on Computer Vision
  and Pattern Recognition (CVPR)} (2019).
\newblock URL: \url{http://arxiv.org/abs/1811.10943v2}.

\bibitem[WTBZ21]{williams2021neuralsplines}
\textsc{Williams F., Trager M., Bruna J., Zorin D.}:
\newblock Neural splines: Fitting 3d surfaces with infinitely-wide neural
  networks.
\newblock In \emph{Proceedings of the IEEE/CVF Conference on Computer Vision
  and Pattern Recognition (CVPR)} (2021).
\newblock URL: \url{http://arxiv.org/abs/2006.13782v3}.

\bibitem[WWB{\etalchar{*}}14]{wald2014embree}
\textsc{Wald I., Woop S., Benthin C., Johnson G.~S., Ernst M.}:
\newblock Embree: a kernel framework for efficient cpu ray tracing.
\newblock \emph{ACM Transactions on Graphics (TOG) 33}, 4 (2014), 1--8.

\bibitem[WWG{\etalchar{*}}21]{wang2021ibrnet}
\textsc{Wang Q., Wang Z., Genova K., Srinivasan P., Zhou H., Barron J.~T.,
  Martin-Brualla R., Snavely N., Funkhouser T.}:
\newblock Ibrnet: Learning multi-view image-based rendering.
\newblock In \emph{Proceedings of the IEEE/CVF Conference on Computer Vision
  and Pattern Recognition (CVPR)} (2021).
\newblock URL: \url{http://arxiv.org/abs/2102.13090v2}.

\bibitem[WWX{\etalchar{*}}21]{wang2021nerf--}
\textsc{Wang Z., Wu S., Xie W., Chen M., Prisacariu V.~A.}:
\newblock Nerf--: Neural radiance fields without known camera parameters.
\newblock \emph{arXiv preprint arXiv:2102.07064} (2021).
\newblock URL: \url{http://arxiv.org/abs/2102.07064v3}.

\bibitem[WWZ{\etalchar{*}}21]{wang2021mirrornerf}
\textsc{Wang Z., Wang L., Zhao F., Wu M., Xu L., Yu J.}:
\newblock Mirrornerf: One-shot neural portrait radiance field from multi-mirror
  catadioptric imaging.
\newblock \emph{arXiv preprint arXiv:2104.02607} (2021).
\newblock URL: \url{http://arxiv.org/abs/2104.02607v2}.

\bibitem[WXB21]{wu2021learning}
\textsc{Wu Z.-F., Xue H., Bai W.}:
\newblock Learning deeper non-monotonic networks by softly transferring
  solution space.
\newblock In \emph{Proceedings of the Thirtieth International Joint Conference
  on Artificial Intelligence (IJCAI)} (2021), International Joint Conferences
  on Artificial Intelligence Organization.

\bibitem[WYN21]{wang2021sceneformer}
\textsc{Wang X., Yeshwanth C., Nie{\ss}ner M.}:
\newblock Sceneformer: Indoor scene generation with transformers.
\newblock In \emph{2021 International Conference on 3D Vision (3DV)} (2021),
  IEEE, pp.~106--115.

\bibitem[WZB22]{wolterink2021implicit}
\textsc{Wolterink J.~M., Zwienenberg J.~C., Brune C.}:
\newblock Implicit neural representations for deformable image registration.
\newblock In \emph{Medical Imaging with Deep Learning} (2022).

\bibitem[XAS21]{xu2021hnerf}
\textsc{Xu H., Alldieck T., Sminchisescu C.}:
\newblock H-nerf: Neural radiance fields for rendering and temporal
  reconstruction of humans in motion.
\newblock URL: \url{http://arxiv.org/abs/2110.13746v2}.

\bibitem[XFYS20]{xu2020ladybird}
\textsc{Xu Y., Fan T., Yuan Y., Singh G.}:
\newblock Ladybird: Quasi-monte carlo sampling for deep implicit field based 3d
  reconstruction with symmetry.
\newblock In \emph{Proceedings of the European Conference on Computer Vision
  (ECCV)} (2020).
\newblock URL: \url{http://arxiv.org/abs/2007.13393v1}.

\bibitem[XHKK21]{xian2021spacetime}
\textsc{Xian W., Huang J.-B., Kopf J., Kim C.}:
\newblock Space-time neural irradiance fields for free-viewpoint video.
\newblock In \emph{Proceedings of the IEEE/CVF Conference on Computer Vision
  and Pattern Recognition (CVPR)} (2021).
\newblock URL: \url{http://arxiv.org/abs/2011.12950v2}.

\bibitem[XPLD21]{xu2021generative}
\textsc{Xu X., Pan X., Lin D., Dai B.}:
\newblock Generative occupancy fields for 3d surface-aware image synthesis.
\newblock In \emph{Advances in Neural Information Processing Systems (NeurIPS)}
  (2021).

\bibitem[XPMBB21]{xie2021fignerf}
\textsc{Xie C., Park K., Martin-Brualla R., Brown M.}:
\newblock Fig-nerf: Figure-ground neural radiance fields for 3d object category
  modelling.
\newblock \emph{arXiv preprint arXiv:2104.08418} (2021).
\newblock URL: \url{http://arxiv.org/abs/2104.08418v1}.

\bibitem[XWC{\etalchar{*}}19]{xu2019disn}
\textsc{Xu Q., Wang W., Ceylan D., Mech R., Neumann U.}:
\newblock Disn: Deep implicit surface network for high-quality single-view 3d
  reconstruction.
\newblock In \emph{Advances in Neural Information Processing Systems (NeurIPS)}
  (2019), Curran Associates, Inc.
\newblock URL: \url{http://arxiv.org/abs/1905.10711v2}.

\bibitem[XXH{\etalchar{*}}21]{xiang2021neutex}
\textsc{Xiang F., Xu Z., Hasan M., Hold-Geoffroy Y., Sunkavalli K., Su H.}:
\newblock Neutex: Neural texture mapping for volumetric neural rendering.
\newblock In \emph{Proceedings of the IEEE/CVF Conference on Computer Vision
  and Pattern Recognition (CVPR)} (2021).
\newblock URL: \url{http://arxiv.org/abs/2103.00762v1}.

\bibitem[YBHK21]{yang2021nfgp}
\textsc{Yang G., Belongie S., Hariharan B., Koltun V.}:
\newblock Geometry processing with neural fields.
\newblock \emph{Advances in Neural Information Processing Systems 34} (2021).

\bibitem[YCFB{\etalchar{*}}20]{yen-chen2021inerf}
\textsc{Yen-Chen L., Florence P., Barron J.~T., Rodriguez A., Isola P., Lin
  T.-Y.}:
\newblock Inerf: Inverting neural radiance fields for pose estimation.
\newblock URL: \url{http://arxiv.org/abs/2012.05877v3}.

\bibitem[YFKT{\etalchar{*}}21]{yu2021plenoxels}
\textsc{Yu A., Fridovich-Keil S., Tancik M., Chen Q., Recht B., Kanazawa A.}:
\newblock Plenoxels: Radiance fields without neural networks.
\newblock \emph{arXiv preprint arXiv:2112.05131} (2021).

\bibitem[YFST18]{yang2018foldingnet}
\textsc{Yang Y., Feng C., Shen Y., Tian D.}:
\newblock Foldingnet: Point cloud auto-encoder via deep grid deformation.
\newblock In \emph{Proceedings of the IEEE conference on computer vision and
  pattern recognition} (2018), pp.~206--215.

\bibitem[YGKL21]{yariv2021volsdf}
\textsc{Yariv L., Gu J., Kasten Y., Lipman Y.}:
\newblock Volume rendering of neural implicit surfaces.
\newblock In \emph{Advances in Neural Information Processing Systems (NeurIPS)}
  (2021), Curran Associates, Inc.
\newblock URL: \url{http://arxiv.org/abs/2106.12052v1}.

\bibitem[YGW21]{yu2021uorf}
\textsc{Yu H.-X., Guibas L.~J., Wu J.}:
\newblock Unsupervised discovery of object radiance fields.
\newblock \emph{arXiv preprint arXiv:2107.07905} (2021).
\newblock URL: \url{http://arxiv.org/abs/2107.07905v1}.

\bibitem[YHA{\etalchar{*}}21]{yeung2021implicitvol}
\textsc{Yeung P.-H., Hesse L., Aliasi M., Haak M., the INTERGROWTH-21st
  Consortium, Xie W., Namburete A. I.~L.}:
\newblock Implicitvol: Sensorless 3d ultrasound reconstruction with deep
  implicit representation.
\newblock \emph{arXiv preprint arXiv:2109.12108} (2021).
\newblock URL: \url{http://arxiv.org/abs/2109.12108v1}.

\bibitem[YKM{\etalchar{*}}20]{yariv2020idr}
\textsc{Yariv L., Kasten Y., Moran D., Galun M., Atzmon M., Basri R., Lipman
  Y.}:
\newblock Multiview neural surface reconstruction by disentangling geometry and
  appearance.
\newblock In \emph{Advances in Neural Information Processing Systems (NeurIPS)}
  (2020), Curran Associates, Inc.
\newblock URL: \url{http://arxiv.org/abs/2003.09852v3}.

\bibitem[YLB{\etalchar{*}}20]{yang2020continuous}
\textsc{Yang Z., Litany O., Birdal T., Sridhar S., Guibas L.}:
\newblock Continuous geodesic convolutions for learning on 3d shapes, 2020.

\bibitem[YLM{\etalchar{*}}22]{yan2022shapeformer}
\textsc{Yan X., Lin L., Mitra N.~J., Lischinski D., Cohen-Or D., Huang H.}:
\newblock Shapeformer: Transformer-based shape completion via sparse
  representation, 2022.

\bibitem[YLSL21]{yuan2021star}
\textsc{Yuan W., Lv Z., Schmidt T., Lovegrove S.}:
\newblock Star: Self-supervised tracking and reconstruction of rigid objects in
  motion with neural rendering.
\newblock In \emph{Proceedings of the IEEE/CVF Conference on Computer Vision
  and Pattern Recognition (CVPR)} (2021).
\newblock URL: \url{http://arxiv.org/abs/2101.01602v1}.

\bibitem[YLT{\etalchar{*}}21a]{yu2021plenoctrees}
\textsc{Yu A., Li R., Tancik M., Li H., Ng R., Kanazawa A.}:
\newblock Plenoctrees for real-time rendering of neural radiance fields.
\newblock \emph{Proc. ICCV} (2021).

\bibitem[YLT{\etalchar{*}}21b]{yu2021nerfsh}
\textsc{Yu A., Li R., Tancik M., Li H., Ng R., Kanazawa A.}:
\newblock Plenoctrees for real-time rendering of neural radiance fields.
\newblock In \emph{Proceedings of the IEEE International Conference on Computer
  Vision (ICCV)} (2021).
\newblock URL: \url{http://arxiv.org/abs/2103.14024v2}.

\bibitem[YRSH21]{yifan2021idf}
\textsc{Yifan W., Rahmann L., Sorkine-Hornung O.}:
\newblock Geometry-consistent neural shape representation with implicit
  displacement fields.
\newblock In \emph{Proceedings of the Thirtieth International Joint Conference
  on Artificial Intelligence (IJCAI)} (2021), International Joint Conferences
  on Artificial Intelligence Organization.
\newblock URL: \url{http://arxiv.org/abs/2106.05187v2}.

\bibitem[YTB{\etalchar{*}}21]{yenamandra2021i3dmm}
\textsc{Yenamandra T., Tewari A., Bernard F., Seidel H.-P., Elgharib M.,
  Cremers D., Theobalt C.}:
\newblock i3dmm: Deep implicit 3d morphable model of human heads.
\newblock In \emph{Proceedings of the IEEE/CVF Conference on Computer Vision
  and Pattern Recognition (CVPR)} (2021).
\newblock URL: \url{http://arxiv.org/abs/2011.14143v1}.

\bibitem[YTS{\etalchar{*}}21]{yan2021continual}
\textsc{Yan Z., Tian Y., Shi X., Guo P., Wang P., Zha H.}:
\newblock Continual neural mapping: Learning an implicit scene representation
  from sequential observations.
\newblock In \emph{Proceedings of the IEEE International Conference on Computer
  Vision (ICCV)} (2021).
\newblock URL: \url{http://arxiv.org/abs/2108.05851v1}.

\bibitem[YWC{\etalchar{*}}21]{yang2021deep}
\textsc{Yang M., Wen Y., Chen W., Chen Y., Jia K.}:
\newblock Deep optimized priors for 3d shape modeling and reconstruction.
\newblock In \emph{Proceedings of the IEEE/CVF Conference on Computer Vision
  and Pattern Recognition (CVPR)} (2021).
\newblock URL: \url{http://arxiv.org/abs/2012.07241v1}.

\bibitem[YWM{\etalchar{*}}21]{yang2021s3}
\textsc{Yang Z., Wang S., Manivasagam S., Huang Z., Ma W.-C., Yan X., Yumer E.,
  Urtasun R.}:
\newblock S3: Neural shape, skeleton, and skinning fields for 3d human
  modeling.
\newblock In \emph{Proceedings of the IEEE/CVF Conference on Computer Vision
  and Pattern Recognition (CVPR)} (2021).
\newblock URL: \url{http://arxiv.org/abs/2101.06571v1}.

\bibitem[YWOSH21]{yifan2021isopoints}
\textsc{Yifan W., Wu S., Oztireli C., Sorkine-Hornung O.}:
\newblock Iso-points: Optimizing neural implicit surfaces with hybrid
  representations.
\newblock In \emph{Proceedings of the IEEE/CVF Conference on Computer Vision
  and Pattern Recognition (CVPR)} (2021).
\newblock URL: \url{http://arxiv.org/abs/2012.06434v2}.

\bibitem[YYTK21]{yu2021pixelnerf}
\textsc{Yu A., Ye V., Tancik M., Kanazawa A.}:
\newblock pixelnerf: Neural radiance fields from one or few images.
\newblock In \emph{Proceedings of the IEEE/CVF Conference on Computer Vision
  and Pattern Recognition (CVPR)} (2021).
\newblock URL: \url{http://arxiv.org/abs/2012.02190v3}.

\bibitem[YZG{\etalchar{*}}21]{yu2021function4d}
\textsc{Yu T., Zheng Z., Guo K., Liu P., Dai Q., Liu Y.}:
\newblock Function4d: Real-time human volumetric capture from very sparse
  consumer rgbd sensors.
\newblock In \emph{Proceedings of the IEEE/CVF Conference on Computer Vision
  and Pattern Recognition} (2021), pp.~5746--5756.

\bibitem[YZP{\etalchar{*}}21]{yang2021recursivenerf}
\textsc{Yang G.-W., Zhou W.-Y., Peng H.-Y., Liang D., Mu T.-J., Hu S.-M.}:
\newblock Recursive-nerf: An efficient and dynamically growing nerf.
\newblock \emph{arXiv preprint arXiv:2105.09103} (2021).
\newblock URL: \url{http://arxiv.org/abs/2105.09103v1}.

\bibitem[YZX{\etalchar{*}}21]{yang2021learning}
\textsc{Yang B., Zhang Y., Xu Y., Li Y., Zhou H., Bao H., Zhang G., Cui Z.}:
\newblock Learning object-compositional neural radiance field for editable
  scene rendering.
\newblock In \emph{Proceedings of the IEEE International Conference on Computer
  Vision (ICCV)} (2021).
\newblock URL: \url{http://arxiv.org/abs/2109.01847v1}.

\bibitem[ZA21]{zobeidi2021a}
\textsc{Zobeidi E., Atanasov N.}:
\newblock A deep signed directional distance function for object shape
  representation.
\newblock \emph{arXiv preprint arXiv:2107.11024} (2021).
\newblock URL: \url{http://arxiv.org/abs/2107.11024v1}.

\bibitem[ZBDB20a]{zhong2019reconstructing}
\textsc{Zhong E.~D., Bepler T., Davis J.~H., Berger B.}:
\newblock Reconstructing continuous distributions of 3d protein structure from
  cryo-em images.
\newblock \emph{Proc. ICLR} (2020).

\bibitem[ZBDB20b]{zhong2020cryodrgn}
\textsc{Zhong E.~D., Bepler T., Davis J.~H., Berger B.}:
\newblock Reconstructing continuous distributions of 3d protein structure from
  cryo-em images.
\newblock In \emph{International Conference on Learning Representations}
  (2020).
\newblock URL: \url{http://arxiv.org/abs/1909.05215v3}.

\bibitem[ZBL{\etalchar{*}}19]{zhou2019continuity}
\textsc{Zhou Y., Barnes C., Lu J., Yang J., Li H.}:
\newblock On the continuity of rotation representations in neural networks.
\newblock In \emph{Proceedings of the IEEE/CVF Conference on Computer Vision
  and Pattern Recognition} (2019), pp.~5745--5753.

\bibitem[ZBW{\etalchar{*}}20]{zheng2020neural}
\textsc{Zheng Q., Babaei V., Wetzstein G., Seidel H.-P., Zwicker M., Singh G.}:
\newblock Neural light field 3d printing.
\newblock \emph{ACM Transactions on Graphics (TOG) 39}, 6 (2020), 1--12.

\bibitem[ZIL{\etalchar{*}}21]{zang2021intratomo}
\textsc{Zang G., Idoughi R., Li R., Wonka P., Heidrich W.}:
\newblock Intratomo: Self-supervised learning-based tomography via sinogram
  synthesis and prediction.
\newblock In \emph{Proceedings of the IEEE International Conference on Computer
  Vision (ICCV)} (2021), IEEE.

\bibitem[ZLCT21]{zehnder2021ntopo}
\textsc{Zehnder J., Li Y., Coros S., Thomaszewski B.}:
\newblock Ntopo: Mesh-free topology optimization using implicit neural
  representations.
\newblock \emph{arXiv preprint arXiv:2102.10782} (2021).
\newblock URL: \url{http://arxiv.org/abs/2102.10782v1}.

\bibitem[ZLLD21]{zhi2021semanticnerf}
\textsc{Zhi S., Laidlow T., Leutenegger S., Davison A.~J.}:
\newblock In-place scene labelling and understanding with implicit scene
  representation.
\newblock In \emph{Proceedings of the IEEE International Conference on Computer
  Vision (ICCV)} (2021).
\newblock URL: \url{http://arxiv.org/abs/2103.15875v2}.

\bibitem[ZLW{\etalchar{*}}21]{zhang2021physg}
\textsc{Zhang K., Luan F., Wang Q., Bala K., Snavely N.}:
\newblock Physg: Inverse rendering with spherical gaussians for physics-based
  material editing and relighting.
\newblock In \emph{Proceedings of the IEEE/CVF Conference on Computer Vision
  and Pattern Recognition (CVPR)} (2021).
\newblock URL: \url{http://arxiv.org/abs/2104.00674v1}.

\bibitem[ZLY{\etalchar{*}}21]{zhang2021stnerf}
\textsc{Zhang J., Liu X., Ye X., Zhao F., Zhang Y., Wu M., Zhang Y., Xu L., Yu
  J.}:
\newblock Editable free-viewpoint video using a layered neural representation.
\newblock \emph{ACM Transactions on Graphics (TOG)} (2021).
\newblock URL: \url{http://arxiv.org/abs/2104.14786v1}, \href
  {https://doi.org/10.1145/3450626.3459756}
  {\path{doi:10.1145/3450626.3459756}}.

\bibitem[ZMX{\etalchar{*}}21]{zhu2021rgbd}
\textsc{Zhu L., Mousavian A., Xiang Y., Mazhar H., van Eenbergen J., Debnath
  S., Fox D.}:
\newblock Rgb-d local implicit function for depth completion of transparent
  objects.
\newblock In \emph{Proceedings of the IEEE/CVF Conference on Computer Vision
  and Pattern Recognition (CVPR)} (2021).
\newblock URL: \url{http://arxiv.org/abs/2104.00622v1}.

\bibitem[ZRL21]{zheng2021rethinking}
\textsc{Zheng J., Ramasinghe S., Lucey S.}:
\newblock Rethinking positional encoding.
\newblock \emph{arXiv preprint arXiv:2107.02561} (2021).
\newblock URL: \url{http://arxiv.org/abs/2107.02561v2}.

\bibitem[ZRSK20]{zhang2020nerf++}
\textsc{Zhang K., Riegler G., Snavely N., Koltun V.}:
\newblock Nerf++: Analyzing and improving neural radiance fields.
\newblock \emph{arXiv preprint arXiv:2010.07492} (2020).
\newblock URL: \url{http://arxiv.org/abs/2010.07492v2}.

\bibitem[ZS20]{zhang2020fisher}
\textsc{Zhang Z., Scaramuzza D.}:
\newblock Fisher information field: an efficient and differentiable map for
  perception-aware planning.
\newblock \emph{arXiv preprint arXiv:2008.03324} (2020).

\bibitem[ZSD{\etalchar{*}}21]{zhang2021nerfactor}
\textsc{Zhang X., Srinivasan P.~P., Deng B., Debevec P., Freeman W.~T., Barron
  J.~T.}:
\newblock Nerfactor: Neural factorization of shape and reflectance under an
  unknown illumination.
\newblock \emph{ACM Trans. Graph. 40}, 6 (dec 2021).
\newblock URL: \url{https://doi.org/10.1145/3478513.3480496}, \href
  {https://doi.org/10.1145/3478513.3480496}
  {\path{doi:10.1145/3478513.3480496}}.

\bibitem[ZSG{\etalchar{*}}18]{zollhofer2018state}
\textsc{Zollh{\"o}fer M., Stotko P., G{\"o}rlitz A., Theobalt C., Nie{\ss}ner
  M., Klein R., Kolb A.}:
\newblock State of the art on 3d reconstruction with rgb-d cameras.
\newblock In \emph{Computer graphics forum} (2018), vol.~37, Wiley Online
  Library, pp.~625--652.

\bibitem[ZYLD21]{zheng2021pamir}
\textsc{Zheng Z., Yu T., Liu Y., Dai Q.}:
\newblock Pamir: Parametric model-conditioned implicit representation for
  image-based human reconstruction.
\newblock \emph{IEEE Transactions on Pattern Analysis and Machine Intelligence}
  (2021).

\bibitem[ZYQ21]{zhang2021learning}
\textsc{Zhang J., Yao Y., Quan L.}:
\newblock Learning signed distance field for multi-view surface reconstruction.
\newblock In \emph{Proceedings of the IEEE International Conference on Computer
  Vision (ICCV)} (2021).
\newblock URL: \url{http://arxiv.org/abs/2108.09964v1}.

\end{thebibliography}


\newpage 
\appendix
\section{Variable Naming Conventions}

\begin{table}[!htb]
\centering
\begin{tabular}{ll}
\toprule
Variable/Function               & Symbol                                \\
\midrule
Neural Field                    & $\field$                              \\
Neural Field Parameters         & $\params$                             \\
Encoder                         & $\mathcal{E}$                         \\
Positional Encoding             & $\gamma$                              \\
Hypernetwork                    & $\Psi$                                \\
Latent Variable                 & $\textbf{z}$                          \\
Coordinate (Input)              & $\mathbf{x}\in\mathcal{X}$            \\
Field Quantity (Output)         & $\mathbf{y}\in\mathcal{Y}$            \\
Field Coordinate Dimensionality & $m$                                   \\
Field Output Dimensionality     & $n$                                   \\
Hypersurface                    & $\mathcal{S}=\partial \mathcal{V}$    \\
Observation                     & $\mathcal{O}$                         \\
Observation (field)             & $\Omega$                              \\
Distribution                    & $\mathcal{D}$                         \\
View Direction                  & $\mathbf{\omega}_o$                   \\
Plenoptic Function              & $L(\mathbf{x},\mathbf{d})$            \\
\bottomrule
\end{tabular}
\label{tab:var_names}
\end{table}

\section{Implicit Surface Representations}

Since neural fields can store arbitrary quantities, they offer a flexible way to represent geometry and other data. We will summarize the popular output types for neural fields that can represent geometry. In addition to geometry, other types of output include radiance, BRDF parameters, deformation/warping parameters, classification weights, etc. We have discussed these output types throughout Part II. In this appendix, we focus on implicit surface representations.

\emph{Distance Functions} represent the distance to the nearest surface in some metric. These distances are useful for tasks like path planning, computational fabrication, approximating occlusion, and more. A special case of the distance function is the \textit{signed} distance function (SDF), where the sign of the distance encodes whether the surface is inside or outside. These signed distance functions can be efficiently visualized using an algorithm like sphere tracing~\cite{hart1996sphere}. A necessary but not sufficient condition of the signed distance function is the Eikonal property, which has been utilized for neural fields in many different contexts.

\emph{Occupancy} represents whether a point is considered to be inside the object or outside the object, usually with a binary $0,1$ value. Since the neural field output is continuous, the binary $0,1$ values is often approximated with a continuous function, and an appropriate isosurface value $b \in [0,1]$ needs to be selected. These occupancy fields can be visualized using raymarching. 

\emph{Volume Density} represents the density of particles that exist. Unlike occupancy, this is not a binary value and are continuous values with no upper bound. While distance functions and occupancy can only represent hard surfaces, volume density is useful for representing physically-realizable volumetric scenes like clouds, fog, or hair, where the geometric structures are too fine to be efficiently modeled as hard surfaces, and a distribution of particles can effectively approximate the true geometry. These densities can be visualized through volume rendering~\cite{novak2018monte}.

\emph{Medial Fields}~\cite{rebain2021dmf} represent the local thickness of the geometry which can be derived from the medial axis. Similarly to SDFs, these quantities can be used for rendering and approximating quantities like ambient occlusion.

\end{document}